\documentclass[preprint,12pt]{elsarticle}

\usepackage{amssymb}

\usepackage{xspace}

\newcommand{\etal}{\xspace{ et al.}\xspace}
\newcommand{\ie}{i.e.,\ }
\newcommand{\eg}{e.g.,\ }

\newcommand{\bftab}{\fontseries{b}\selectfont}

\usepackage{amsmath} %
\usepackage{amssymb}  %

\usepackage{subcaption} %

\graphicspath{{./images/}}

\newlength{\figurewidth}
\newlength{\figureheight}
\usepackage{hyperref}
\usepackage[capitalize,noabbrev]{cleveref}
\usepackage[binary-units=true]{siunitx}
\usepackage{array,multirow}
\usepackage[binary-units=true]{siunitx}
\DeclareSIUnit[number-unit-product = {}]{\inch}{^{\prime\prime}}

\crefname{algocf}{Algorithm}{Algorithms}
\crefname{table}{Table}{Tables}
\crefname{chapter}{Chapter}{Chapters}
\crefname{equation}{Equation}{Equations}
\crefname{section}{Section}{Sections}

\usepackage{tikz}
\usetikzlibrary{arrows}
\usetikzlibrary{positioning,calc}
\usetikzlibrary{decorations.pathreplacing}
\usetikzlibrary{decorations.markings}
\usetikzlibrary{fit}
\usetikzlibrary{shapes.callouts}

\tikzset{
  big arrow/.style={
    decoration={markings,mark=at position 1 with {\arrow[scale=1.7]{latex}}},
    postaction={decorate},
    shorten >=0.4pt}}
\tikzset{
  big 2arrow/.style={
    decoration={markings,mark=at position 0 with {\arrow[scale=-1.7]{latex}},mark=at position 1 with {\arrow[scale=1.7]{latex}}},
    postaction={decorate},
    shorten >=0.4pt}}
\journal{Robotics and Autonomous Systems}

\begin{document}

\begin{frontmatter}

\title{Feature-based Visual Odometry Prior\\ for Real-time Semi-dense Stereo SLAM}

\author{Nicola Krombach}
\ead{krombach@ais.uni-bonn.de}
\author{David Droeschel}
\ead{droeschel@ais.uni-bonn.de}
\author{Sebastian Houben}
\ead{houben@ais.uni-bonn.de}
\author{Sven Behnke}
\ead{behnke@ais.uni-bonn.de}

\address{Autonomous Intelligent Systems Group, Computer Science Institute VI\\
University of Bonn, Friedrich-Ebert-Allee 144, 53113 Bonn, Germany 
}

\begin{abstract}

Robust and fast motion estimation and mapping is a key prerequisite for autonomous operation of mobile robots. The goal of performing this task solely on a stereo pair of video cameras is highly demanding and bears conflicting objectives: on one hand, the motion has to be tracked fast and reliably, on the other hand, high-level functions like navigation and obstacle avoidance depend crucially on a complete and accurate environment representation. 
In this work, we propose a two-layer approach for visual odometry and SLAM with stereo cameras that runs in real-time and combines feature-based matching with semi-dense direct image alignment.
Our method initializes semi-dense depth estimation, which is computationally expensive, from motion that is tracked by a fast but robust keypoint-based method.
Experiments on public benchmark and proprietary datasets show that our approach is faster than state-of-the-art methods without losing accuracy and yields comparable map building capabilities. Moreover, our approach is shown to handle large inter-frame motion and illumination changes much more robustly than its direct counterparts.

\end{abstract}

\begin{keyword}

visual simultaneous localization and mapping, visual odometry, feature-based SLAM, semi-dense SLAM

\end{keyword}

\end{frontmatter}
	\begin{tikzpicture}[remember picture,overlay]
		        \node[anchor=north west,align=left,font=\sffamily,yshift=-1.2cm] at (current page.north west) {%
				In: Robotics and Autonomous Systems (RAS) 
						                    };
								                      \node[anchor=north east, align=right,font=\sffamily,yshift=-1.2cm] at (current page.north east) {%
											                                  DOI: \href{https://doi.org/10.1016/j.robot.2018.08.002}{10.1016/j.robot.2018.08.002}
															                              };
	\end{tikzpicture}%

\section{Introduction}

A key feature of nearly all mobile robots is a reliable, robust, and fast state estimation that is essential for most high-level operations like autonomous navigation or exploration.
Many mobile robots rely on cameras since they are inexpensive and lightweight and can be used for a variety of tasks including visual obstacle detection, 3D scene reconstruction, visual odometry, and even visual simultaneous localization and mapping~(SLAM).

Visual odometry~(VO) means estimating the egomotion solely from images captured by a monocular or stereo camera system. There are a large variety of VO methods that can be classified into feature-based and direct methods. SLAM broadens this task by also requiring to compute a representation of the robot's surrounding referred to as map.
Most VO and SLAM methods are feature-based and work by detecting keypoints and matching them between frames. 
In contrast, direct methods estimate the camera motion by minimizing the photometric error over all pixels.
Since this minimization consists of aggregating the matching cost over all image pixels, it is computationally more demanding than determining the reprojection error of sparse set of feature points. Hence, direct methods are often computationally more demanding, yet more accurate, than their feature-based counterparts.
\begin{figure*}[th]
\centering
      \includegraphics[width=\linewidth]{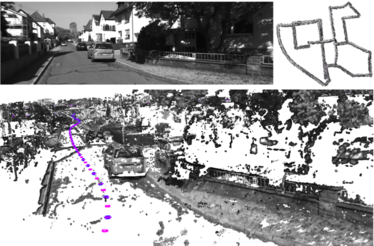}
    \caption[Semi-dense 3D reconstruction of KITTI Dataset 00]{Semi-dense 3D reconstruction of KITTI 00: Top left: Camera image. Bottom: Semi-dense 3D reconstruction with colored camera trajectory (key frames blue, feature-based tracked frames pink). Top right: Bird's eye view of the complete reconstructed scene.}
    \label{fig:kittireconstruction3}
\end{figure*}
In this work, we propose a novel approach that combines direct image alignment with sparse feature matching for stereo cameras.
By combining both paradigms, we are able to process images with high frame rate and to also track large inter-frame motion while maintaining the accuracy and quality of a direct method.
Due to the distinctiveness of the tracked features, our method performs well on datasets with low frame rates, which is often a problem for direct methods as they need sufficient image overlap.

We extend monocular LSD-SLAM~\cite{engel14eccv} to work with a stereo setup and restrict semi-dense matching to key frames for achieving a higher frame rate. In order to estimate the motion between key frames, we employ a feature-based VO method and use the estimated motion as initialization for the direct image alignment. Thus, we restrict the search space for direct image alignment and gain real-time performance. This paper builds upon our recent work~\cite{krombach2016combining} where we introduced a VO algorithm deploying both feature-based and semi-direct matching techniques. Here, we expand this approach to a fully-fledged SLAM system.

\section{Related Work}

Visual odometry (VO) and visual SLAM are both vivid areas of research and have seen rapid progress in the past years. Currently, feature-based and direct approaches present two of the dominant paradigms.

\subsection{Feature-based Methods}

The general pipeline for feature-based methods can be summarized as follows:
Image features are detected and either matched between frames or tracked over time. Based on these feature correspondences, the relative motion between two frames is computed. To compensate for drift, many methods make use of pose-graph optimization.

Popular feature-based methods are MonoSLAM~\cite{monoslam} and Parallel Tracking and Mapping (PTAM)~\cite{klein07parallel}. PTAM is a widely used feature-based monocular SLAM method, which allows robust state estimation in real-time and has been successfully used on MAVs with monocular cameras~\cite{weiss11monocular}.
ORB-SLAM~\cite{ORB} has been proposed as a monocular visual SLAM method that tracks ORB features in real-time and furthermore uses them for local and global bundle adjustment, and candidate retrieval for loop closing. 
Most monocular methods have a dedicated initialization stage where both the map and the camera movement are estimated at the same time. To this end, the initial movement must bear a certain amount of parallax and can, thus, not be completely arbitrary. On a second note, when using monocular methods, additional sensors are needed to estimate the absolute scale of a scene. In contrast, stereo, depth camera or multi-camera methods~\cite{Geiger2011IV,bacs} have a constant measurement of scale and, hence, do neither suffer from scale drift nor do they need a particular initialization stage. A later extension of ORB-SLAM~\cite{mur2017orb2} incorporates stereo and depth cameras by using a dense depth estimation in order to initialize new ORB map points. On the other hand S-PTAM~\cite{SPTAM,SPTAM2017} matches sparse features directly and does not rely on a stereo depth images that would need to be computed in a preprocessing step. Likewise, a multi-camera version of ORB-SLAM~\cite{houben2016orbslam} has been presented that does not rely on dense depth images but triangulates sparse features if stereo pairs are given. 
Due to their complementary nature, feature-based methods also incorporate readings from an inertial measurement unit~(IMU) as high-frequency short-term estimates between frames. Straightforwardly VO and inertial readings are fused in a filter-based approach~\cite{achtelik11onboard,Leutenegger15122014} which is termed \emph{loose coupling}. In particular the work by Forster~\cite{forster2017_manifold_preintegration} has allowed for \emph{tight-coupling}, \ie integrating both IMU readings and visual odometry in a single non-linear cost function. This technique has since then found its way into another variant of ORB-SLAM~\cite{mur2017visual}.

In our work, we rely on a well-established and efficient feature-based library for stereo visual odometry~\cite{Geiger2011IV} which provides a good trade-off between accuracy and runtime.

\subsection{Direct Methods}

In contrast to feature-based methods, which abstract images into a sparse set of feature points, direct methods use the entire image information in order to minimize the photometric error. In an early work~\cite{comport}, Comport \etal formulate pixel-wise quadrifocal constraints for sparse corresponding stereo matches in two subsequent pairs of images from a stereo setup.
If extended to the entire image data, these methods are computationally more intensive than feature-based methods. First introduced for monocular cameras by Engel\etal\cite{engel2013semi,engel14eccv}, LSD-SLAM (Large-scale Semi-Dense SLAM) estimates a pixel-wise inverse depth for a reference frame by means of successive small-baseline stereo estimations. The inverse depth and the according pixel-wise variance is propagated to a new keyframe as soon as the stereo baseline becomes too large. As the pose and point-wise depth estimation is done by minimizing a cost function on image data via gradient descent, the motion must be small or a good initial pose estimate must be given in order to not converge to a local minimum. This is the reason that large inter-frame motion is problematic. The relative poses between the keyframes are asynchronously optimized in a pose graph approach in which two keyframes are connected by a rigid transform with an additional scaling factor. 
Direct approaches have been extended to stereo and RGB-D cameras. Engel \etal~\cite{engel2015_stereo_lsdslam} use both fixed-baseline stereo and temporal stereo (as in monocular LSD-SLAM) to refine the depth estimate of the current reference keyframe. Since static stereo is performed initially for every new stereo keyframe, a more reliable depth estimate is available right from the beginning. Hence, the pose and depth refinement become more robust and can deal with larger inter-frame motion.
With RGB-D cameras, direct methods are more straightforward as a point-wise depth estimate with constant variance is given in every frame. Stückler and Behnke~\cite{mrsm} transform the depth image into a coarser representation, named a surfel map, for aggregation and track the camera motion with ORB features. They later utilize this approach for dense image registration and combine it with a sparse feature matcher in order to compute visual odometry \cite{gutt}. 
Dense direct methods often need to use GPUs to achieve real-time performance~\cite{newcombe2010dense_recon,pizzoli2014remode}. By using only pixels with sufficient gradient, LSD-SLAM~\cite{engel14eccv} reduces the computational demand and real-time semi-dense SLAM becomes possible with a strong CPU. The extension to stereo cameras \cite{engel2015_stereo_lsdslam} uses both fixed-baseline stereo depth and temporal multi-view stereo in order to estimate a semi-dense environment representation.
Recently, a method for directly optimizing the depth of sparse feature points for visual odometry, DSO (Direct Sparse Odometry), has been proposed by Engel\etal~\cite{engel2018_dso}. Building upon the same optimization scheme like LSD-SLAM, they optimize for all parameters (including the depth of numerous sparsely chosen image points) for a sliding window of a few keyframes. In order to increase robustness, the cameras must be calibrated photometrically and exposure times have to be taken into account.
Sch\"{o}ps \etal~\cite{schops2015_3drecon} use a visual-inertial odometry approach to compute a short time horizon of camera poses and obtain a dense stereo estimation by plane sweeping multiple images\cite{gallup2007_planesweep}. The latter method can be scheduled in parallel and achieves real-time applicability by use of a GPU.

\subsection{Hybrid Methods}

Regarding the reconstruction of the environment, direct methods have the advantage of estimating a dense map while feature-based methods can only rely on the sparse features that have been tracked.
Dense direct methods are computationally demanding and are often executed as a final step for estimating a globally consistent dense map after pose tracking with sparse interest-points succeeded.
To speed up global optimization, already tracked sparse feature-points can be used as initialization for dense mapping~\cite{ORB-SemiDense}.
The semi-direct method by Forster \etal~\cite{Forster2014ICRA} uses direct motion estimation for initial feature extraction and continues by using only these features. A novel release includes edgelets as features, encompasses IMU readings, and yields a significant speedup \cite{forster2017svo}. 
A recent paper by Piazza \etal~\cite{piazza2018_3dmeshrecon} presents a real-time capable algorithm to compute and update a 3D manifold mesh on a CPU. This allows for deriving a dense 3D map from a set of sparse points as provided by any of the above SLAM systems.
A combination of a feature-based and a direct method has been presented by Younes \etal~\cite{younes2018_fdmo} as feature-assisted direct monocular odometry. They present a VO method that is based on DSO\cite{engel2018_dso} but uses feature-based tracking when optimization yields little relative improvement.
In contrast, we always continuously combine feature-based and semi-dense direct tracking over time, taking advantage of the fast tracking from the feature-based method and the accurate alignment of image gradients from direct methods. The feature-based tracking result is immediately fed to the direct tracking at runtime as an initial guess.

\section{Method}
Our method is mainly based on the monocular version of LSD-SLAM that we extended to work with stereo cameras.
By using stereo cameras instead of a single monocular camera, the absolute scale of the scene becomes observable, eliminating scale ambiguity and the need for additional sensors, \eg inertial measurement units.

To ensure a high frame rate, we restrict the semi-dense direct alignment to key frames only and estimate the motion for all other frames by the feature-based method LIBVISO2~\cite{Geiger2011IV}.
This motion estimate is used as initial estimate for direct alignment of key frames. The semi-dense environment mapping runs in a parallel thread. 

\subsection{Notation}
We follow the notation of Engel et al.~\cite{engel14eccv}.
The monochrome stereo images captured at time $i$ are denoted with $I^{l/r}_{i}:\Omega \subset \mathbb{R}^2 \rightarrow \mathbb{R}$, with image domain $\Omega$.
Each key frame $KF_i = \{ I_i^l, I_i^r, D_i, V_i\}$ consists of the left and right stereo images $I^{l/r}_i$, the semi-dense inverse depth map $D_i: {\Omega_D}_i \rightarrow \mathbb{R}^+$, and the corresponding pixel-wise variance map $V_i: {\Omega_D}_i \rightarrow \mathbb{R}^+$.
The inverse of the depth $z$ of a pixel is denoted as $d=z^{-1}$.
Camera motions are represented as twist coordinates $\boldsymbol{\xi} \in \mathfrak{se}(3)$ with corresponding transformation $\boldsymbol{T_{\xi}} \in SE(3)$.
A 3D point $\boldsymbol{p} = (p_x,p_y,p_z)^T$ is projected into image coordinates $\boldsymbol{u} = (u_x, u_y, 1)^T$ by the projection function $\pi(\boldsymbol{p}) := \boldsymbol{K} \left(p_x/p_z, p_y/p_z, 1\right)^T$ 
with intrinsic camera matrix $\boldsymbol{K}$.
Thus, the inverse projection function $\pi^{-1}(\boldsymbol{u},d)$ maps a pixel with corresponding inverse depth to a 3D point $\boldsymbol{p} = \pi^{-1}(\boldsymbol{u},d) := \left(d^{-1}\boldsymbol{K}^{-1}\boldsymbol{u}\right)^T$.

\subsection{LSD-SLAM}

The processing pipeline of LSD-SLAM~\cite{engel14eccv} consists of the three main components: Tracking, depth map estimation, and global map optimization.

Tracking, \ie frame-wise relative pose estimation, is based on maximizing photo-consistency and thus minimizing the photometric error between the current frame and the most recent key frame using Gauss-Newton optimization:
\begin{equation}
E(\boldsymbol{\xi}) := I_{KF}(\pi(\boldsymbol{p})) - I(\pi(\boldsymbol{T_{\xi} ~p})) \enspace , \forall p \in \Omega
\label{tracking}
\end{equation}
where $\boldsymbol{p}$ is warped from $I$ to $I_{KF}$ by $\boldsymbol{\xi}$.
New frames are tracked towards a key frame and the rigid body motion of the camera $\boldsymbol{\xi} \in \mathfrak{se}(3)$ is estimated. 

In the depth map estimation, tracked frames are then used to refine the existing depth map of the key frame by many short-baseline stereo comparisons.
Given the transformation between a tracked frame and the key frame,
that has been estimated prior in the tracking, the epipolar lines are calculated.
Afterwards, for each pixel with sufficient gradient its depth hypothesis is updated
with stereo measurements. The depth is calculated by finding the best matching
point along the epipolar line, that is the point which minimizes the SAD error
measured over five equidistant points along the epipolar line.
Given the estimated depth the depth map of the most recent key frame is then refined by either creating new depth hypotheses or improving existing ones.
New key frames are created when the distance exceeds a certain threshold and are initialized by propagating depth of the previous key frame towards the new frame.
Once a key frame is replaced, it is added to the pose-graph for further refinement and loop closing.

\subsection{LIBVISO2}
LIBVISO2~\cite{Geiger2011IV} is a fast feature-based VO library for monocular and stereo cameras.
Similar to other feature-based methods, it consists of feature matching over subsequent frames and egomotion estimation by minimizing the reprojection error.
Features are extracted by filtering the images with a corner and blob mask and performing non-maximum and non-minimum suppression on the filtered images.
Starting from all feature detections in the current left image, candidates are matched in a circular fashion over the previous left image, the previous right image, the current right image, and back to the current left image. 
If the first and last features of such a circle match differ, the match is rejected.
Based on all found matches, the egomotion is then estimated by minimizing the reprojection error using Gauss-Newton %
and outliers are removed using RANSAC.

\subsection{Semi-dense Alignment of Stereo Key Frames}

We build upon the open source release of monocular LSD-SLAM and extend it with stereo functionality.
In contrast to monocular visual odometry, stereo allows to compute absolute depth maps and, thus, does not suffer from scale drift.
By extending LSD-SLAM to stereo, we combine the existing depth map computation over time with instant stereo depth from the current image pair.
While monocular LSD-SLAM uses a random initialization and has to bootstrap over the first frames, we take advantage of using stereo cameras and initialize our method with absolute depth values.
We use ELAS~\cite{Geiger2010ACCV} to compute the depth map of the initial key frame. 
The following key frames are registered with their previous key frame by minimizing the photometric error of their left reference frames as well as the depth error.
While in the monocular case, absolute depth is not observable; with stereo cameras absolute depth is observable for every incoming stereo image pair.
This allows us to minimize the depth error in addition to the photometric error.
Hence, for direct tracking with stereo, we extend the minimization of the photometric residual $r_p$ to take the depth residual $r_d$ into account:

\begin{equation}
\begin{split}
r_p(\boldsymbol{p},\boldsymbol{\xi}) &= \left\| I_{KF_i}(\pi(\boldsymbol{p})) - I_j(\pi(\boldsymbol{T_{\xi}~p})) \right\|,  \\
r_d(\boldsymbol{p},\boldsymbol{\xi}) &= \left\| D_{KF_i}(\pi(\boldsymbol{p})) - D_{stereo_j}(\pi(\boldsymbol{T_{\xi}~p})) \right\|,
\end{split}
\end{equation}

\noindent where $\boldsymbol{\xi}$ is the camera motion from the i-th key frame to the new j-th frame and $D_{stereo_j}$ is the initial instant stereo depth map of the j-th frame.
The minimization is performed using a weighted least squares formulation and solved with the Gauss-Newton method.
The residual is formulated as stacked residual $\boldsymbol{r}$ and is weighted with a $2\times2$ weight matrix $\boldsymbol{W}$:
\begin{equation}
  \begin{split}
  \boldsymbol{r}(\boldsymbol{\xi}) &= \boldsymbol{W}(\boldsymbol{\xi}) \sum_{\boldsymbol{p}\in{\Omega_D}_i} \begin{pmatrix} h(r_p(\boldsymbol{p},\boldsymbol{\xi}))\\h(r_d(\boldsymbol{p},\boldsymbol{\xi}))\end{pmatrix}; \\
  \quad \boldsymbol{W}(\boldsymbol{\xi}) &= \sum_{\boldsymbol{p}\in{\Omega_D}_i}\begin{pmatrix} w_p(r_p(\boldsymbol{p},\boldsymbol{\xi}))&0\\0&w_d(r_d(\boldsymbol{p},\boldsymbol{\xi})\end{pmatrix},
  \end{split}
\end{equation}

\noindent where both residuals are weighted with the Huber norm denoted as $h(\cdot)$.

\subsection{Hybrid Odometry Estimation}
\begin{figure*}[ht!]
      \centering
  	\includegraphics[width=0.7\textwidth]{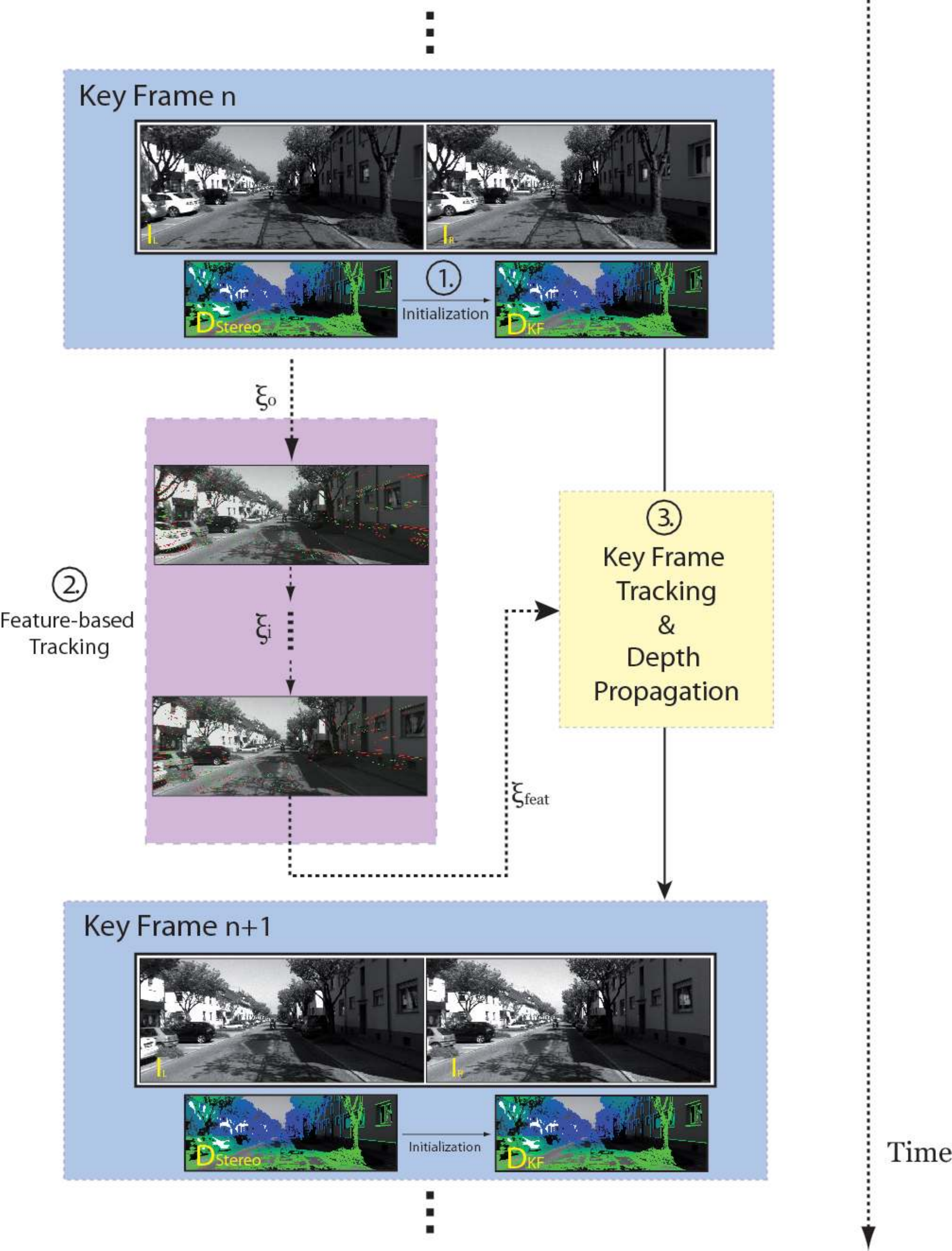} %
      \caption{Overview of our combined semi-direct approach. While direct tracking is only performed on key frames, feature-based tracking is performed for frames in between. The output of the feature-based odometry serves as prior for the direct tracking. }
      \label{overview}
\end{figure*}
Our idea is to take advantage of the different strengths of both approaches and, thereby,  
combine fast feature matching with precise semi-dense image alignment for efficient and reliable state estimation.
The modular structure of our approach is illustrated in \cref{overview}.

We initialize the first key frame with a dense depth map computed by ELAS (\cref{overview} (1)).
Subsequent frames are then tracked towards the key frame incrementally using feature-based LIBVISO2 (\cref{overview} (2)). 
The relative poses of the tracked frames are concatenated and form the relative pose of the camera to the key frame:
\begin{equation}
\boldsymbol{\xi}_{feat} = {\boldsymbol{\xi}_i}_n \circ {\boldsymbol{\xi}_i}_{n-1} \circ \cdots \circ {\boldsymbol{\xi}_i}_0 \enspace .
\end{equation} 
The current absolute pose of the camera at step $j$ and key frame $i$ can be retrieved by:
\begin{equation}
{\boldsymbol{\xi}_i}_j = {\boldsymbol{\xi}_{KF}}_{i} \circ {\boldsymbol{\xi}_i}_{j-1} \enspace .
\end{equation}

We perform feature-based odometry as long as the motion is sufficiently small. As soon as the motion exceeds the motion threshold
\begin{equation}
\epsilon_{motion} = \frac{1}{n} \sum^n_{k=1} \sqrt{\left(\boldsymbol{u}^k_i-\boldsymbol{u}^k_{i-1}\right)^2},
\end{equation}
we perform direct registration again  (\cref{overview} (3)) and the previous key frame is replaced with the new frame, where $n$ is the number of matched feature points and $(\boldsymbol{u}^k_i)$ and $(\boldsymbol{u}^k_{i-1})$ are corresponding feature matches between the current and the previous image. 

The motion $\boldsymbol{\xi}_{feat}$ serves as initial estimate for the direct registration of the new frame towards the key frame:
\begin{equation}
 {\boldsymbol{\xi}_{KF}}_{i+1} = {\boldsymbol{\xi}_{KF}}_{i} \circ \boldsymbol{\xi}_{feat} \enspace .
\end{equation}

This allows us to track larger motions faster and more robustly. 
The depth map of a new key frame is initialized by instant stereo correspondences and then fused with the previous depth map by propagation as described in the next section. 
Once a new key frame is initialized, we start feature-based matching again.

\subsection{Map Update}\label{sec:map_update}

\begin{figure}
      \centering
  	\includegraphics[width=0.8\linewidth]{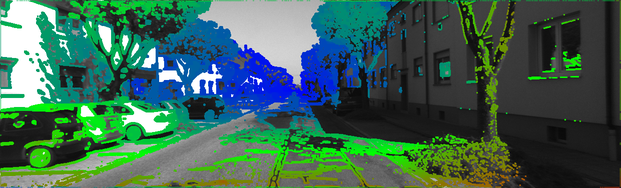}\\\vspace{1mm}
  	\includegraphics[width=0.8\linewidth]{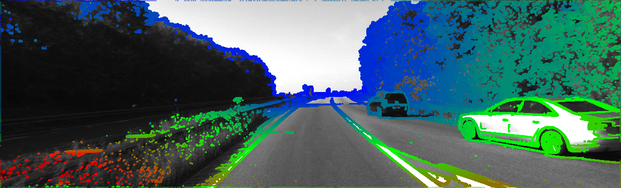}%
      \caption{Computed semi-dense depth maps for KITTI datasets (sequences 00 and 01). Color depicts distance to the sensor.}
      \label{depthmaps}
 \end{figure}
The depth map of each key frame is updated with instant stereo measurements as well as with propagated depth from the previous key frame.
If a new key frame is created, the depth map is computed by instant stereo from the left and right images.
For reasons of runtime, we use a simple but fast block matching along epipolar lines instead of ELAS~\cite{Geiger2010ACCV} which is used for the first keyframe only.
Corresponding pixels are found by minimizing the sum of absolute distances~(SAD) error over a 15$\times$15 pixel window.
After initializing the depth map with stereo measurements, the depth estimates are refined by propagating depth hypotheses of the previous %
depth map to the new frame:
\begin{equation}
\boldsymbol{p}_{new}(\boldsymbol{p}) = \boldsymbol{R}_{C,KF} ~\boldsymbol{p}+\boldsymbol{t}_{C,KF},
\end{equation}
where $\boldsymbol{p}$ is the 3D point in the previous key frame.
The rotation $\boldsymbol{R}_{C,KF}$ and translation $\boldsymbol{t}_{C,KF}$ describe the coordinate transformation from the key frame coordinate system $KF$ to the candidate coordinate system $C$. 
If the residual between the instant and propagated depth is high, the depth value with smaller variance is chosen. 
Otherwise both estimates---$d_{stereo}$ and $d_{prop}$---are fused to a new depth estimate $d_{new}$ as a variance-weighted sum:
\begin{equation}
d_{new} = (1-\omega) ~d_{stereo} + \omega ~d_{prop}\enspace .
\end{equation}

The variance $\omega$ for each depth hypothesis is determined as described by Engel et al.~\cite{engel2013iccv}.

For fish eye lenses we extend this weighting scheme: as fish eye lenses suffer from distortion at the image borders, we further increase the variance of a pixels depth hypothesis depending on the distance $r(u,v)$ to the optical center $(c_x,c_y)$:
\begin{equation}
r(u,v) = \sqrt{(u-c_x)^2+(v-c_y)^2}.
\end{equation} 
\cref{depthmaps} shows the resulting semi-dense depth maps for two KITTI sequences.

\subsection{Global Map Optimization}\label{sec:global_map_optimization}

  \begin{figure}
      \centering
  	\includegraphics[width=0.9\linewidth]{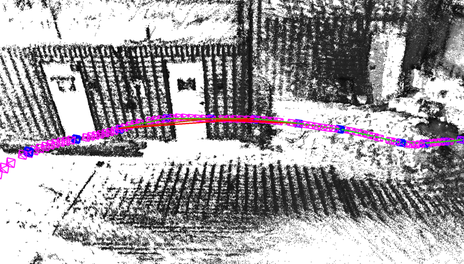}\\\vspace{3mm}
  	 \includegraphics[width=0.9\linewidth]{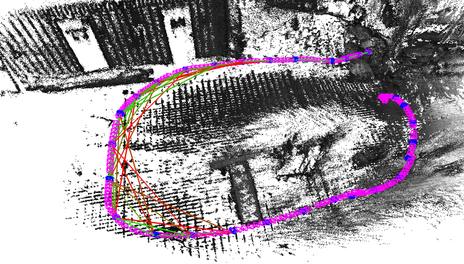}
      \caption[Global Map Optimization]{Trajectory of a camera and reconstructed map. Pink frusta indicate camera poses for every frame taken, blue frusta denote keyframes. Edges between keyframes are color-coded using the reprojection error between them. The coding ranges from green (low reprojection error) to red (high reprojection error).
      In the global map optimization edges between key frames are added to the pose-graph. Nearby key frames usually match better than distant key frames (visualized as line color: green for a good match, red for a poor match). The top picture shows the graph before optimization with many red edges indicating poor consistency. Below: After graph optimization the key frame poses have been refined yielding a lower overall error.}
      \label{fig:slam1}
   \end{figure}

So far, we presented an approach performing incremental visual odometry by directly tracking incoming stereo images in combination with semi-dense depth reconstruction. This gradual pose estimation technique accumulates errors over time.

In order to alleviate this caveat we use G\ensuremath{^2}O~\cite{g2o} for global pose graph optimization.
The pose graph is constructed from the key frame poses as vertices and their relative transformations as edges.
Instead of optimizing $SIM(3)$ constraints, \ie assuming that two separate camera frames are related via a rigid-body motion with an additional unknown scaling factor, as in the original proposal by Engel \etal \cite{engel2013iccv}, we set constraints between key frames as their $SE(3)$ rigid-body motion as due to our stereo setup we have no scalar ambiguity.
Once a key frame is created, its pose is added to the key frame graph as a vertex $V_l$.
Subsequently, we search the existing vertices in the graph for additional constraints that can refine the pose graph.
To this end, the closest $n$ key frames, that have sufficient scene overlap in terms of parametrizable euclidean distance as well as parametrizable angular overlap, are matched against the newly created key frame:
We estimate the transformation between both frames both ways, by registering the constraint candidate against the key frame and vice versa using semi-dense direct matching as described in \cref{tracking}. 
Only if the matching succeeds for both directions and the resulting transformations agree, they are added to the graph as an additional constraint.
All edge constraints $E_{ij}$ between the vertices $V_i, V_j$ define a cost function $C$ that is optimized using G\ensuremath{^2}O:
\begin{equation}
C(V_0, ..., V_l) = \sum\limits_{E_{ij}} \left\| T_V(V_i,V_j) - T_E(E_{ij}) \right\|^2.
\end{equation} 
where $T_V$ denotes the relative pose between two key frames corresponding to $V_i$ and $V_j$ and $T_E$ yields the relative pose held by the edge $E_{ij}$.
\cref{fig:slam1} shows an exemplifying result of the global optimization stage.

\section{Evaluation\label{chap:Evaluation}}

For the evaluation of our semi-direct approach we perform experiments on three challenging stereo datasets:
the well-known KITTI-dataset~\citep{Geiger2012CVPR}, the EuRoC dataset \cite{Burri25012016} and a proprietary dataset recorded with our high-performance MAV presented in \cref{sec:MAV}.
The datasets differ in terms of frame rate, apparent motion, stereo baseline, and base platform.
All experiments have been conducted on an Intel Core i7-4702MQ running at \SI{2.2}{\giga\hertz} with \SI{8}{\giga\byte} RAM.

We compare the quality of our combined approach in terms of accuracy and runtime to LSD-SLAM~\citep{engel14eccv} and LIBVISO2~\citep{Geiger2011IV}, as well as to state-of-the-art methods like S-PTAM~\citep{SPTAM} and ORB-SLAM~\citep{ORB}. The execution of the referred methods has been obtained using the provided default parameters.

As ground truth for all sequences is available, we employ the evaluation metrics by Sturm \cite{sturm12iros} and measure the absolute trajectory error (ATE) by computing the root mean squared error (RMSE) over the whole trajectory.
In addition, we also provide the median error for better insight, because single outliers can greatly affect the final result.
The ATE is a popular measure for the evaluation of visual SLAM systems as it measures the Euclidean distance between ground truth poses and estimated poses at corresponding timestamps, and thereby allows to evaluate the global consistency of SLAM systems.
In a first step the trajectories are rigidly aligned because they stem from different coordinate systems.
Moreover, a similarity alignment is performed for the monocular systems to estimate the absolute scale of the estimated trajectory.
For an intuitively accessible visualization, trajectories are always shown in bird's eye perspective neglecting height differences in the trajectory.
In the following sections we first present detailed results for each dataset, individually.
Moreover, we evaluate the performance of visual SLAM compared to pure visual odometry and provide quantitative result.
Afterwards, we shortly summarize the obtained average results for accuracy and runtime and conclude with qualitative results of our 3D reconstruction.  
Finally, the following videos summarize our visual SLAM approach on the EuRoc dataset \footnote{https://www.youtube.com/watch?v=7NkHf6syRIo} and visual odometry on the KITTI dataset \footnote{https://www.youtube.com/watch?v=PRYgnIBDVGI}.

\subsection{KITTI}
\begin{table}
\small
\begin{center}
\begin{tabular}{c||c||c||c||c}
    KITTI & \multicolumn{4}{c} {\textbf{Absolute Trajectory Error RMSE (Median) in m}}\\
Sequence & Ours 					& LIBVISO2 & ORB-SLAM & S-PTAM\\
\hline
    00 & \bftab{\hphantom{0}5.79 \hphantom{0}(4.54)}   	& 29.71 (18.49)  			& \hphantom{00}8.30 \hphantom{00}(6.04)            	& \hphantom{00}7.83 \hphantom{00}(6.30) \\
    01 & \bftab{61.55 (54.57)} 				& 66.54 (60.46) 			& 		335.52 (303.79)       		 	& 204.65 (157.10) \\
    02 & 18.99 (14.38)          			& 34.26 (27.36) 			& \bftab{\hphantom{0}18.66 \hphantom{0}(15.03)} 	& \hphantom{0}20.78 \hphantom{0}(17.28) \\
    03 & \bftab{\hphantom{0}0.63 \hphantom{0}(0.52)}   	&\hphantom{0}1.67 \hphantom{0}(1.54)   	& \hphantom{0}11.91 \hphantom{00}(9.19)           	& \hphantom{0}10.53 \hphantom{0}(10.41) \\
    04 & \bftab{\hphantom{0}0.67 \hphantom{0}(0.46)}   	& \hphantom{0}0.80 \hphantom{0}(0.66)   & \hphantom{00}2.15 \hphantom{00}(1.73)            	& \hphantom{00}0.98 \hphantom{00}(0.88) \\
    05 & \hphantom{0}5.47 \hphantom{0}(4.14)           	& 22.14 (19.07) 			& \hphantom{00}4.93 \hphantom{00}(4.73)            	& \bftab{\hphantom{00}2.80 \hphantom{00}(2.24)} \\
    06 & \bftab\hphantom{0}{2.06 \hphantom{0}(1.80)}   	& 11.54 (10.26) 			& \hphantom{0}16.01 \hphantom{0}(15.56)          	& \hphantom{00}4.00 \hphantom{00}(4.01) \\
    07 & \hphantom{0}2.34 \hphantom{0}(1.67)            & \hphantom{0}4.41 \hphantom{0}(4.37)   & \hphantom{00}4.30 \hphantom{00}(3.65)           	& \bftab{\hphantom{00}1.80 \hphantom{00}(1.53)} \\	
    08 & \hphantom{0}8.42 \hphantom{0}(7.04)            & 47.67 (34.84) 			& \hphantom{0}38.80 \hphantom{0}(18.12)          	& \bftab{\hphantom{00}5.13 \hphantom{00}(4.26)} \\
    09 & \bftab{\hphantom{0}5.46 \hphantom{0}(3.33)}   	& 89.83 (77.57) 			& \hphantom{00}7.46 \hphantom{00}(6.91)            	& \hphantom{00}7.27 \hphantom{00}(4.61) \\
    10 & \bftab{\hphantom{0}1.68 \hphantom{0}(1.37)}   	& 49.35 (36.00) 			& \hphantom{00}8.35 \hphantom{00}(7.55)            	& \hphantom{00}2.08 \hphantom{00}(1.70) \\
\hline
mean   & \bftab{10.28 \hphantom{0}(8.53)}  		& 32.54	(26.42) 			& \hphantom{0}41.49 \hphantom{0}(35.66) 		&  \hphantom{0}25.74 \hphantom{0}(20.26) \\
w/o S 01 & \bftab{\hphantom{0}5.15 \hphantom{0}(3.93)} & 29.14 (23.02) 			& \hphantom{0}12.09 \hphantom{00}(8.85) 		& \hphantom{00}7.85 \hphantom{00}(6.57)\\
\hline
\end{tabular}
\end{center}
\caption{ATE Results on KITTI Dataset}
\label{tab:kitti_ate}
\end{table}

The KITTI dataset~\citep{Geiger2012CVPR} is a very popular dataset for the evaluation of visual and laser-based odometry or SLAM methods.
It contains $22$ stereo sequences accompanied by laser scans, and ground truth from a localization unit consisting of a GPS and an IMU.
The stereo camera rig and the laser scanner are mounted on top of a standard station wagon---the autonomous driving platform Annieway~\citep{ROB:ROB20252}. The stereo rig has a baseline of approximately \SI{54}{\cm}.

\begin{figure*}[ht!]
  \centering
    \includegraphics[width=0.5\linewidth]{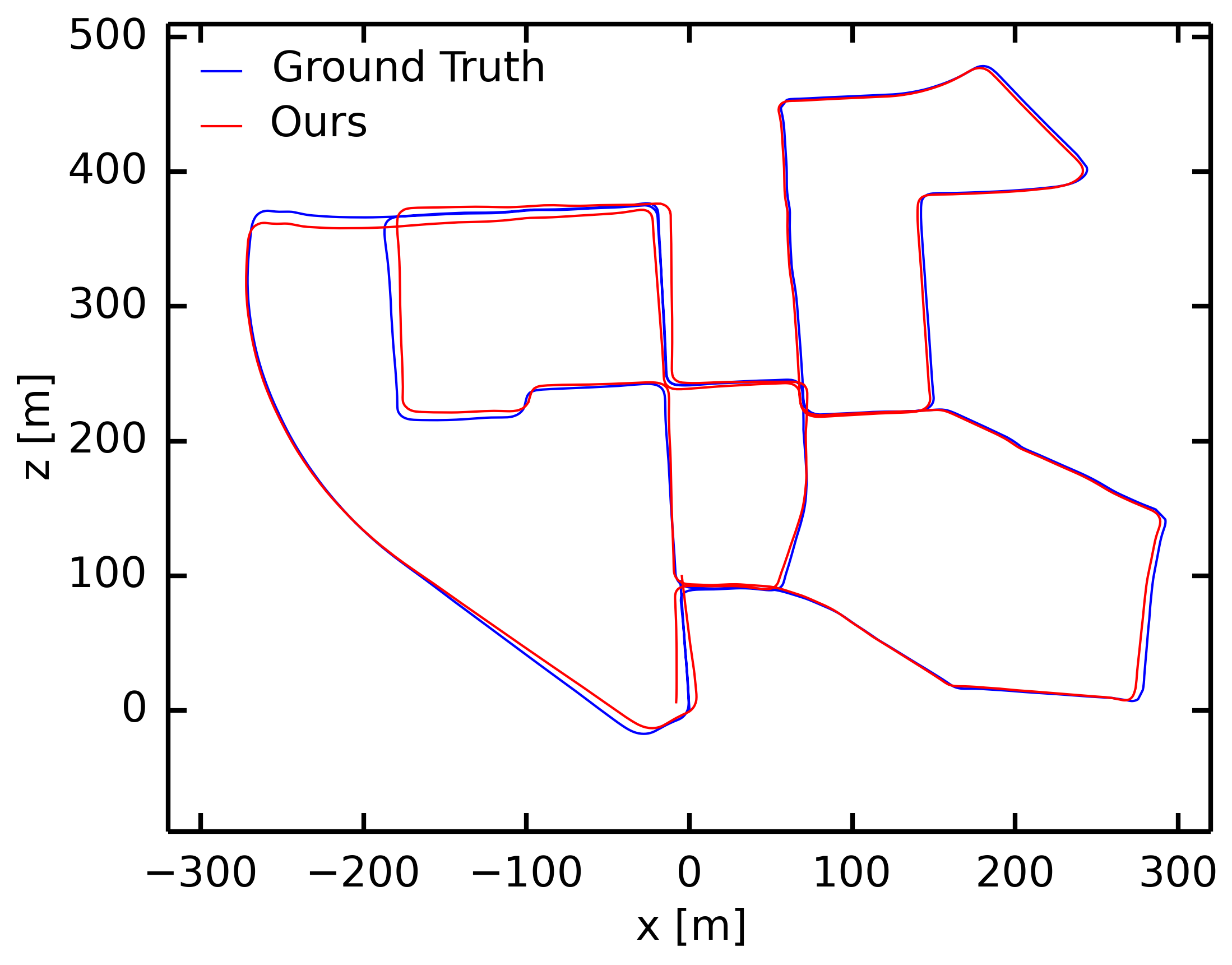}\hfill
    \includegraphics[width=0.5\linewidth]{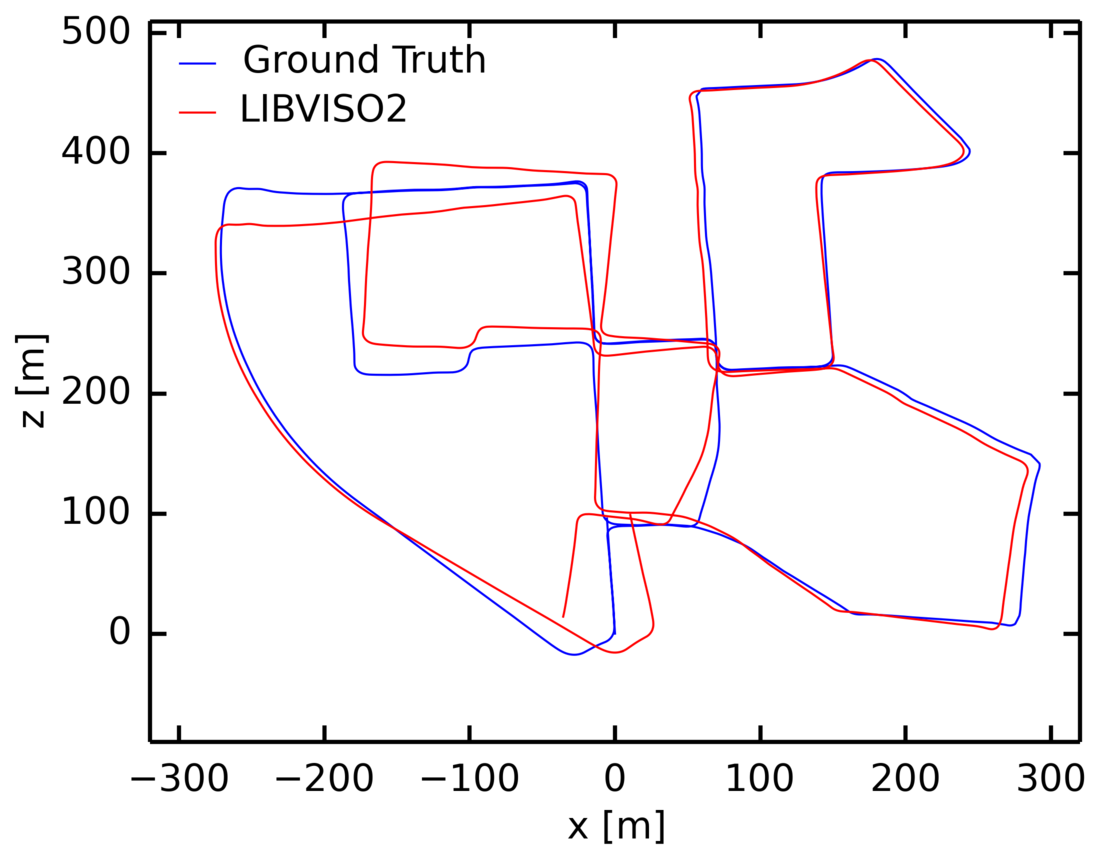}\\
    \includegraphics[width=0.5\linewidth]{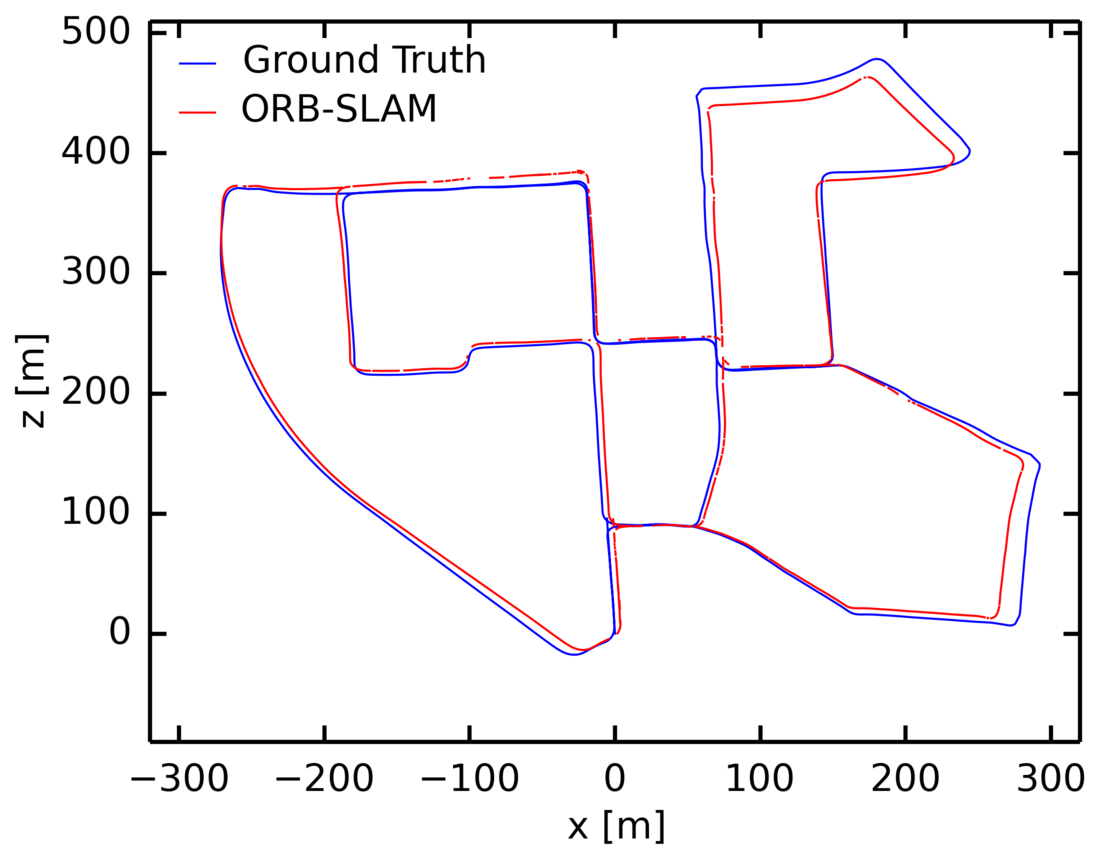}\hfill
    \includegraphics[width=0.5\linewidth]{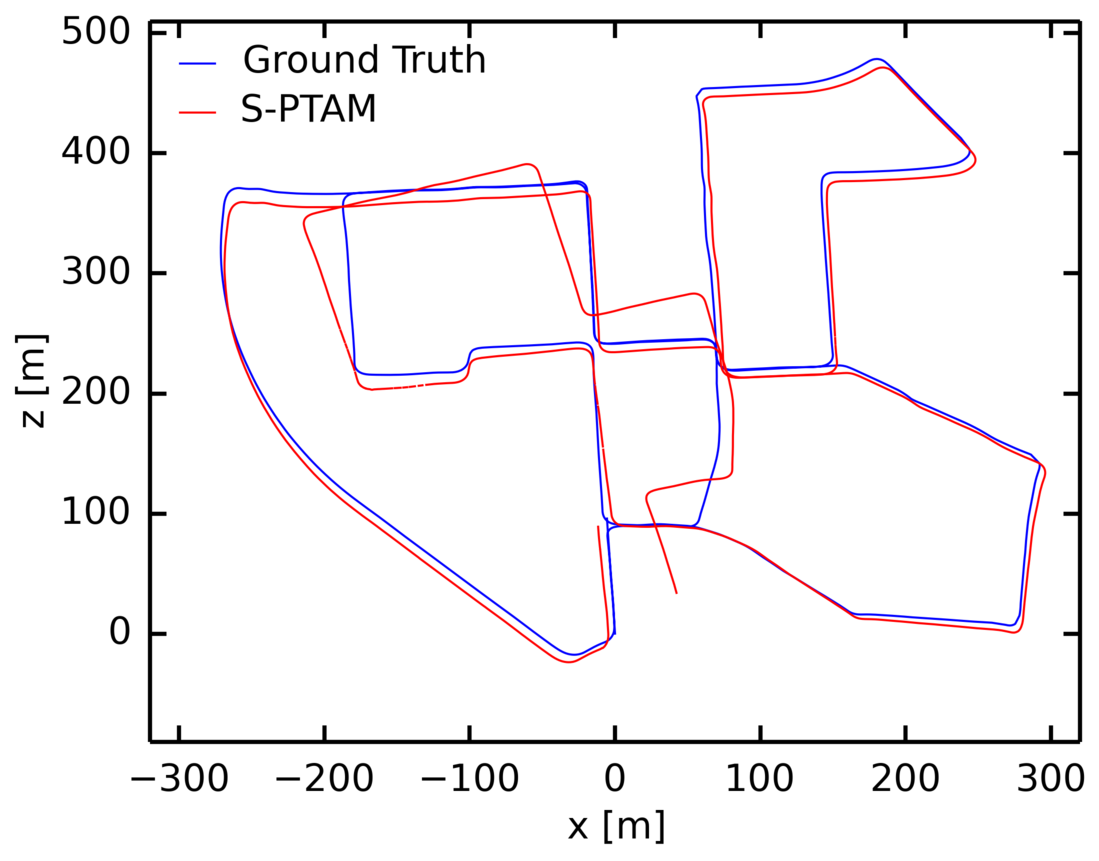}\\
    \caption[Results for KITTI Sequence 00]{Results for KITTI Sequence 00. Comparison of our method to LIBVISO2 (top row), ORB-SLAM and S-PTAM (bottom row). Our methods achieves the lowest ATE (\SI{5.79}{\m})}
    \label{fig:kitti00all}
\end{figure*}

Rectified images are provided with \SI{10}{\hertz} at a resolution of $1240\times376$ pixels. 
The sequences are recorded in real-world driving situations along urban, residual and countryside roads.
The distance traveled ranges from a few $100$ meters up to $5$ kilometers with driving speeds up to \SI[per-mode=symbol]{80}{\km\per\hour}.

The dataset is very challenging because the low frame rate in combination with fast driving speed leads to large inter-frame motions of up to \SI{2.8}{\m} per frame.
This strongly restricts the number of possible feature correspondences.
Moreover, moving obstacles in form of passing vehicles, bicycles, or pedestrians that have great impact on the performance of visual odometry systems, are included frequently.

We compare the performance of our semi-direct method with four state-of-the-art methods for visual odometry and SLAM.

We selected LIBVISO2 and LSD-SLAM for reference as our method is built upon them.
Moreover, we chose two established feature-based SLAM algorithms that presented promising results: ORB-SLAM as a monocular and S-PTAM as a stereo method.
All processing is done on the original image resolution of the rectified images of $1240 \times 376$.

Both error measures---RMSE and Median---for the training sequences $00$ to $10$ of the KITTI dataset are listed in \cref{tab:kitti_ate}. 

Unfortunately, LSD-SLAM fails on all sequences of the KITTI dataset. This is probably caused by too large inter-frame motion for a pure monocular direct method, as sufficient scene overlap is important for successful tracking.
Moreover, it can be seen, that all SLAM methods lack performance on sequence $01$, resulting in a a very high ATE.
Sequence $01$ contains images from driving on a highway, thus it is hard to find re-occurring feature points in subsequent frames.

\begin{figure*}[t]
    \centering
    \includegraphics[width=0.5\linewidth]{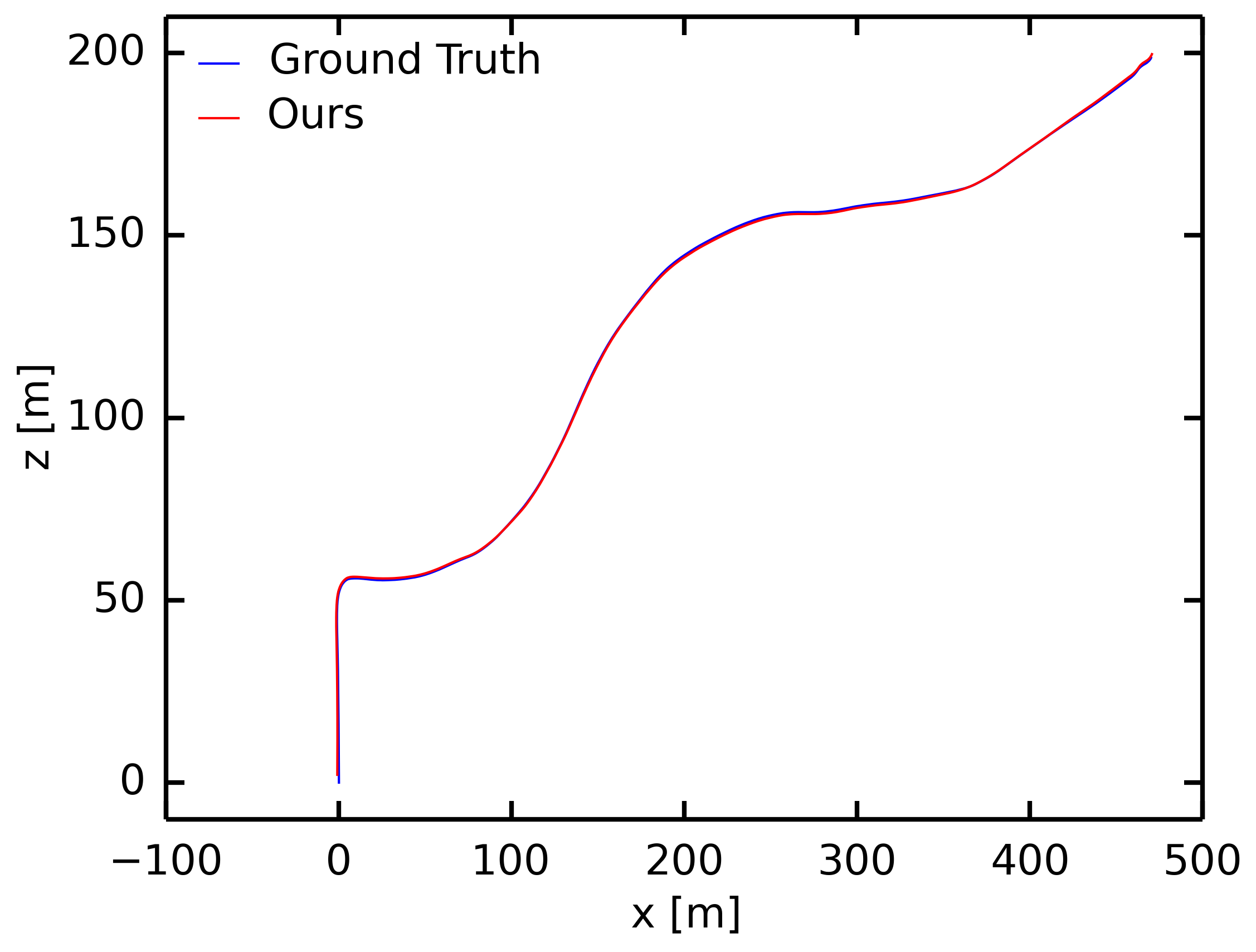}\hfill
    \includegraphics[width=0.5\linewidth]{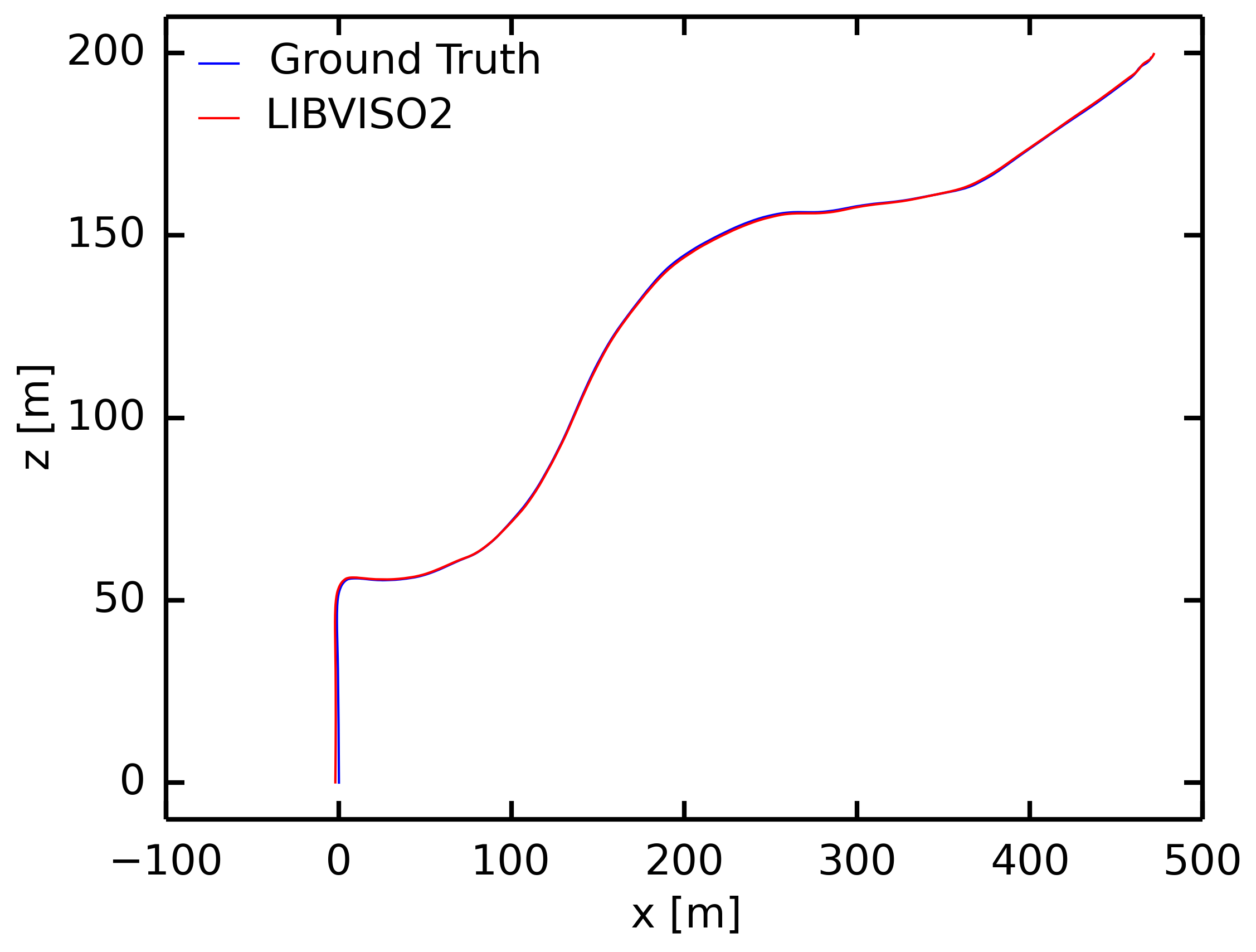}\hfill\\
    \includegraphics[width=0.5\linewidth]{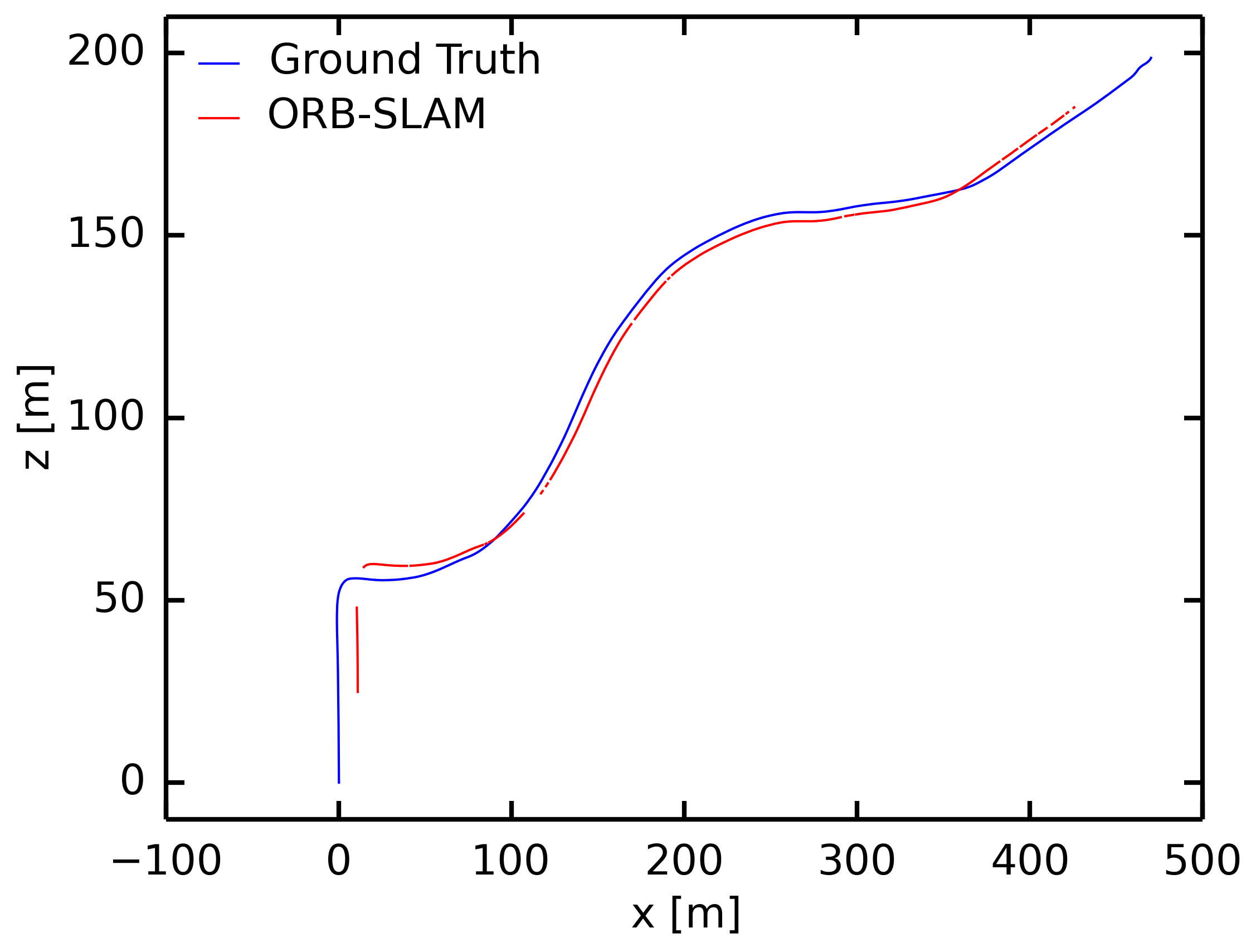}\hfill
    \includegraphics[width=0.5\linewidth]{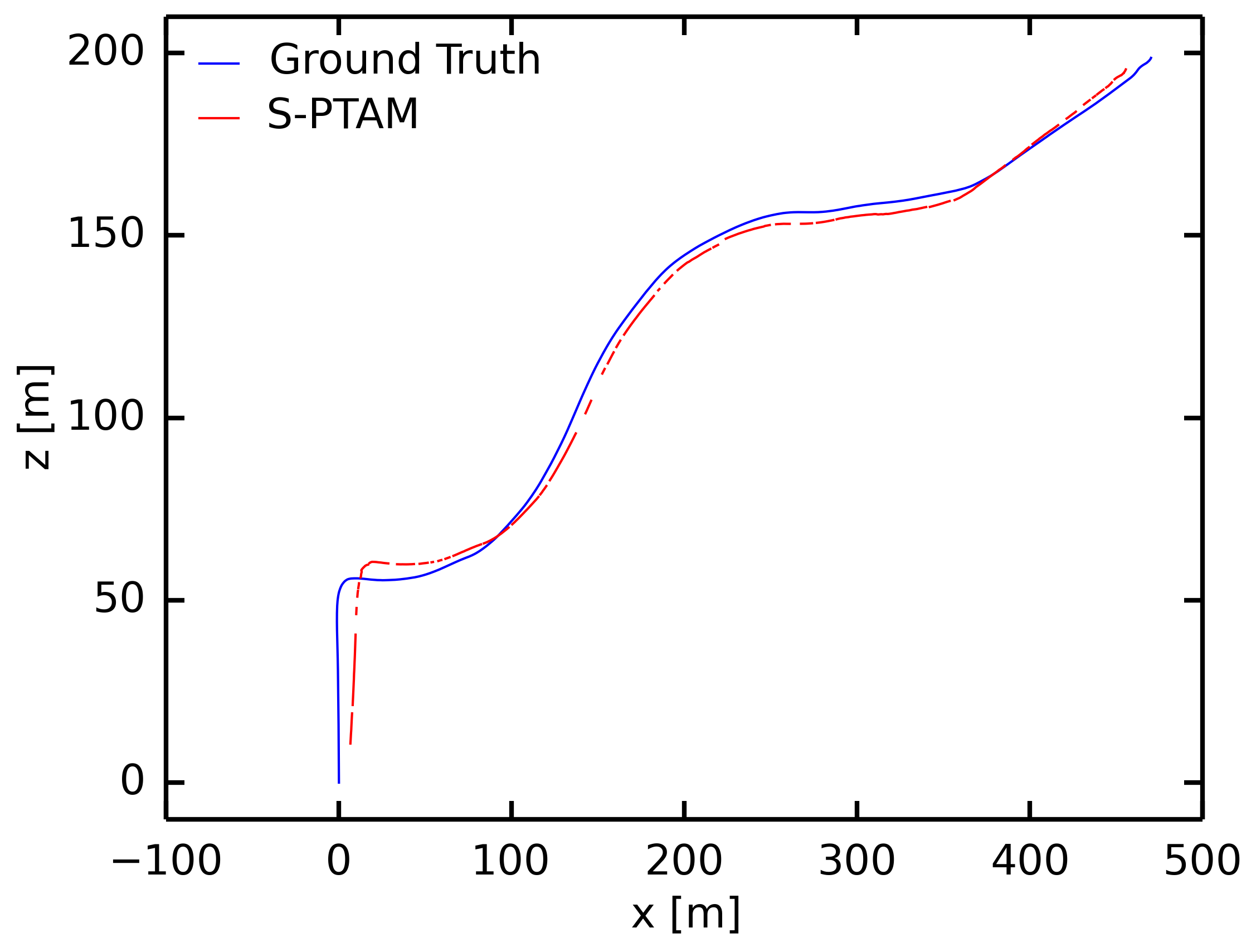}\hfill\\
    \caption[Results for KITTI Sequence 03]{Results for KITTI Sequence 03. Comparison of our method to LIBVISO2 (top row), ORB-SLAM and S-PTAM (bottom row). Our method and LIBVISO2 show accurate trajectories.}
    \label{fig:kitti03all}
\end{figure*}

Overall our method is equally good and in seven of eleven cases even better than state-of-the-art methods.
Especially sequences $03$ and $04$ show very accurate results below \SI{1}{\m} ATE. 
In three of the cases S-PTAM and in one case (sequence $02$) ORB-SLAM performs better.
As LIBVISO2 is a pure odometry method, it performs significantly worse than the SLAM methods on all datasets.

\begin{figure*}[t]
    \centering
    \includegraphics[width=0.5\linewidth]{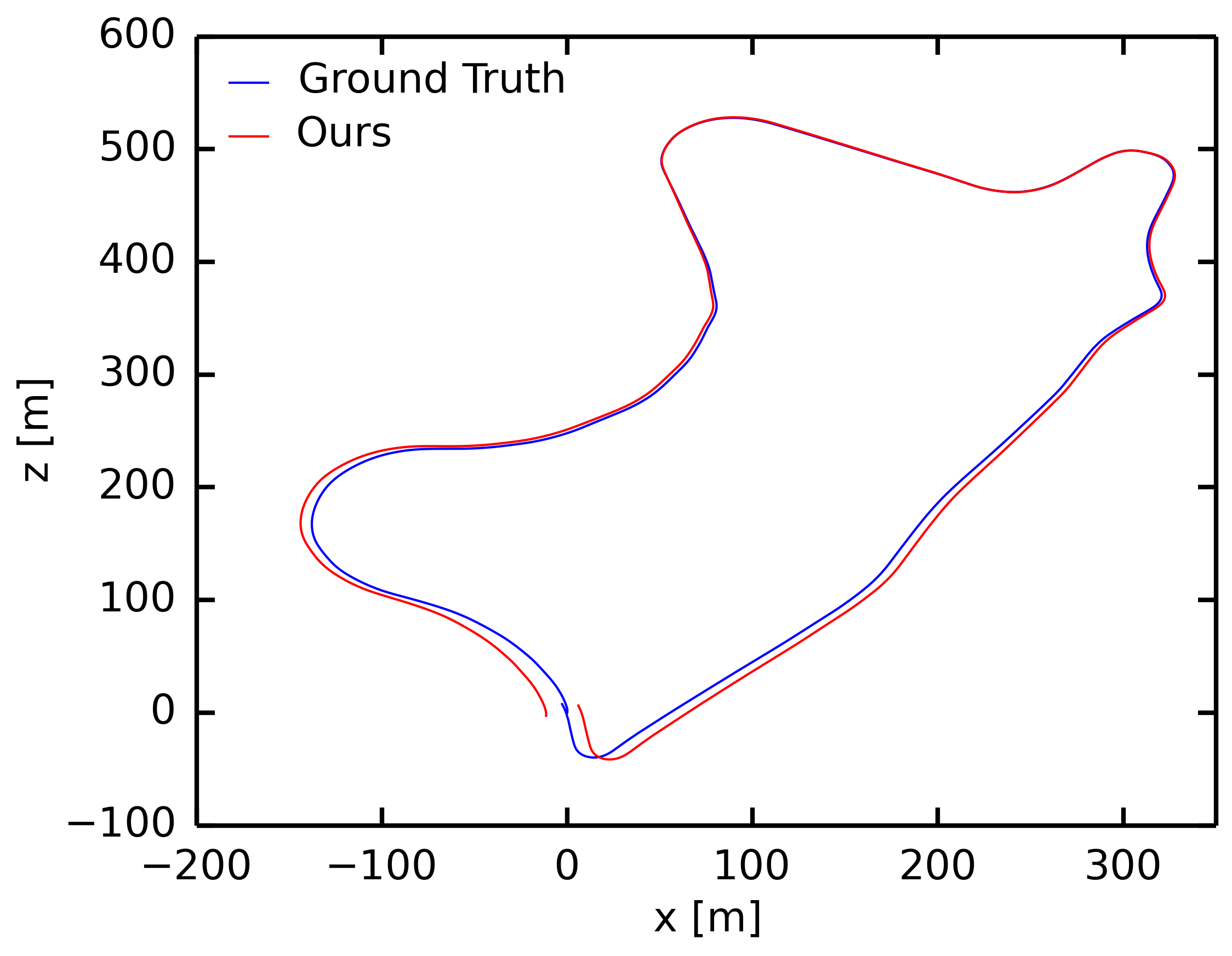}\hfill
    \includegraphics[width=0.5\linewidth]{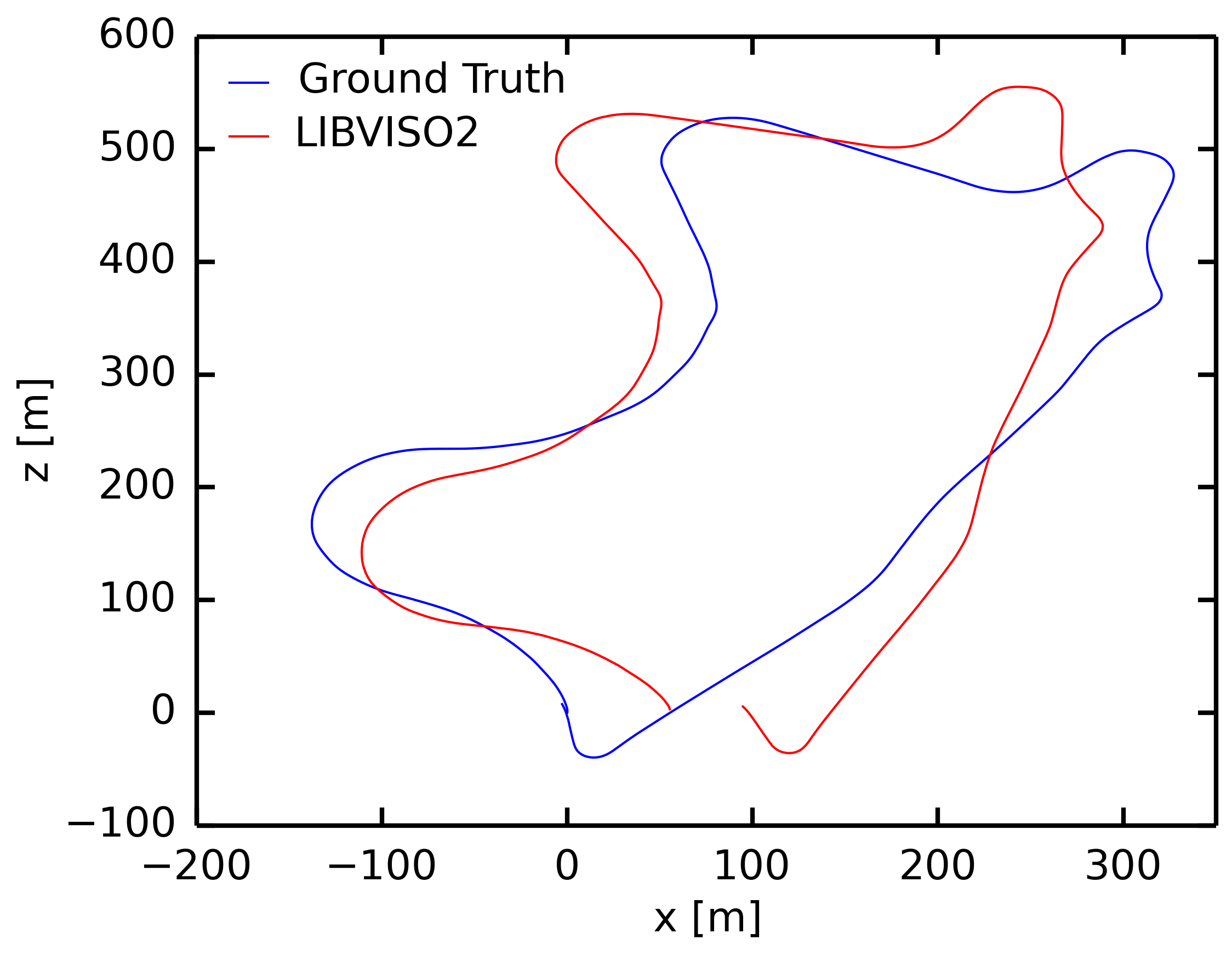}\hfill\\
    \includegraphics[width=0.5\linewidth]{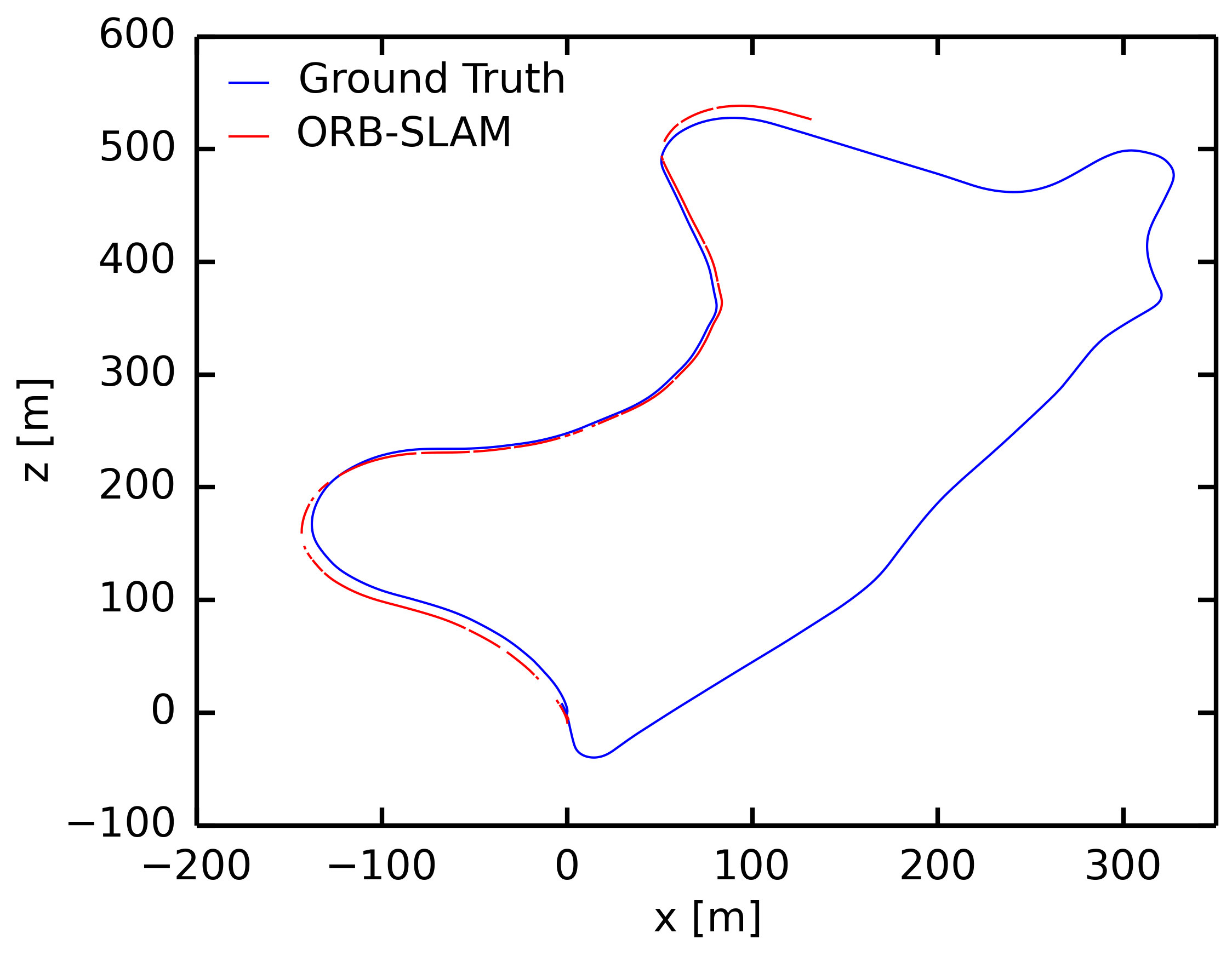}\hfill
    \includegraphics[width=0.5\linewidth]{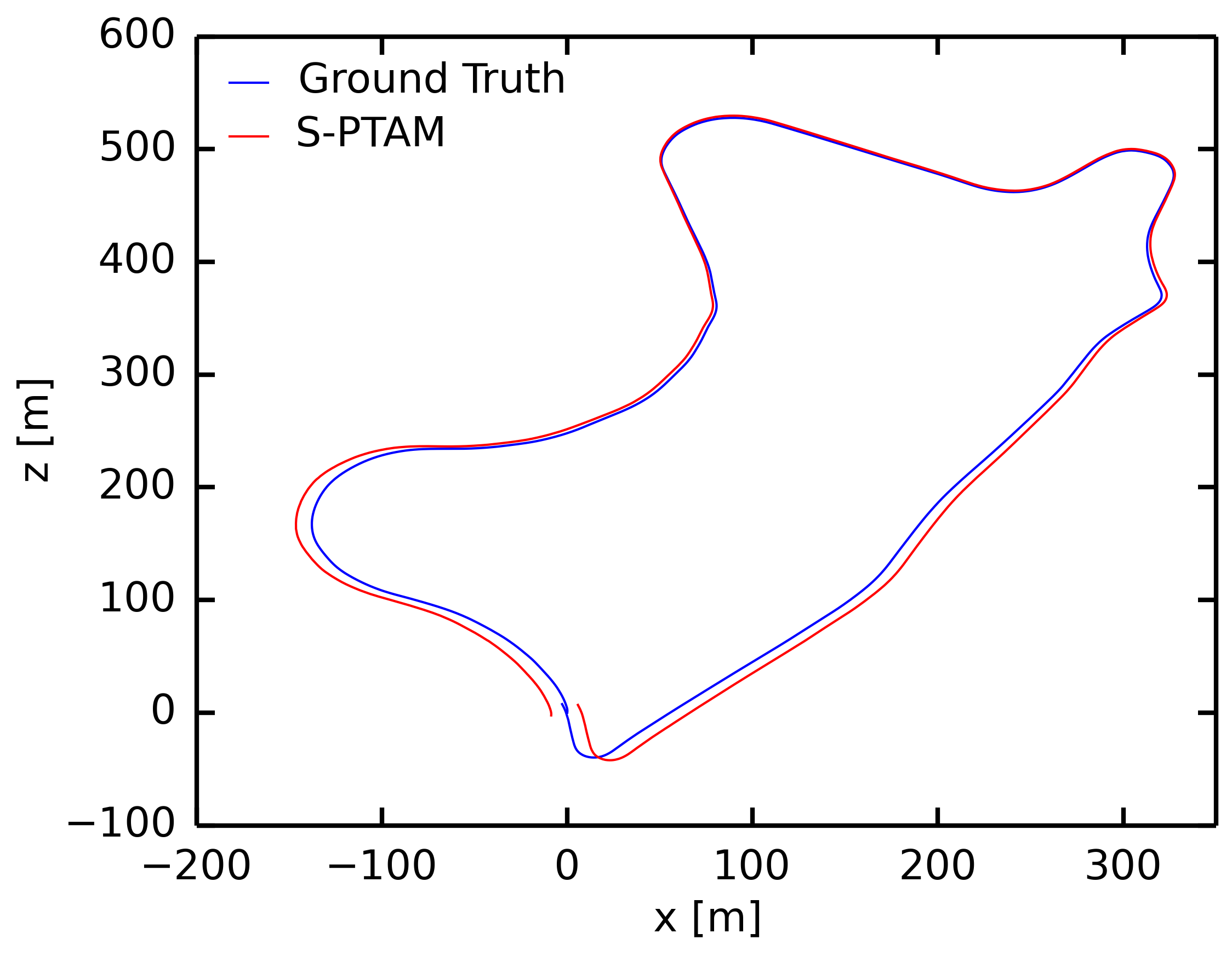}\hfill\\
    \caption[Results for KITTI Sequence 09]{Results for KITTI Sequence 09. Comparison of our method to LIBVISO2 (top row), ORB-SLAM and S-PTAM (bottom row). While ORB-SLAM looses track and LIBVISO2 accumulates drift, S-PTAM and our method stay close to the ground truth.}
    \label{fig:kitti09all}
\end{figure*}

When averaging over the eleven training sequences our method ranks first, followed by S-PTAM, ORB-SLAM and LIBVISO2.
However, the bad results from sequence $01$ greatly affect the final average computation as all methods accumulate high ATEs in sequence $01$.
One could argue that such high ATE values count as outlier or failure.
Therefore, we also show resulting means when omitting sequence $01$ for all methods.
It follows that these results show significantly lower ATEs.
When omitting sequence $01$ our method achieves a mean (median) ATE of \SI{5.15}{\m} (\SI{3.99}{\m}) compared to the S-PTAM result of \SI{7.85}{\m} (\SI{6.57}{\m}).

For a better visualization exemplary trajectories are shown in birds-eye perspective for the sequences $00, 03, 09$ and $10$. 

Sequence 00 is shown in \cref{fig:kitti00all}.
It can be seen, that our approach performs best, followed by ORB-SLAM, S-PTAM and LIBVISO2.
Moreover, limitations of the approaches become visible:
as LIBVISO2 is a pure odometry method, it accumulates more drift over time, and as ORB-SLAM is a monocular method, scale is not always estimated correctly. Rotations are challenging for all methods. 
In this sequence S-PTAM fails to track rotations frequently and exhibits drift for the last part of the trajectory.

\begin{figure*}[t]
    \centering
    \includegraphics[width=0.5\linewidth]{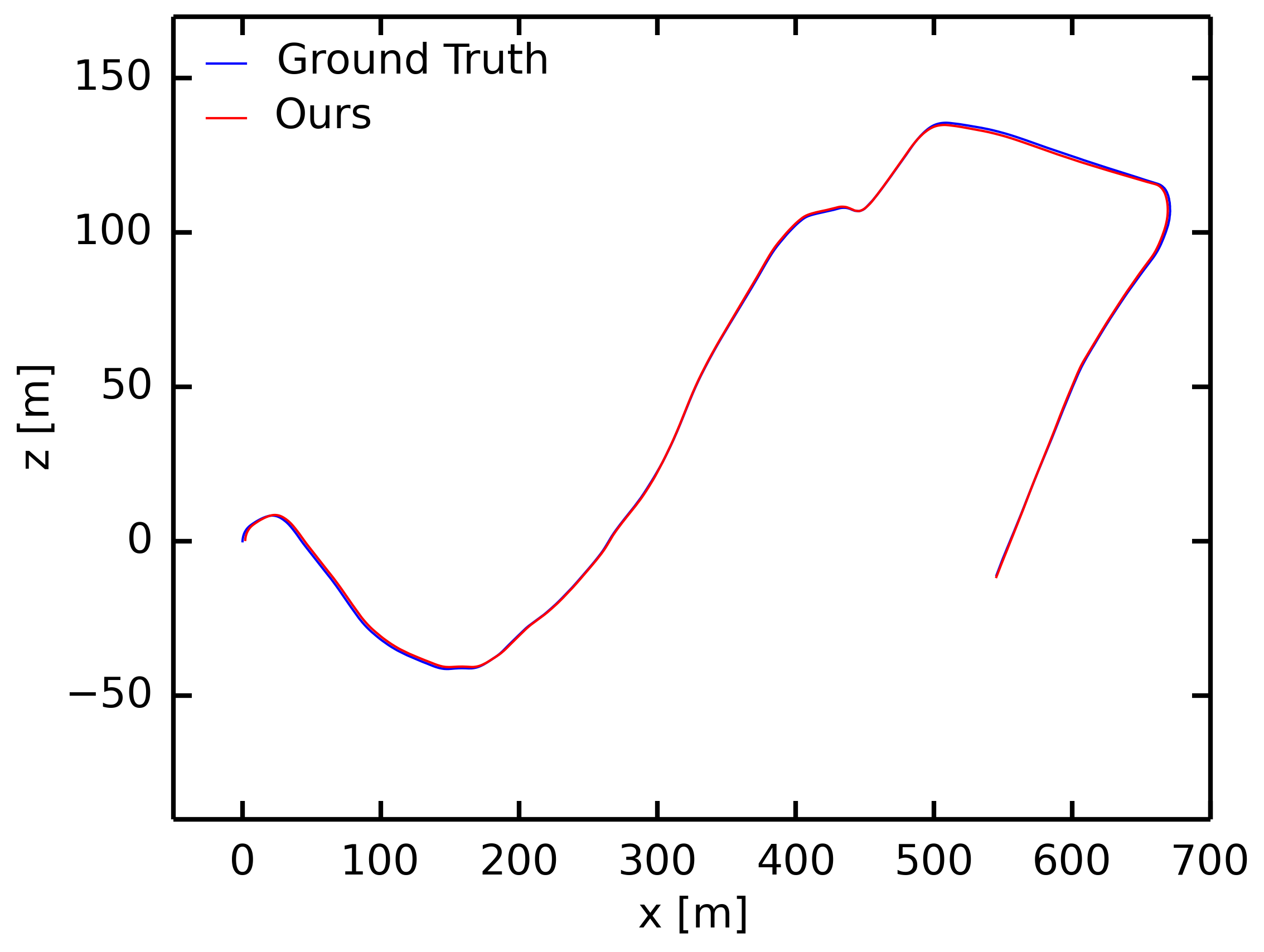}\hfill
    \includegraphics[width=0.5\linewidth]{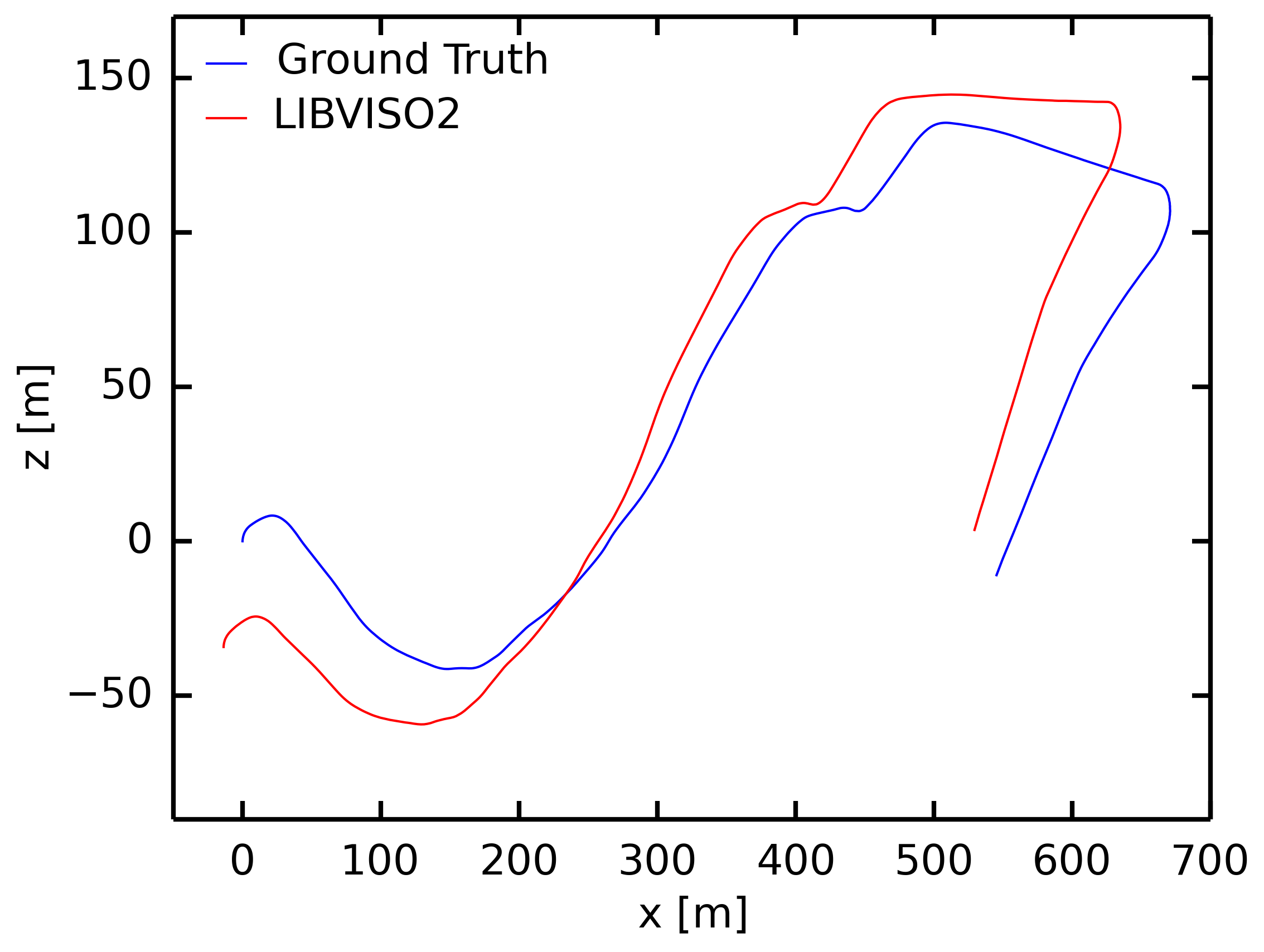}\hfill\\
    \includegraphics[width=0.5\linewidth]{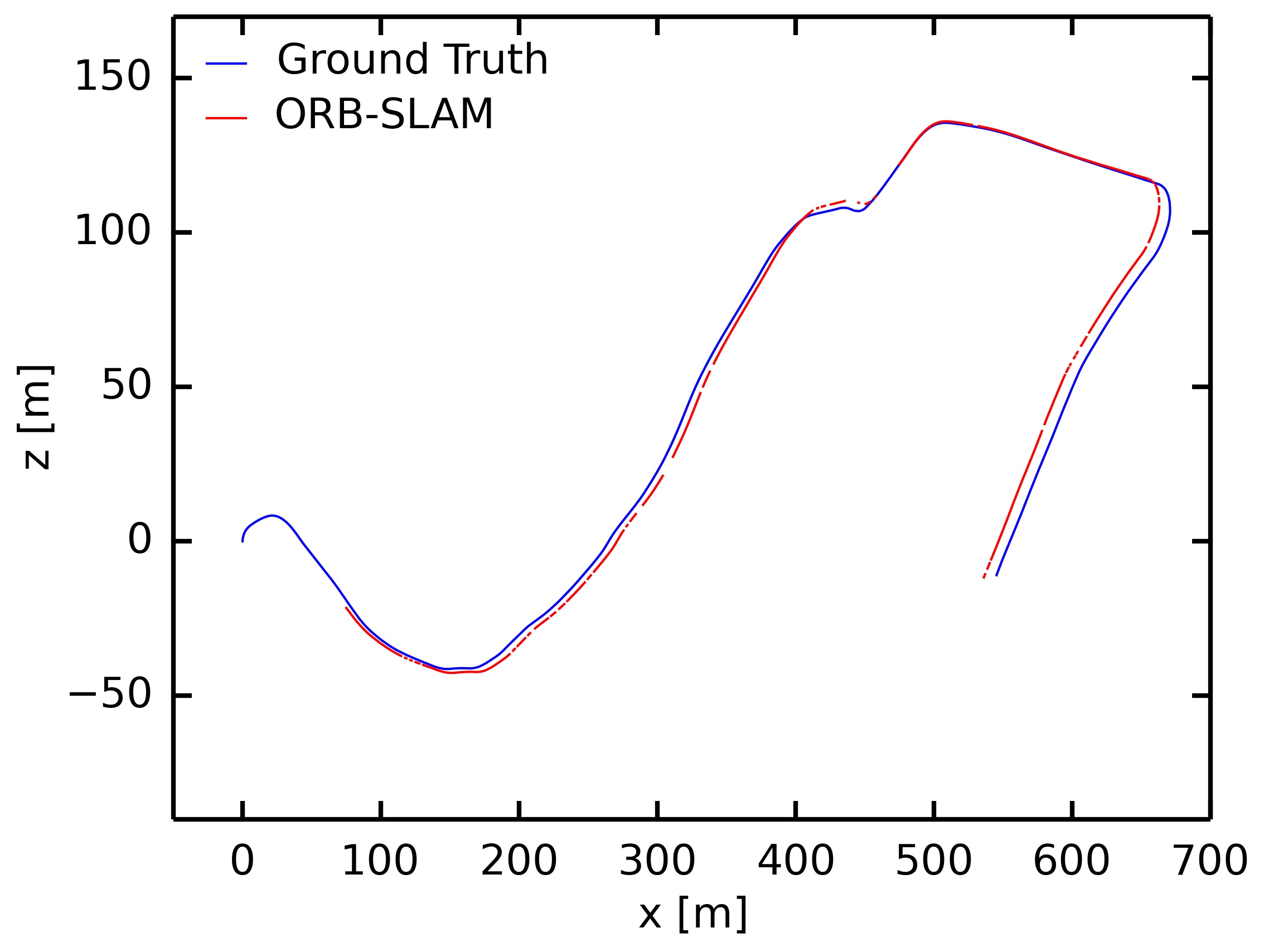}\hfill
    \includegraphics[width=0.5\linewidth]{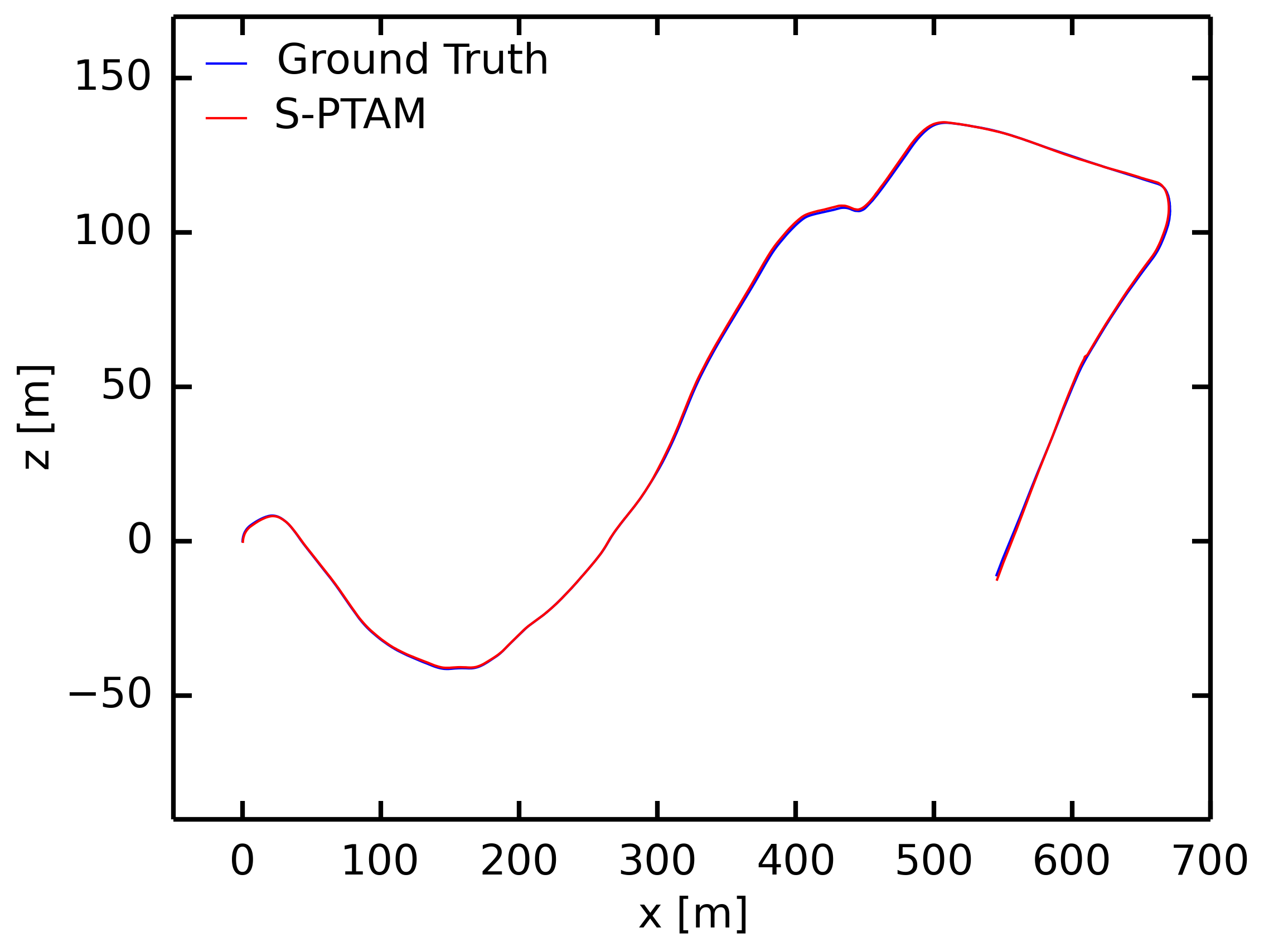}\hfill\\
    \caption[Results for KITTI Sequence 10]{Results for KITTI Sequence 10. Comparison of our method to LIBVISO2 (top row), ORB-SLAM and S-PTAM (bottom row) on a longer trajectory without loop closures.}
    \label{fig:kitti10all}
\end{figure*}

\cref{fig:kitti03all} shows the results for sequence $03$, a trajectory without full loop closure. We choose this sequence to compare the drift over time, when no full loop can be closed.
All estimated trajectories are close to the ground truth.
However, our method is---with \SI{0.63}{\m} ATE---distinctively more accurate than ORB-SLAM (\SI{11.91}{\m}) and S-PTAM (\SI{10.53}{\m}).
Additionally, LIBVISO2 also shows accurate results with an ATE of \SI{1.67}{\m} and does not accumulate much drift for this trajectory.

In sequence $09$ a full loop closure appears at the very end of the trajectory, that is not always detected from the SLAM methods before the sequence ends. 
This behavior is shown in \cref{fig:kitti09all}.
Again, LIBVISO2 suffers from drift over time while the results from our Semi-Direct SLAM (\SI{5.46}{\m}) are more accurate than the results from S-PTAM (\SI{7.27}{\m}).
Moreover, it can be seen, that ORB-SLAM lost track at some point and failed to relocalize. Thus, more than half of the trajectory remains uncovered.
This is not reflected in the error measure, because the ATE is only computed over existing measurements.

Sequence $10$ is similar to sequence $03$ as it contains no full loop but covers a longer path and performs more rotations.
Results for this sequence are visualized in \cref{fig:kitti10all}.
They show that our method performs well even if the path of LIBVISO2 drifts over time.
ORB-SLAM fails to initialize directly from the beginning but retrieves a trajectory consistent with the ground truth later on, even with little offset.
S-PTAM again shows similar results to Semi-Direct SLAM although Semi-Direct SLAM performs slightly better (\SI{1.68}{\m} to \SI{2.08}{\m} respectively).

However, as LSD-SLAM fails on the KITTI sequences, we compare our semi-direct approach to its fully direct version without feature-based initial estimates.	
In particular, we compare the combined semi-direct approach to its building blocks---LIBVISO2 and direct stereo tracking---separately.
As LIBVISO2 is a pure odometry method, we evaluate it against results from our semi-direct odometry without closing loops.

\begin{figure*}
    \centering
    \includegraphics[width=0.45\linewidth]{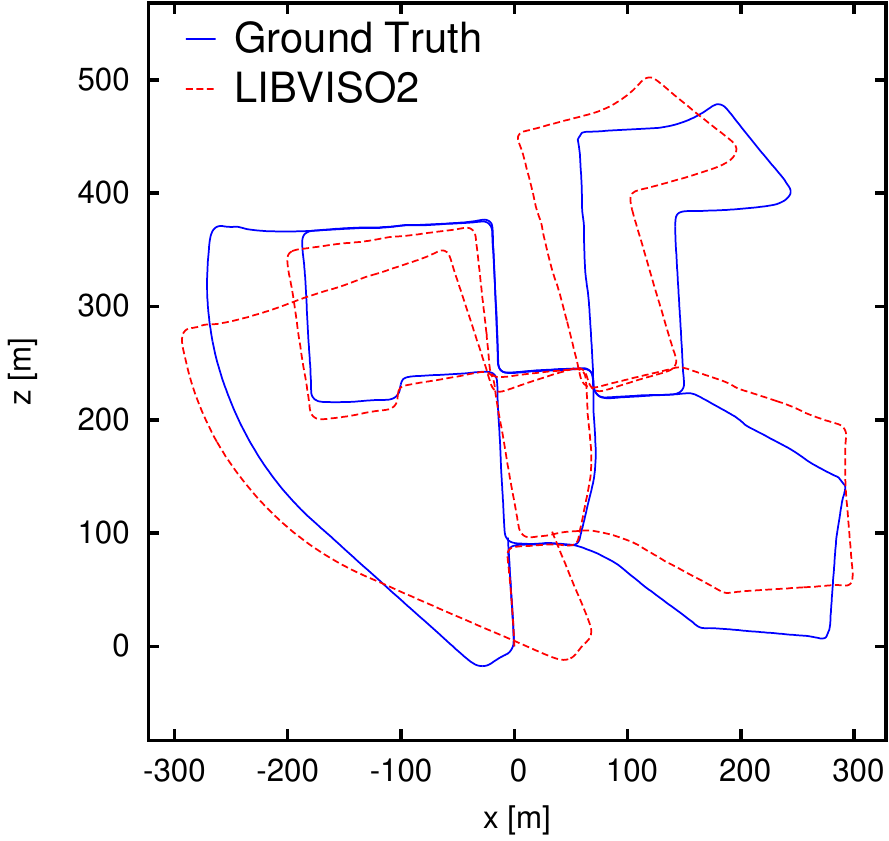}\hfill
    \includegraphics[width=0.45\linewidth]{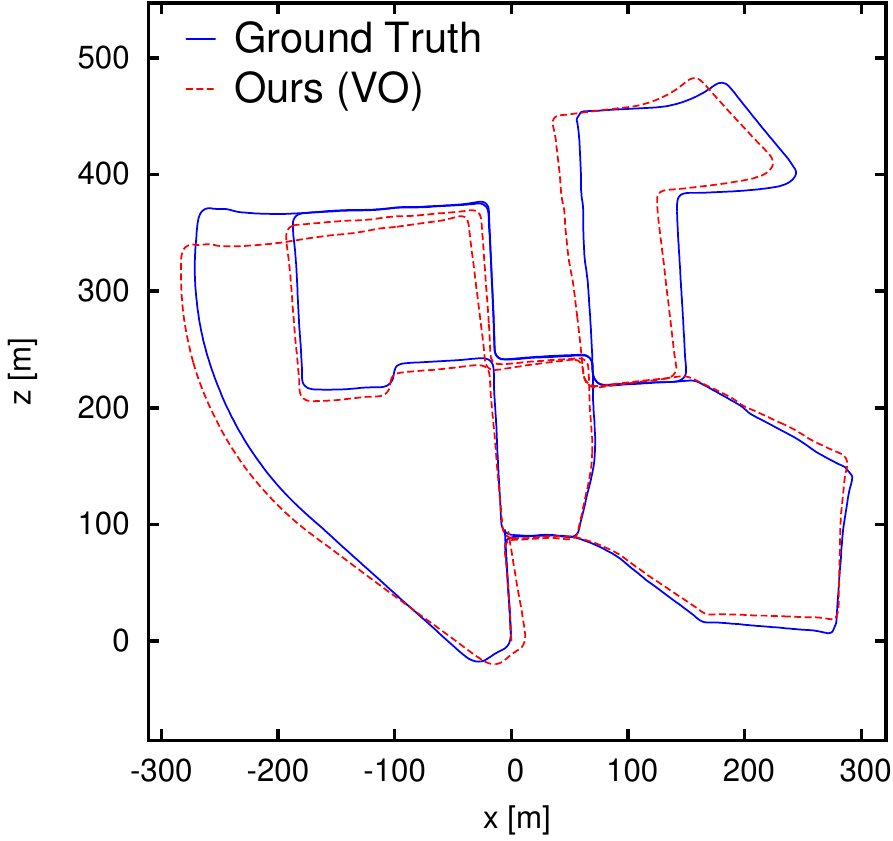}\hfill\\
    \includegraphics[width=0.45\linewidth]{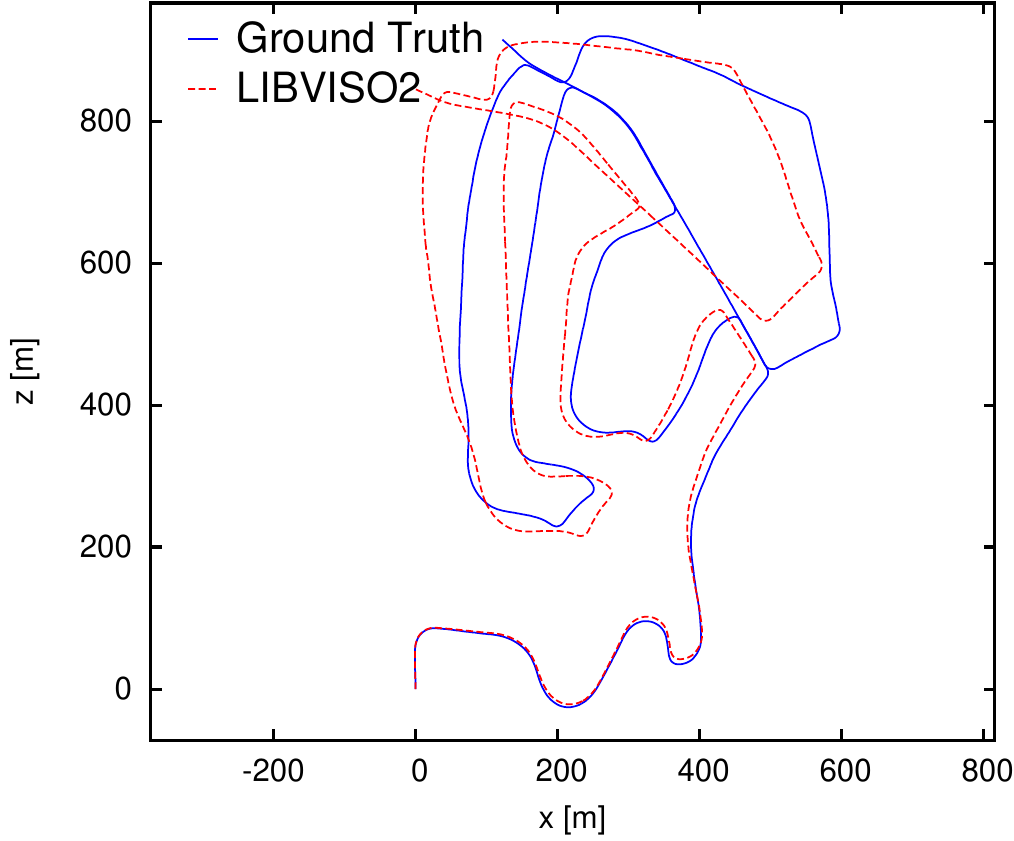}\hfill
    \includegraphics[width=0.45\linewidth]{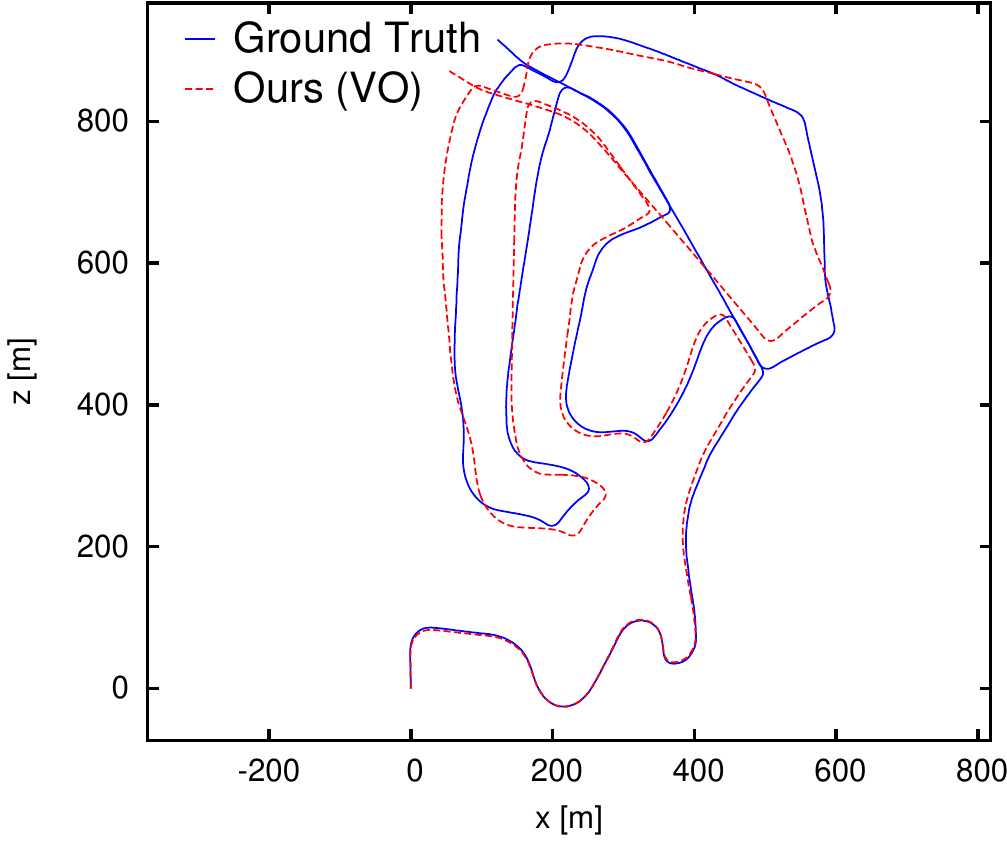}\hfill\\
    \includegraphics[width=0.45\linewidth]{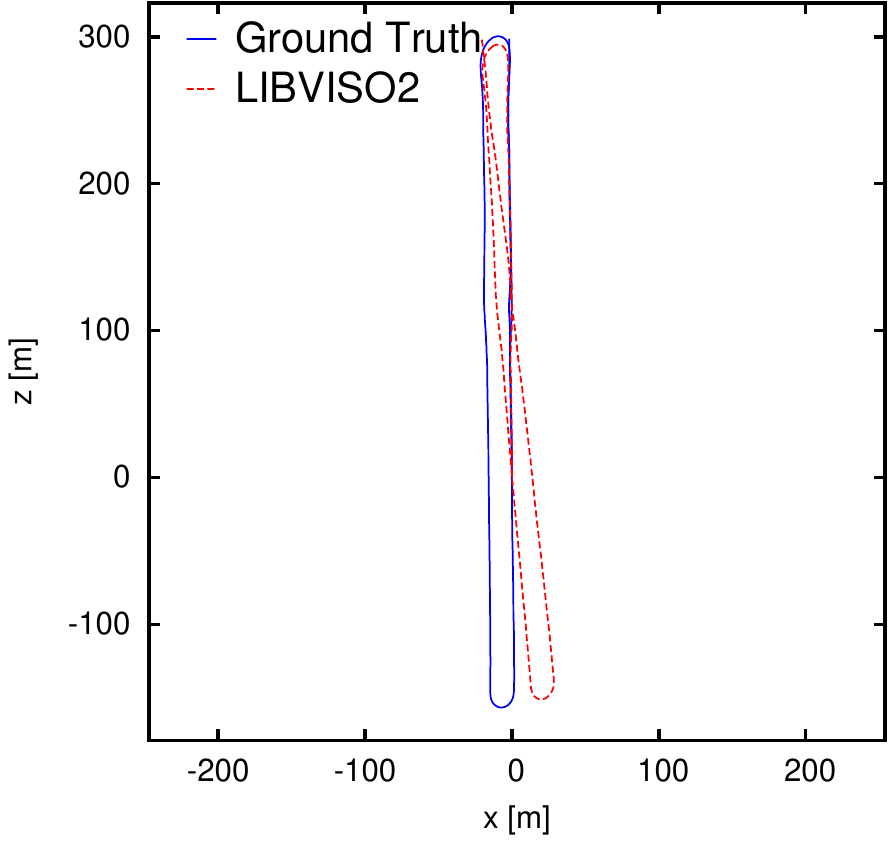}\hfill
    \includegraphics[width=0.45\linewidth]{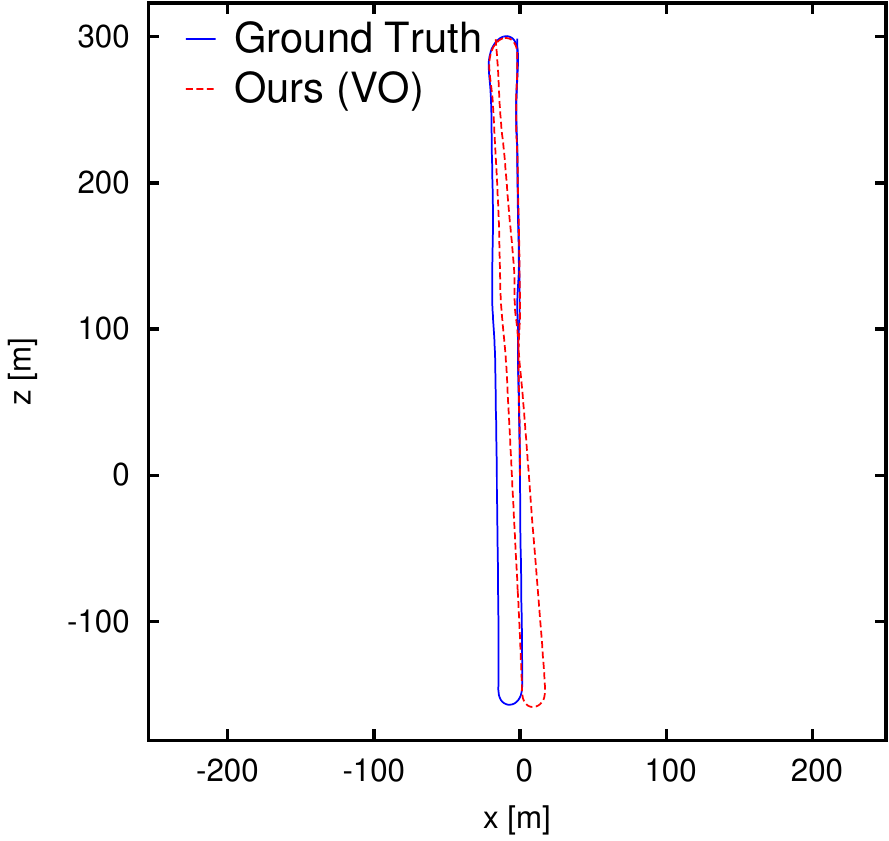}\hfill\\
    \caption[Comparison of LIBVISO2 and our Odometry on KITTI]{Comparison of the results from LIBVISO2 (left) to our semi-direct odometry (right). Top Row: KITTI Sequence 00, Middle Row: KITTI Sequence 02, Bottom Row: KITTI Sequence 06. In direct comparison to LIBVISO2 our method accumulates less drift.}
    \label{fig:kittisemilib}
\end{figure*}

\begin{figure}
    \centering
    \includegraphics[width=0.45\linewidth]{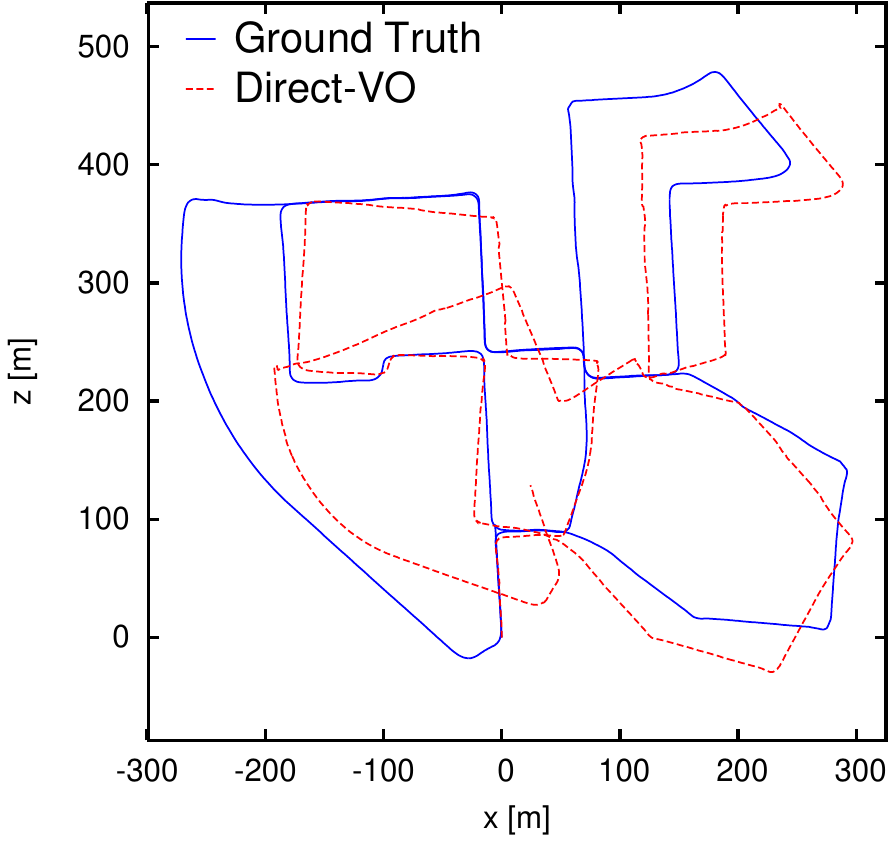}\hfill
    \includegraphics[width=0.45\linewidth]{img_plots_semidirect_00.pdf}\\
    \includegraphics[width=0.45\linewidth]{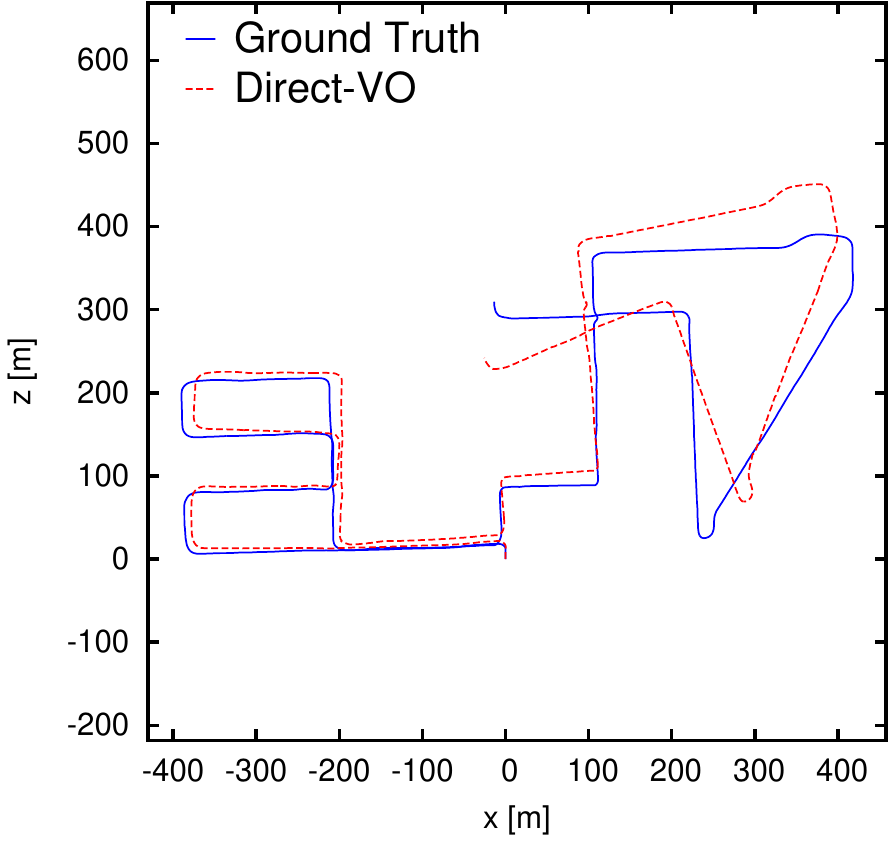}\hfill
    \includegraphics[width=0.5\linewidth]{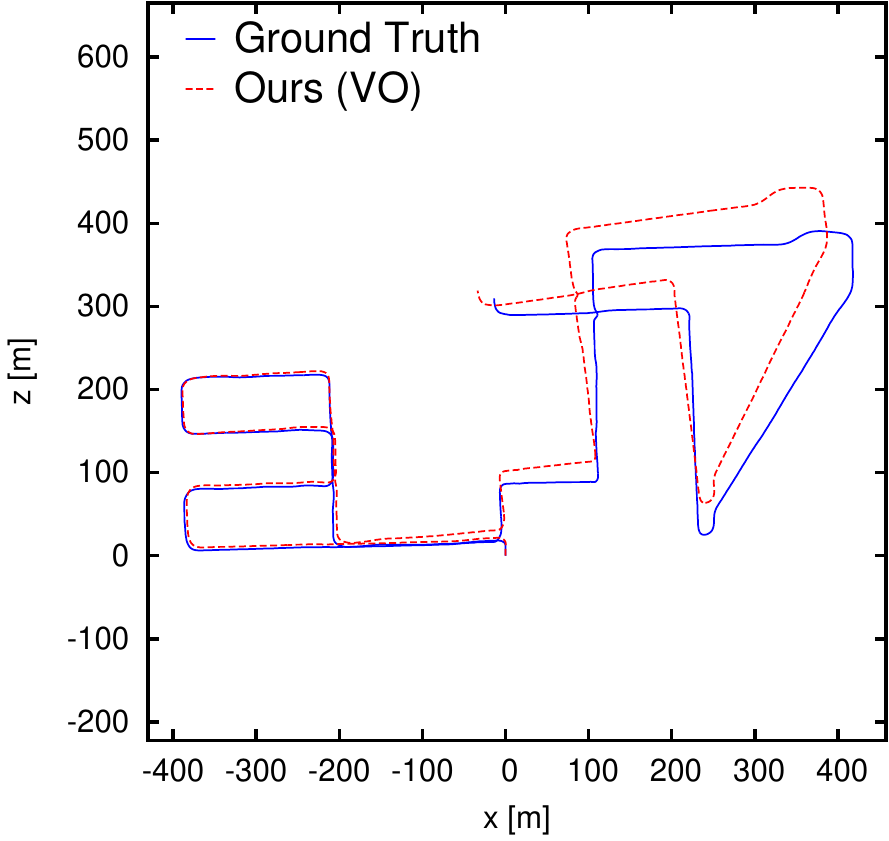}\\
    \caption[Comparison of Direct Odometry and our Odometry on KITTI]{Comparison of the results from direct odometry (left) to our semi-direct odometry (right). Top Row: KITTI Sequence 00, Bottom Row: KITTI Sequence 08. In direct comparison to the direct visual odometry our method is clearly more robust to fast rotations and to large motions.}
    \label{fig:kittisemidir}
\end{figure}

In \cref{fig:kittisemilib} the results from three different datasets ($00, 02$ and $06$) are shown in birds-eye perspective. 
The left column shows the resulting path LIBVISO2 computed and the right column the path from the semi-direct odometry.
It can be seen that LIBVISO2 accumulates more drift over time than the semi-direct approach, while at the same time the semi-direct approach remains closer to the ground truth trajectory.

When comparing our semi-direct approach to its fully direct version without feature-based odometry as initial estimate, we noticed that a fully direct version has problems with strong turns in the dataset.
Moreover, the dataset is very challenging for a fully direct method as it contains large inter-frame motions and difficult lighting changes.
Large inter-frame motions are problematic for direct methods because they assume small pixel displacements~\citep{Directmethods}. Without a good initial estimate they often fail to retrieve large displacements.
Difficult lighting changes, induced by auto-exposure and changing sunlight, are tough, as they violate the brightness constancy assumption.
Thereby, it can be seen in \cref{fig:kittisemidir} that the fully direct odometry accumulates more drift over time than our semi-direct version.
Again, while the semi-direct approach is shown in the right column, the fully direct approach is visualized in the left column for sequences $00$ and $06$.
Fully direct tracking tends to fail especially at strong turns and at street crossings where lighting changes increase because the car leaves shadowed street canyons.
In contrast, our approach is more robust to strong rotations and illumination changes.

The semi-direct method performs better than its isolated building blocks. The direct tracking is in principle more accurate but has problems with large motions.
However, when a good initial estimate is available, as in our case from LIBVISO2, direct tracking succeeds even at large motions and with a low frame rate.

Generally speaking, a combined semi-direct odometry performs better than both---feature-based and direct---odometries alone. 
Overall, our approach shows promising results on the KITTI dataset when compared to other state-of-the-art methods.

\subsection{EuRoC}
In addition to the evaluation on the KITTI dataset, we perform further experiments on the well-known visual-inertial EuRoC MAV dataset that contains stereo images and synchronized IMU readings from the on-board computer of an Asctec Firefly hex-rotor helicopter. We choose six trajectories with different difficulties from the two Vicon datasets V$0$ and V$1$.
The data has been collected from flights in a room that is equipped with a Vicon motion capture system offering 6D ground truth poses.
	
The MAV carries a visual-inertial sensor~\citep{NikolicRBGLFS14} that captures stereo images of WVGA resolution at \SI{20}{\hertz} and synchronized IMU measurements at \SI{200}{\hertz}.
		
\begin{table}
\small
\begin{center}
\begin{tabular}{c||c||c||c||c||c}
EuRoC & \multicolumn{5}{c} {\textbf{Absolute Trajectory Error RMSE (Median) in m}}\\
Dataset & Ours & Libviso2 & \hspace{-0.1em}LSD-SLAM\hspace{-0.1em} & ORB-SLAM & S-PTAM\\
\hline
V1\_01 & \bftab{0.12 (0.11)} & 0.31 (0.31) & 0.19 (0.10) & 0.79 (0.62) & 0.28 (0.19) \\
V1\_02 & \bftab{0.11 (0.10)} & 0.29 (0.27) & 0.98 (0.92) & 0.98 (0.87) & 0.50 (0.35) \\
V1\_03 & \bftab{0.75 (0.45)} & 0.87 (0.64) & X           & 2.12 (1.38) & 1.36 (1.09) \\
V2\_01 & \bftab{0.18 (0.12)} & 0.40 (0.31) & 0.45 (0.41) & 0.50 (0.42) & 2.38 (1.78) \\
V2\_02 & \bftab{0.27 (0.22)} & 1.29 (1.08) & 0.51 (0.48) & 1.76 (1.39) & 4.58 (4.18) \\
V2\_03 & \bftab{0.87 (0.66)} & 1.99 (1.66) & X & X               & X \\
\hline
mean   & \bftab{0.38 (0.28)}   & 0.85 (0.71) & 0.53 (0.48) & 1.23 (0.94) & 1.82 (1.52) \\
\hline
\end{tabular}
\end{center}
\caption{ATE Results on EuRoC Dataset}
\label{tab:euroc_ate}
\end{table}
		
Both datasets contain three trajectories with increasing difficulty named as: easy ($\_01$), medium ($\_02$) and difficult ($\_03$).
The easy trajectories have good illumination, are feature rich, and show no motion blur and only low optical flow and low varying scene depth. They capture a static scene.
The difficulty increases in the medium trajectories by adding difficult lighting conditions, high optical flow and medium varying scene depth. However, they still show a static scene and a feature rich environment without motion blur.
In contrast, the difficult scenes contain areas with only few visual features and more repetitive structures. Moreover, they add motion blur and highly unstable lighting conditions.
The MAV performs very aggressive flight maneuvers resulting in high optical flow and highly varying scene depth in a non-static scene.

\begin{figure*}[t!]
    \centering
    \includegraphics[width=0.5\linewidth]{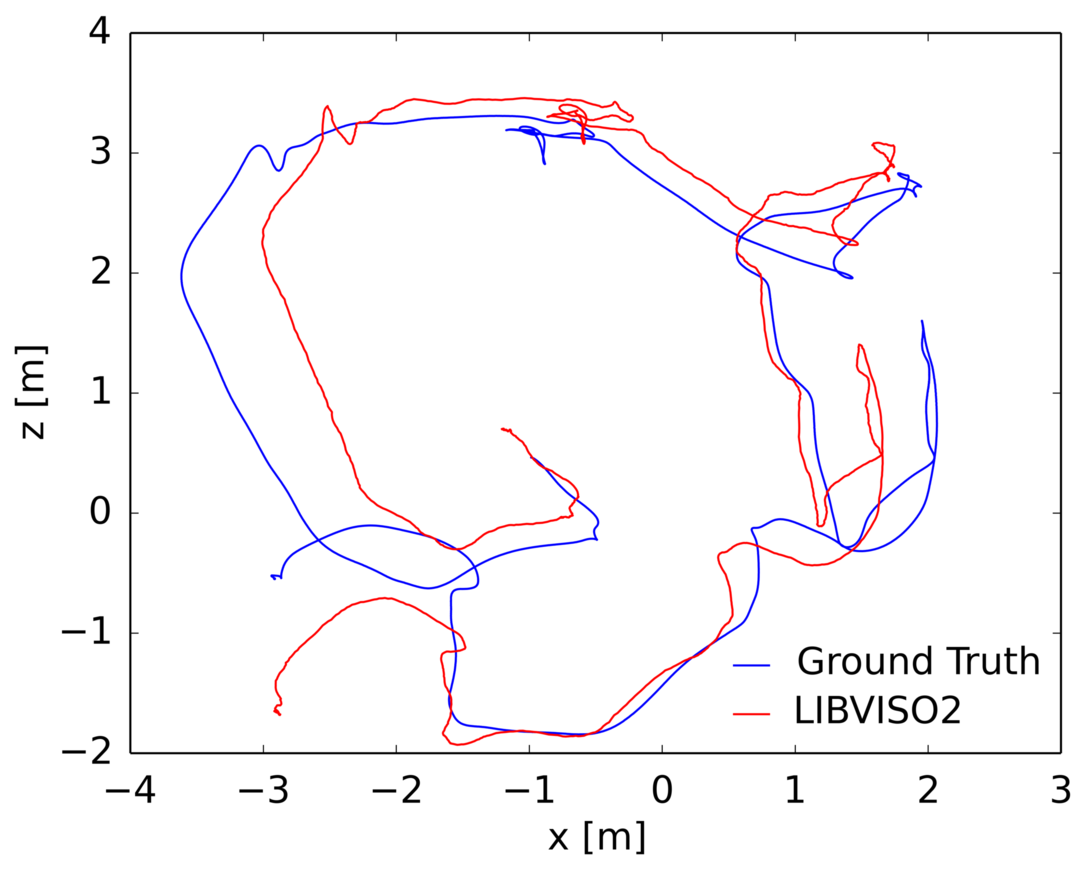}\hfill
    \includegraphics[width=0.5\linewidth]{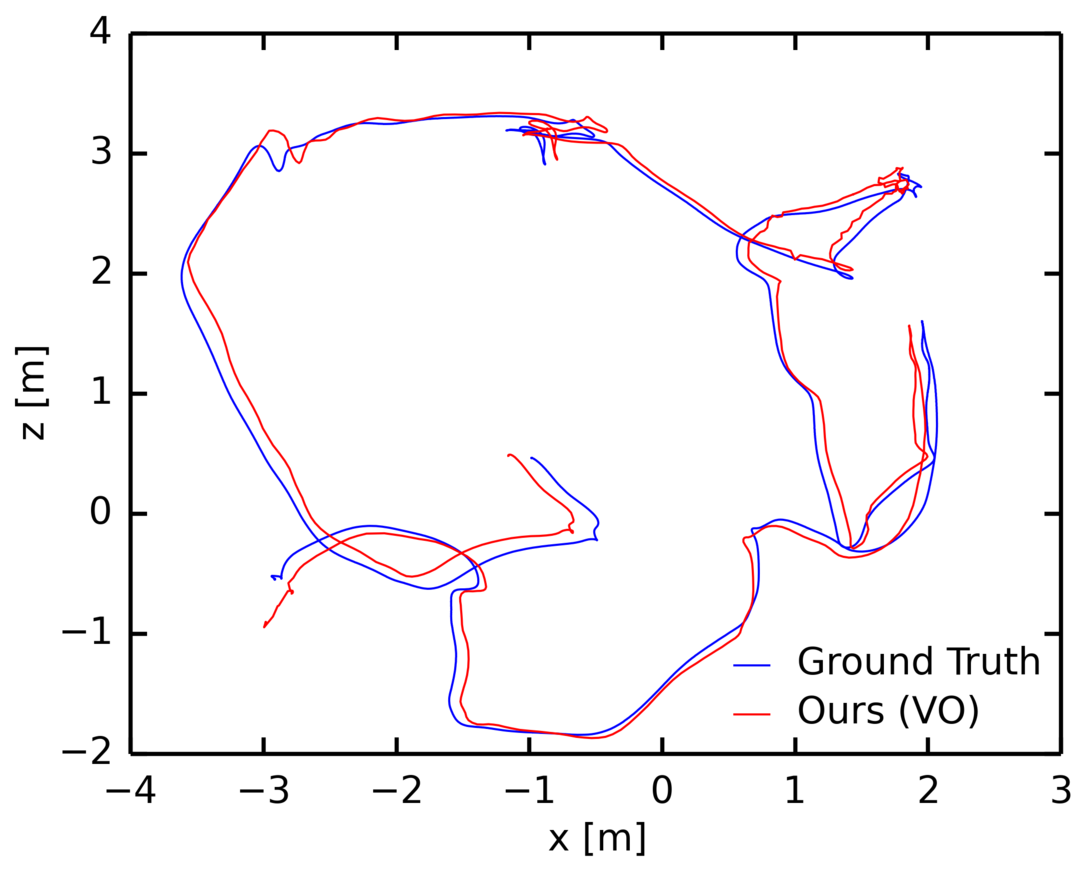}\\
    \includegraphics[width=0.5\linewidth]{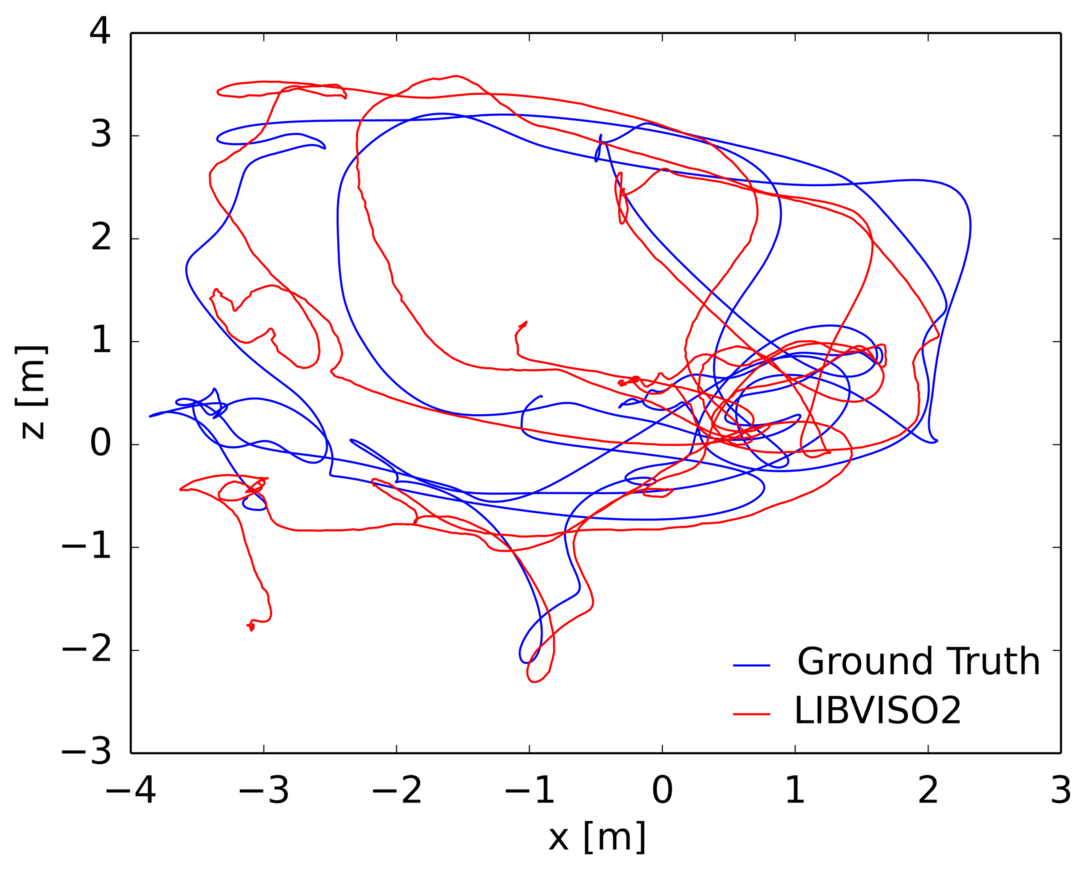}\hfill
    \includegraphics[width=0.5\linewidth]{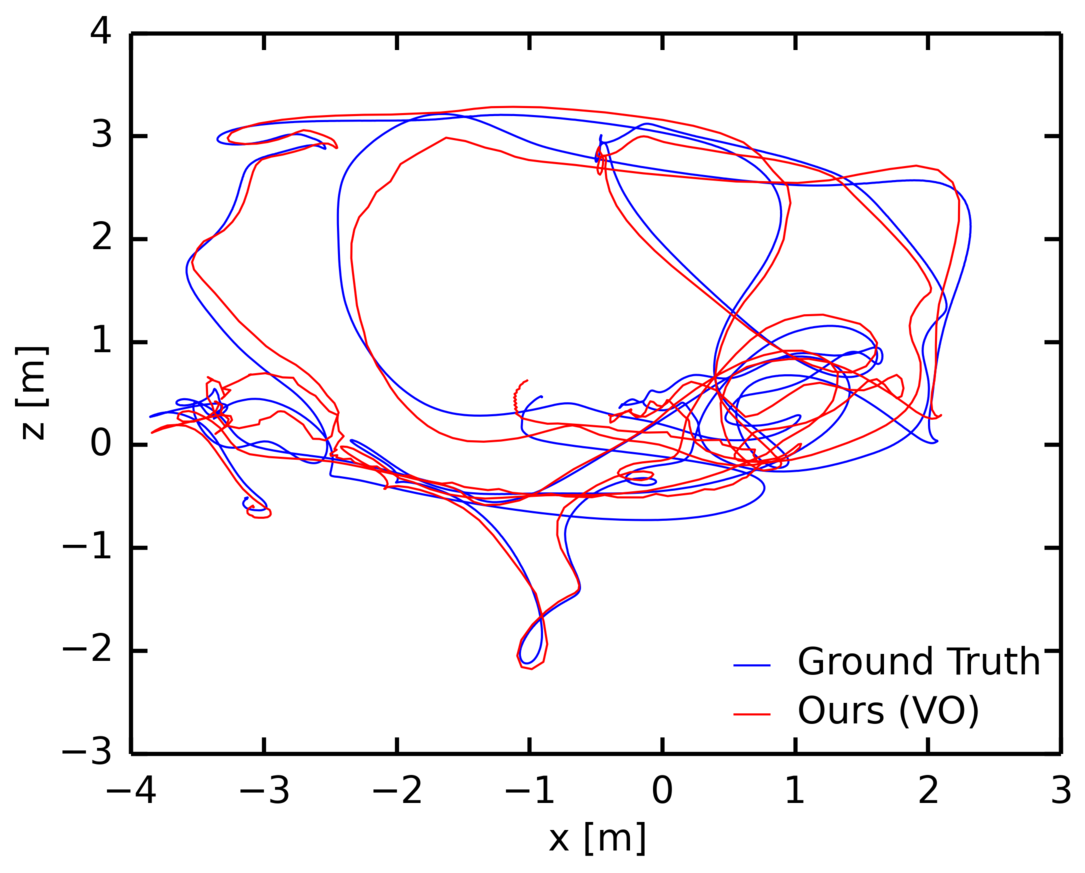}\\
    \caption[Comparison of LIBVISO2 and our Odometry on EuRoC]{Comparison of the results from LIBVISO2 (left) to our semi-direct odometry (right) on datasets V2\_01 and V2\_02 with ground truth from a Vicon motion capture system. Again our method is much closer to the ground truth even without SLAM.}
    \label{fig:euroc2libviso}
\end{figure*}

The dataset is known to have different issues that make a reliable state-estimation more challenging:
for example, the stereo images were captured using an automatic exposure control that is independent for both cameras. Therefore, shutter times are different, which results in different image brightnesses, making stereo matching and feature tracking more challenging. This is especially important, as direct methods minimize the photometric error.

Moreover, as the ground truth is recorded from a different physical device than the images, the accuracy depends on the synchronization scheme used~\citep{Burri25012016}. 

\begin{figure*}[t]
    \centering
    \includegraphics[width=0.5\linewidth]{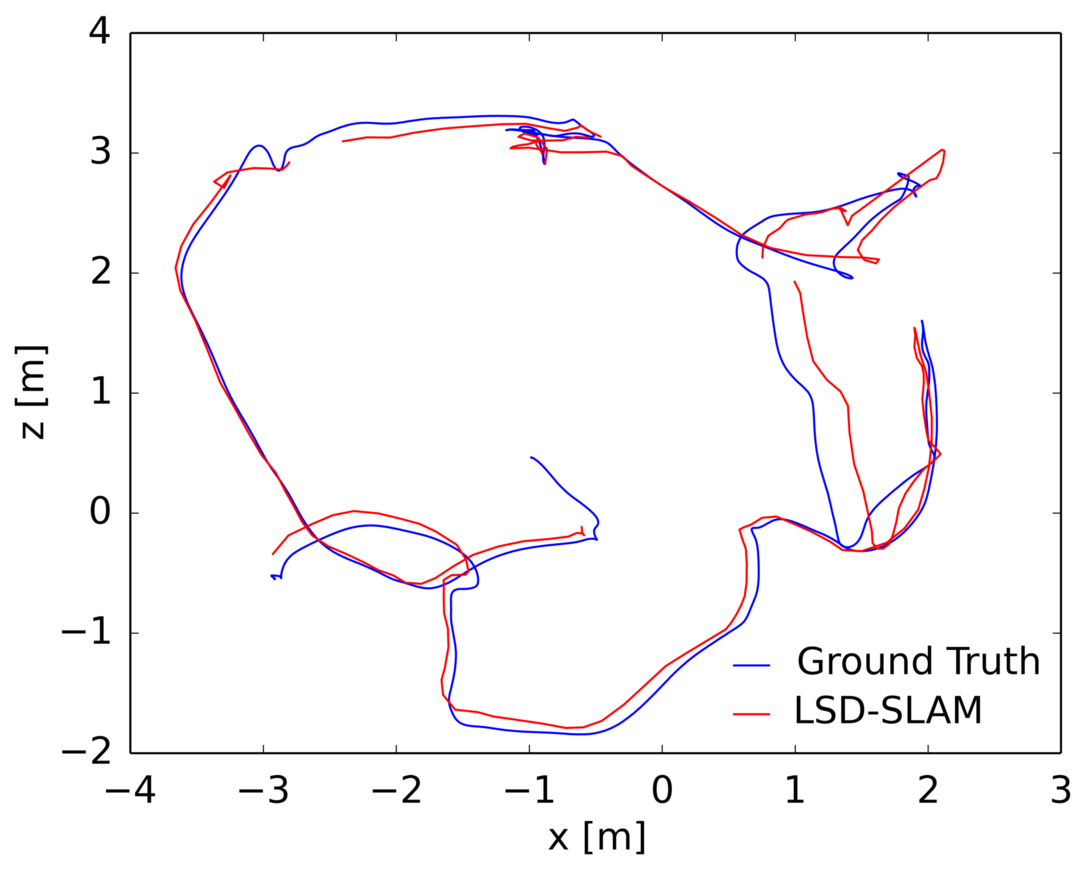}\hfill
    \includegraphics[width=0.5\linewidth]{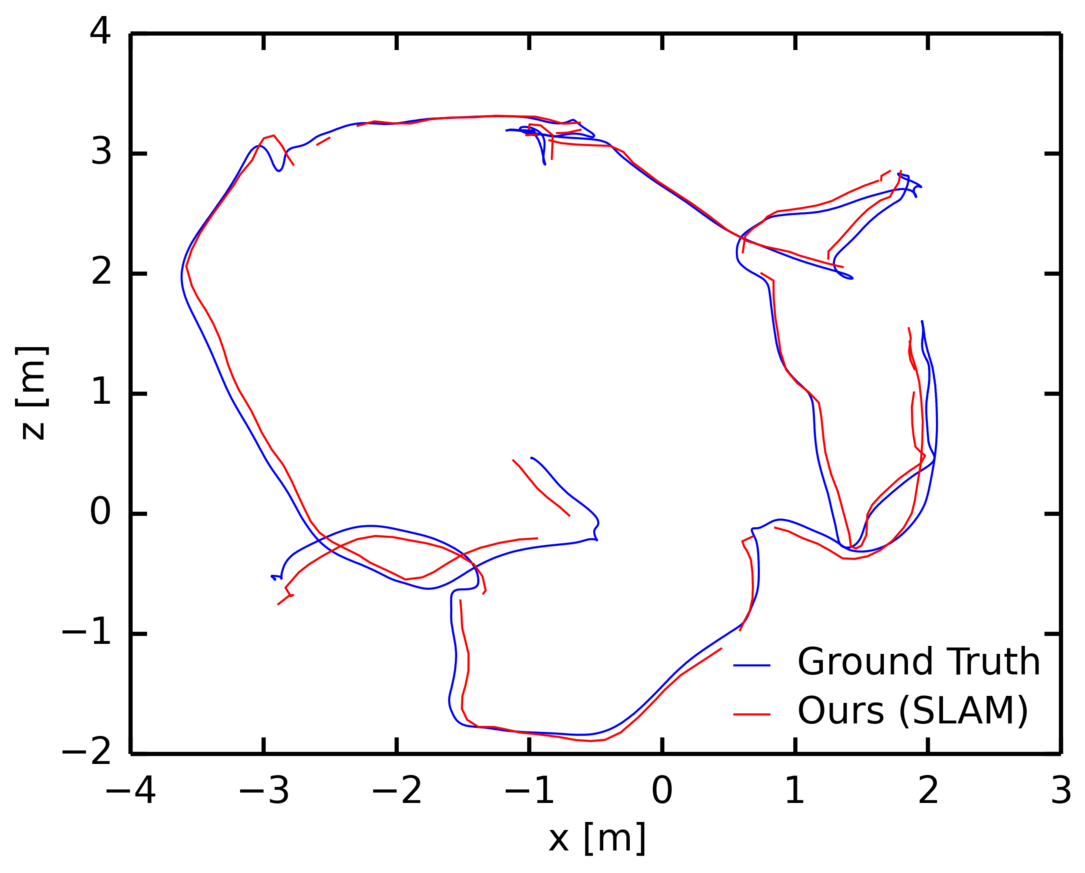}\\
    \caption[Comparison of LSD-SLAM and our SLAM on EuRoC]{Comparison of the results from LSD-SLAM (left) to our semi-direct SLAM (right) on dataset V2\_01 with ground truth from a Vicon motion capture system. While LSD-SLAM shows an ATE of \SI{0.45}{\m} our methods performs better with an ATE of \SI{0.18}{\m}}
    \label{fig:euroc2direct}
\end{figure*}

The resulting ATEs are listed in \cref{tab:euroc_ate}. As the difficult datasets V1\_03 and V2\_03 contain very dynamic movements and fast rotations with an MAV, LSD-SLAM often loses track after a few seconds and is then unable to re-localize for the rest of the trajectory. In \cref{tab:euroc_ate} this is denoted as failure (X).
Similarly, S-PTAM and ORB-SLAM lose track for the difficult trajectory V2\_03. This dataset shows very challenging conditions with strong motion blur and fast aggressive maneuvers.
Moreover, the absence of sufficient visual features makes it hard for feature-based methods to succeed.

\cref{tab:euroc_ate} also shows, that our approach outperforms the other methods and reliably recovers the motion for all test sequences. Additionally, it can be seen, that the results of LIBVISO2 are improved on every trajectory.

On average semi-direct SLAM achieves a higher accuracy, with \SI{0.38}{\m} ATE, than LSD-SLAM, with \SI{0.53}{\m}, ORB-SLAM, with \SI{1.23}{\m} ATE, and S-PTAM with \SI{1.82}{\m} ATE.
LSD-SLAM, ORB-SLAM and S-PTAM often perform poorly at fast motions in combination with rotations, and then tend to lose track temporarily.

\begin{figure*}[ht!]
    \centering
    \includegraphics[width=0.5\linewidth]{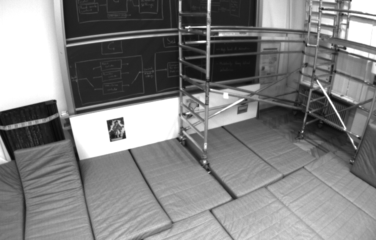}\hfill
    \includegraphics[width=0.5\linewidth]{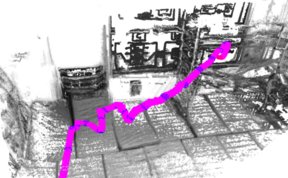}
    \caption[Resolved Trajectory on EuRoC]{Exemplary results of flight Vicon\_m: While the left image shows a capture of the recorded scene, the right image shows the retrieved camera trajectory and reconstructed semi-dense depth. Key frames are shown in blue, while feature-based tracked frames are shown in pink.}
    \label{fig:euroc}
\end{figure*}

Additionally, we again directly compare results from Semi-Direct Visual Odometry to LIBVISO2 and to Direct Odometry from LSD-SLAM.
\cref{fig:euroc2libviso} shows the resulting trajectories for datasets V2\_01 and V2\_02 of LIBVISO2 and Semi-Direct Visual Odometry. Both methods were performed without loop closures, and, thus, accumulate small errors in the estimates over time. It can clearly be seen that the Semi-Direct Odometry is closer to the ground truth from the Vicon system than LIBVISO2. 
Even though the datasets contain rapid rotations, our method stays close to the ground truth path.

In comparison to LSD-SLAM, our approach is more robust to fast rotations in the trajectory as can be seen in \cref{fig:euroc2direct}. While LSD-SLAM computes wrong estimates at strong turns, our method follows the path more precisely.

In addition to the official datasets we performed one manual flight in the Vicon room, named Vicon\_m, where we evaluated the mapping abilities of our approach.
As an example sequence for our mapping abilities, \cref{fig:euroc} shows a sequence captured on a manual flight: the MAV captures a corner of the Vicon room and is able to reconstruct a semi-dense 3D representation of the recorded scene.
As can be seen, details of the scaffold are retrieved as well as the ground plane and drawings on the blackboard. The recovered camera trajectory is shown as well. Key frames are colored in blue while frames that were tracked feature-based with LIBVISO2 are shown in pink.

In total, we showed that our method is more robust to dynamic motions than the other evaluated methods and achieves a lower ATE on all evaluated datasets.

\subsection{MAV}\label{sec:MAV}
\begin{figure}
      \centering
  	\includegraphics[width=0.6\linewidth]{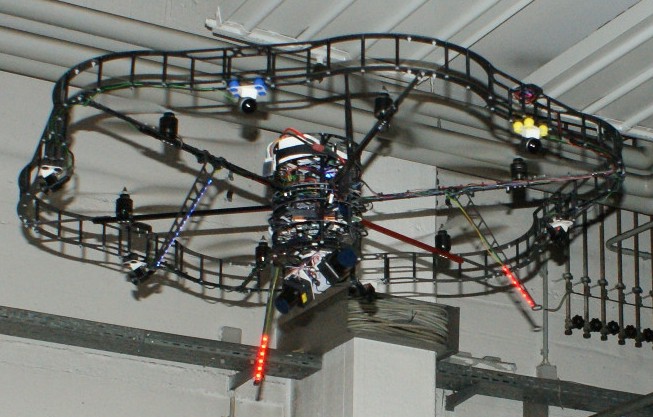}
      \caption[MAV with Omnidirectional Sensor Setup]{High performance MAV during flight. The omnidirectional sensor setup includes three fish eye stereo pairs covering a wide field of view for autonomous navigation.}
      \label{fig:mav_photo}
\end{figure}

In the previous section, we demonstrated that our semi-direct approach is capable of accurate pose estimation with standard stereo cameras.
We furthermore evaluate the performance of our approach on different datasets, that have been acquired with our MAV, shown in \cref{fig:mav_photo}. 
In contrast to the setups in previous datasets, our MAV is equipped with fish eye lenses and a wide baseline.

Our MAV is built as high-performance platform with a multimodal omnidirectional sensor setup~\citep{Beul2017}.
As MAVs have very limited payload, we use only lightweight components and are capable of navigating indoor and outdoor.
Especially for (fully) autonomous navigation in unknown and dynamic environments, a multimodal and omnidirectional sensor setup is of great advantage.
The strengths of the different sensors can be combined and their measurements can be fused in an occupancy grid map.

The MAV is built as hexarotor with six  $\SI{14}{\inch}$ propellers each connected to a MK3644/24 motor.
For better stability and collision protection the MAV is surrounded with a non-rigid milled frame that does not only protect the rotors, but also serves as mount for various sensors.
For on-board computation in real-time, the MAV is equipped with a mini-ITX board, namely a Gigabyte GB-BXi7-4770R with an Intel Core i7-4770R quad-core CPU, 16~GB DDR-3 memory and a 480~GB SSD to process all sensor outputs.

We employ a multimodal sensor setup consisting of IMU, laser scanners and cameras.
Moreover, our system is equipped with two laser scanners and six cameras for high-level autonomous operation and navigation.
In particular, we use two rotating Hokuyo UST-20LX laser scanners, each with a scan range of \SI{20}{\meter} and \ang{270} apex angle.
Together they can perform a full 3D scan of the environment with \SI{4}{\hertz}.
They are used for obstacle perception and SLAM-based 6DOF localization~\citep{droeschel2016multilayered}.

\begin{figure}
      \centering
  	  \includegraphics[width=0.49\linewidth]{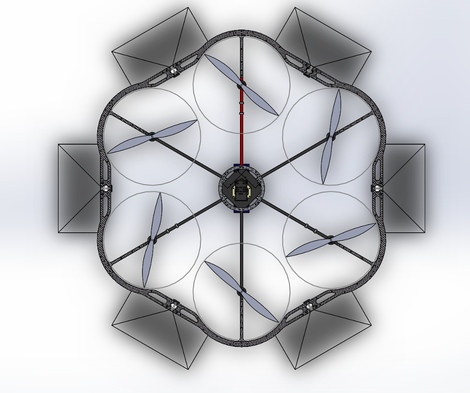}\hfill
	\includegraphics[width=0.49\linewidth]{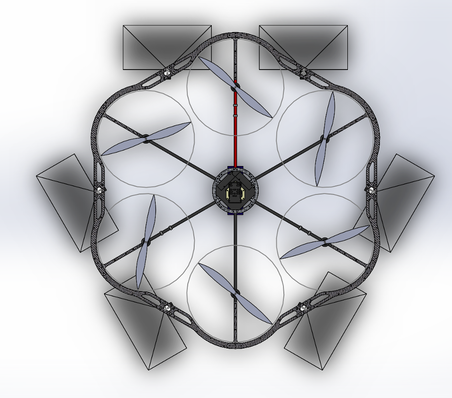}
      \caption[Camera Setup of our MAV]{Top down view of possible camera configurations: the left image shows a fully omnidirectional setup with independent optical axis, while on the right a stereo setup consisting of three independent stereo pairs is shown.}
      \label{fig:mav_cameras}
\end{figure}

For visual obstacle detection and visual SLAM, the MAV is equipped with an omnidirectional camera setup.
The cameras are mounted to the non-rigid body frame using dampers to filter out vibrations induced by the six propellers.
The camera mounting can easily be switched from a fully omnidirectional setup with independent optical axes to a stereo setup with three stereo camera pairs, as can be seen in \cref{fig:mav_cameras}.
The multi-camera setup allows omnidirectional perception of the environment and allows robust state estimation due to redundant information sources, i.e., even if one stereo pair faces a homogeneous wall with no texture the other two pairs still allow for robust localization. 
We use XIMEA MQ013MG-E2 global-shutter monochrome USB 3.0 cameras with 1.3~MP resolution in combination with Lensagon BF2M2020S23 fish-eye lenses for a wide field of view.
By making use of the available independent USB controllers of the on-board system, we distribute the USB traffic and thus can achieve high camera frame rates at full resolution.
Each stereo pair is connected to a USB 3.0 HUB, which is connected to a dedicated on-board USB 3.0 port that offers full USB 3.0 speed.
Through this setup we ensure that for each camera enough bandwidth is available. Assuming that each HUB offers \SI{2400}{\mega\bit / \second} (\SI{300}{\mega\byte/\second}), each camera may use up to \SI{1200}{\mega\bit / \second} (\SI{150}{\mega\byte/\second}).
Theoretically, each camera can achieve the best possible frame rate of \SI{60}{\hertz} in 8-bit mode and \SI{57}{\hertz} in 16-bit mode.

However, the real data rate is limited by additional system and protocol overhead when reading and writing from the connected devices.
Under real lighting conditions and depending on exposure times we achieve up to \SI{50}{\hertz} for each camera in 16-bit mode.
Our camera driver not only ensures that the images are published synchronously, but also offers advanced functionality like downsampling, gamma correction or rectification.

We use laser-based SLAM~\citep{droeschel2016multilayered} as ground truth and again compare the results with those of state-of-the-art SLAM methods.
In total we captured four flights in a decommissioned car service station with challenging lighting conditions.
While on the first two flights, named rect1 and rect2, the MAV covers a rectangular path without many loop closures, the other two flights, loop1 and loop2, include three to four full loops.  

\begin{figure*}[t!]
    \centering
    \includegraphics[width=0.5\linewidth]{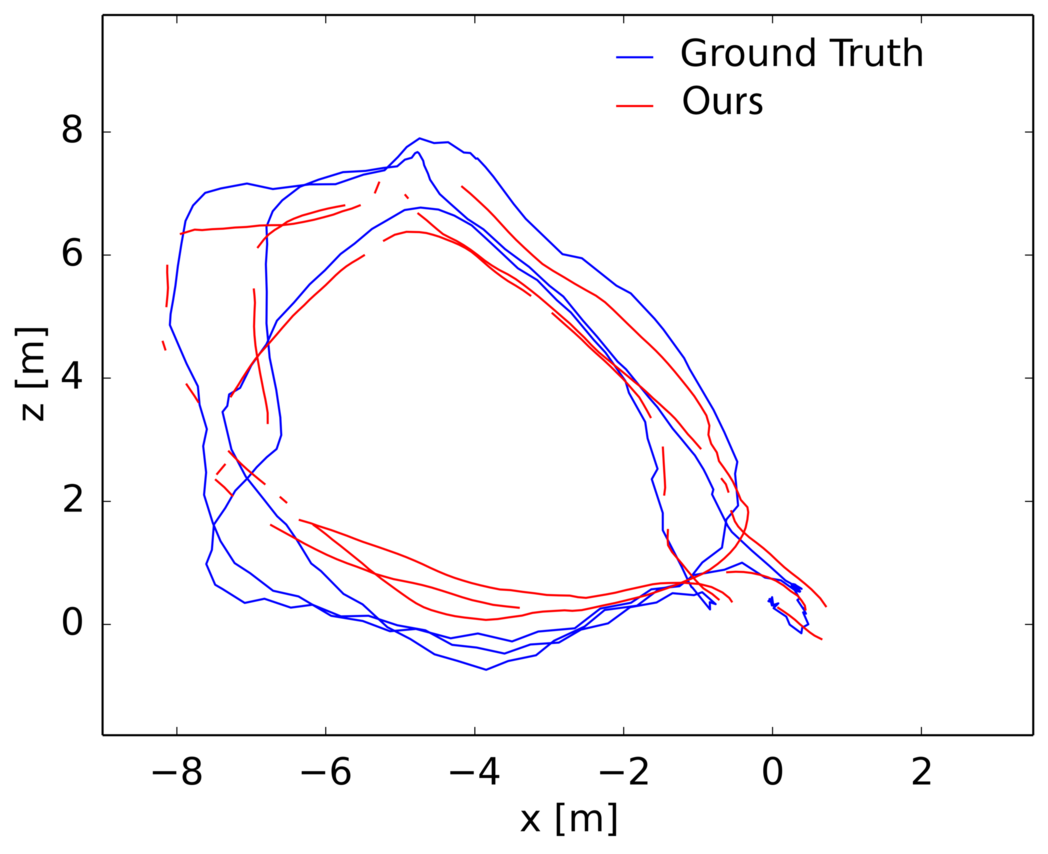}\hfill
    \includegraphics[width=0.5\linewidth]{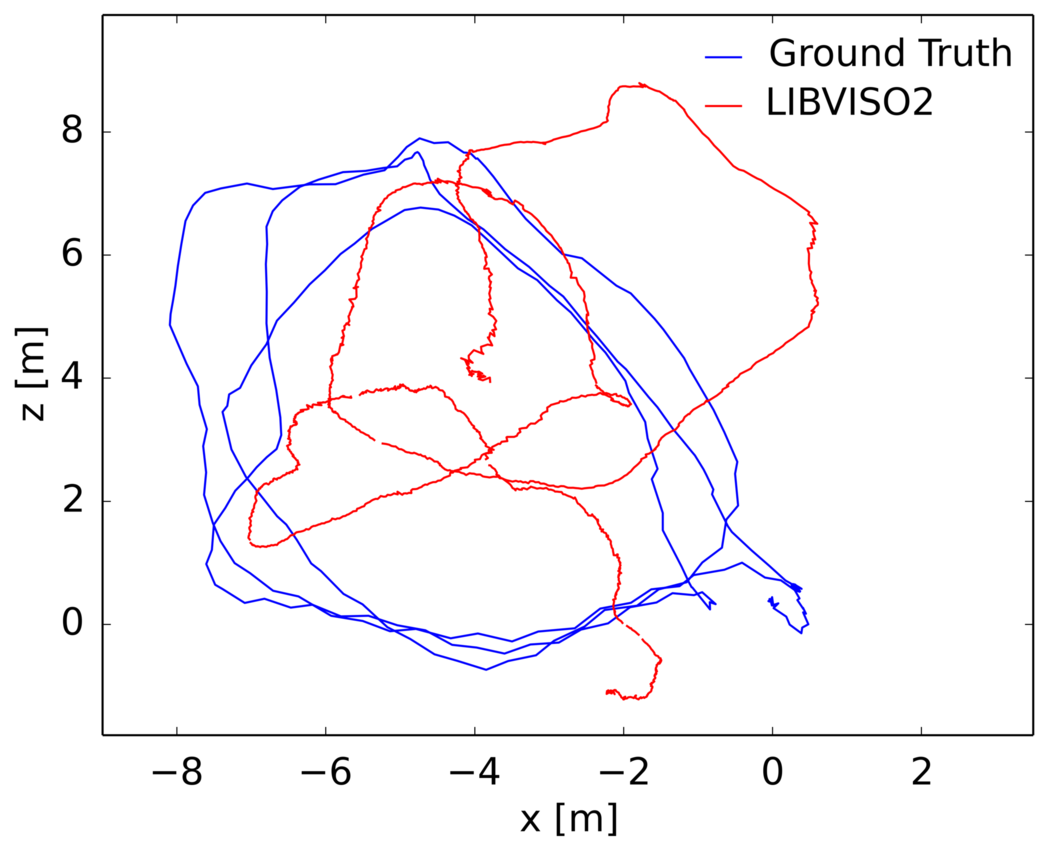}\hfill\\
    \includegraphics[width=0.5\linewidth]{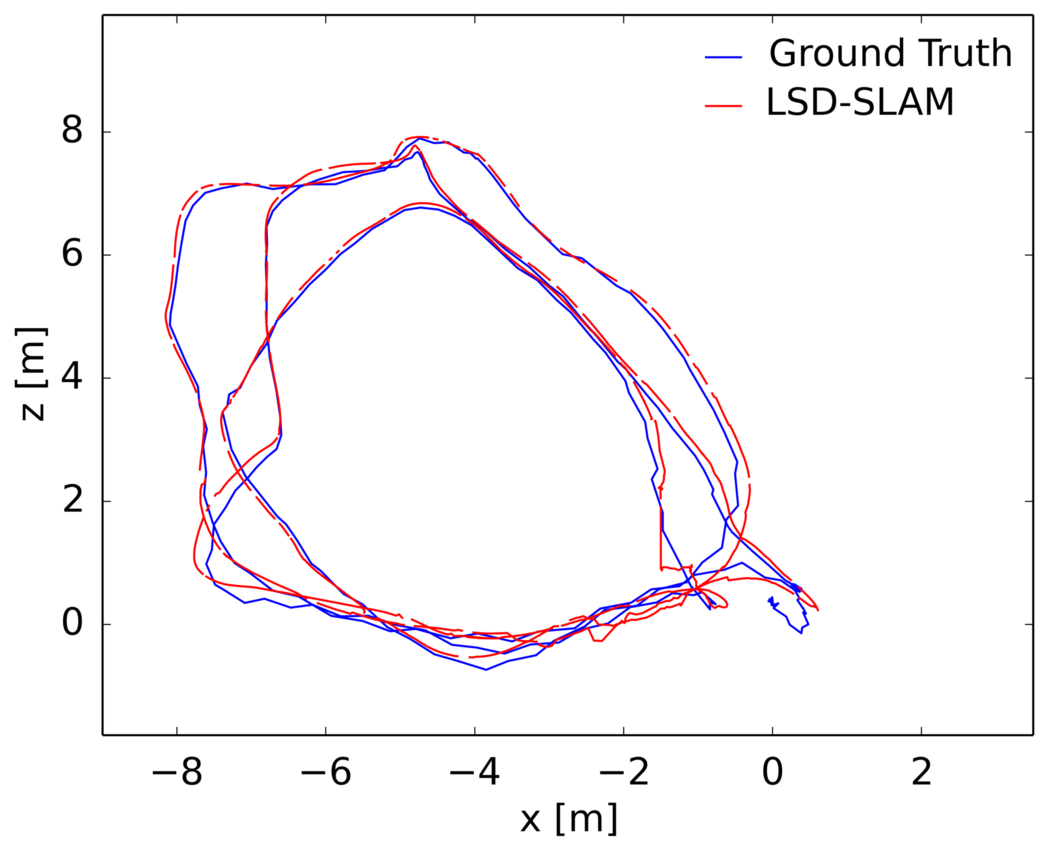}\hfill
    \includegraphics[width=0.5\linewidth]{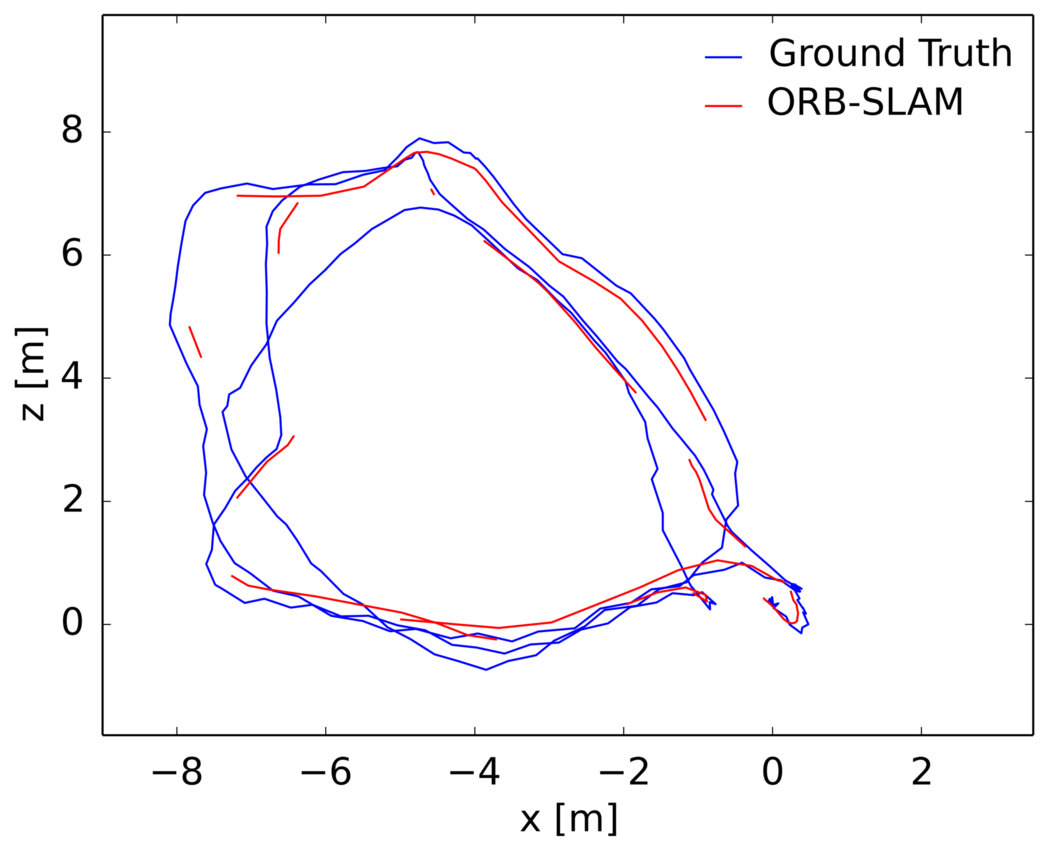}\hfill\\
    \caption[Results for our MAV]{Results for MAV loop 1. Comparison of our method to LIBVISO2 (top row), LSD-SLAM and ORB-SLAM (bottom row) on a challenging dataset that contains large loop closures. As can be seen the monocular methods perform best while LIBVISO2 accumulates strong drift. Still, our method is able to reconstruct the trajectory with an ATE of \SI{0.63}{\m} while S-PTAM fails completely.}
    \label{fig:loop1}
\end{figure*}

\begin{figure}[t!]%
\includegraphics[width=0.45\textwidth]{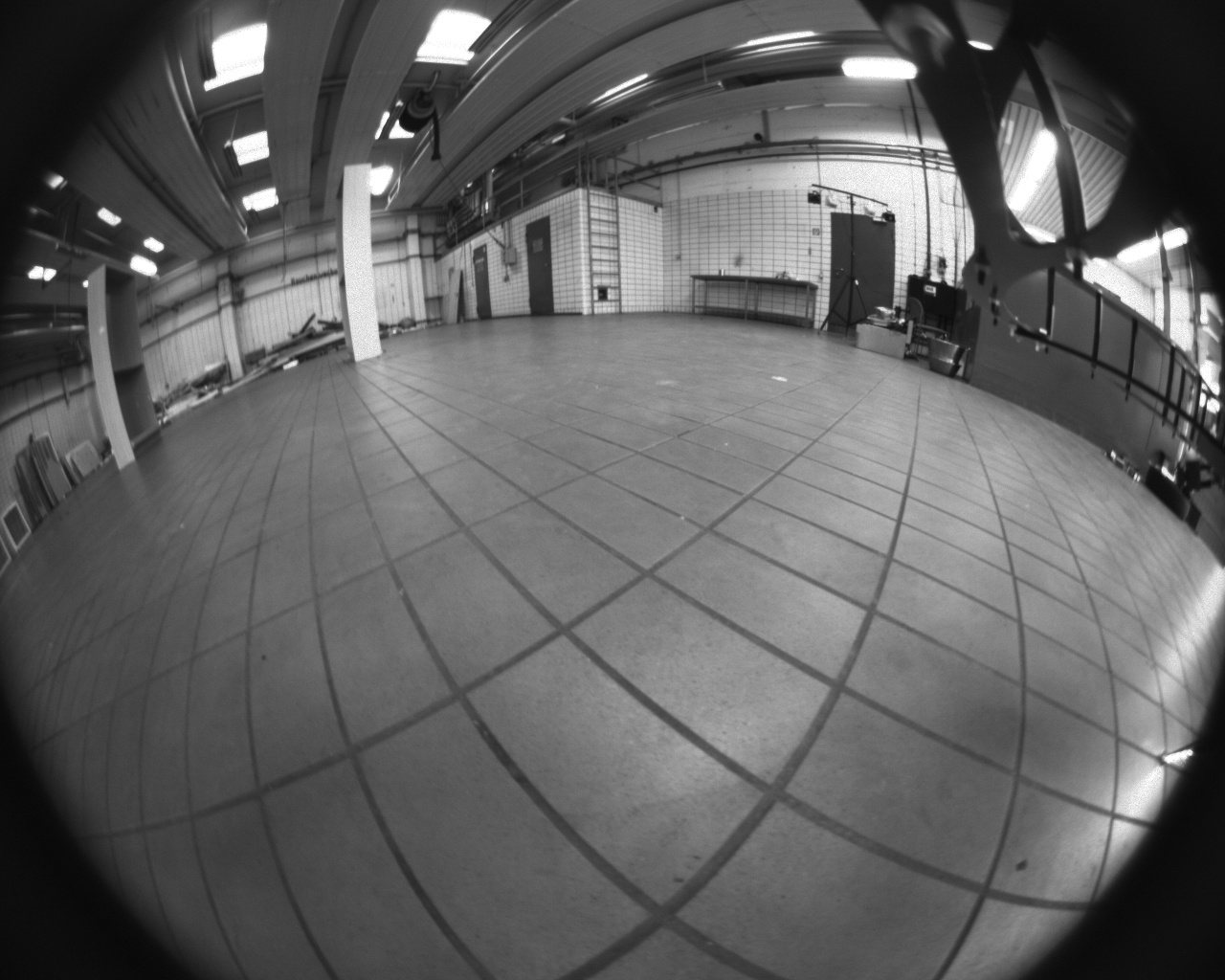}\hfill
\includegraphics[width=0.45\textwidth]{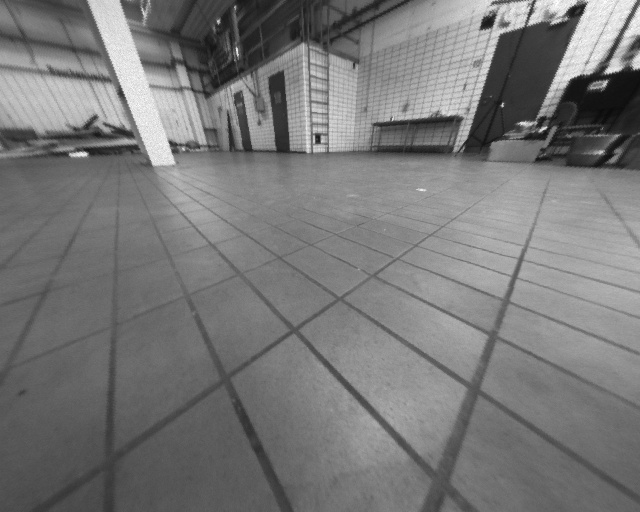}
\caption[Stereo Rectification]{Stereo Rectification: Raw fish eye images (left) are rectified onto a plane with half the resolution (left).}%
\label{fig:mav_rectification}%
\end{figure}

A general prerequisite for stereo computation is to rectify the images. %
To allow different models for calibration we build a general rectification nodelet in ROS, that rectifies the images given respective look-up tables as input.
The look-up tables can be either calculated offline beforehand or online using, \eg, the computer vision library OpenCV.
The rectification nodelet publishes rectified images together with camera info messages that contain the necessary calibration parameters from intrinsic and extrinsic calibration.
Moreover, we added functionality to down-sample the rectified images by a factor $c$ for further run time enhancement.
The images are captured with full resolution of $1280\times 1024$ in \SI{16}{\bit}-encoding and are down-sampled to half the resolution and \SI{8}{\bit} in the rectification step.
\cref{fig:mav_rectification} shows the result from the rectification step on an image from the recorded dataset.

\begin{figure}
    \centering
    \includegraphics[width=0.5\linewidth]{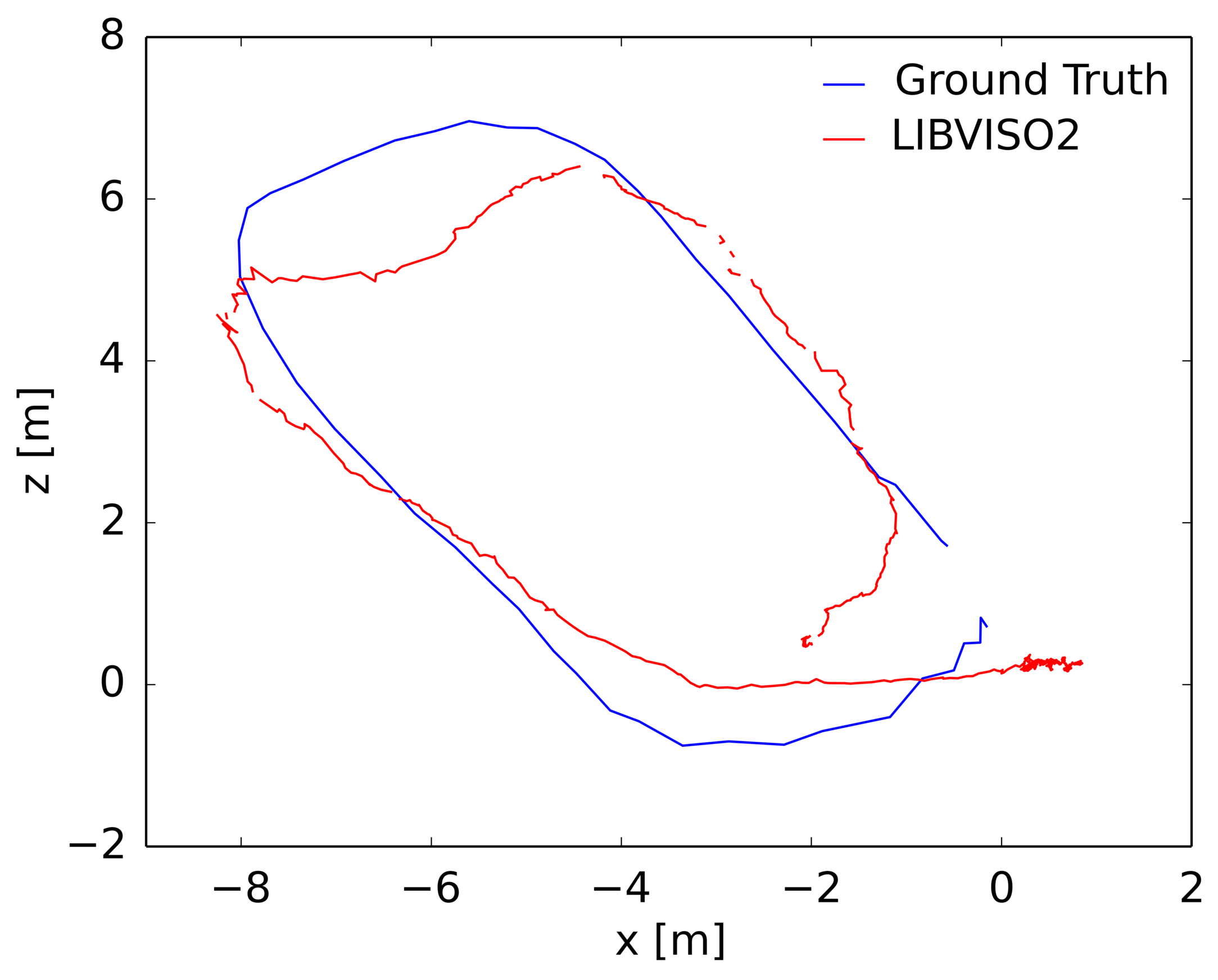}\hfill
    \includegraphics[width=0.5\linewidth]{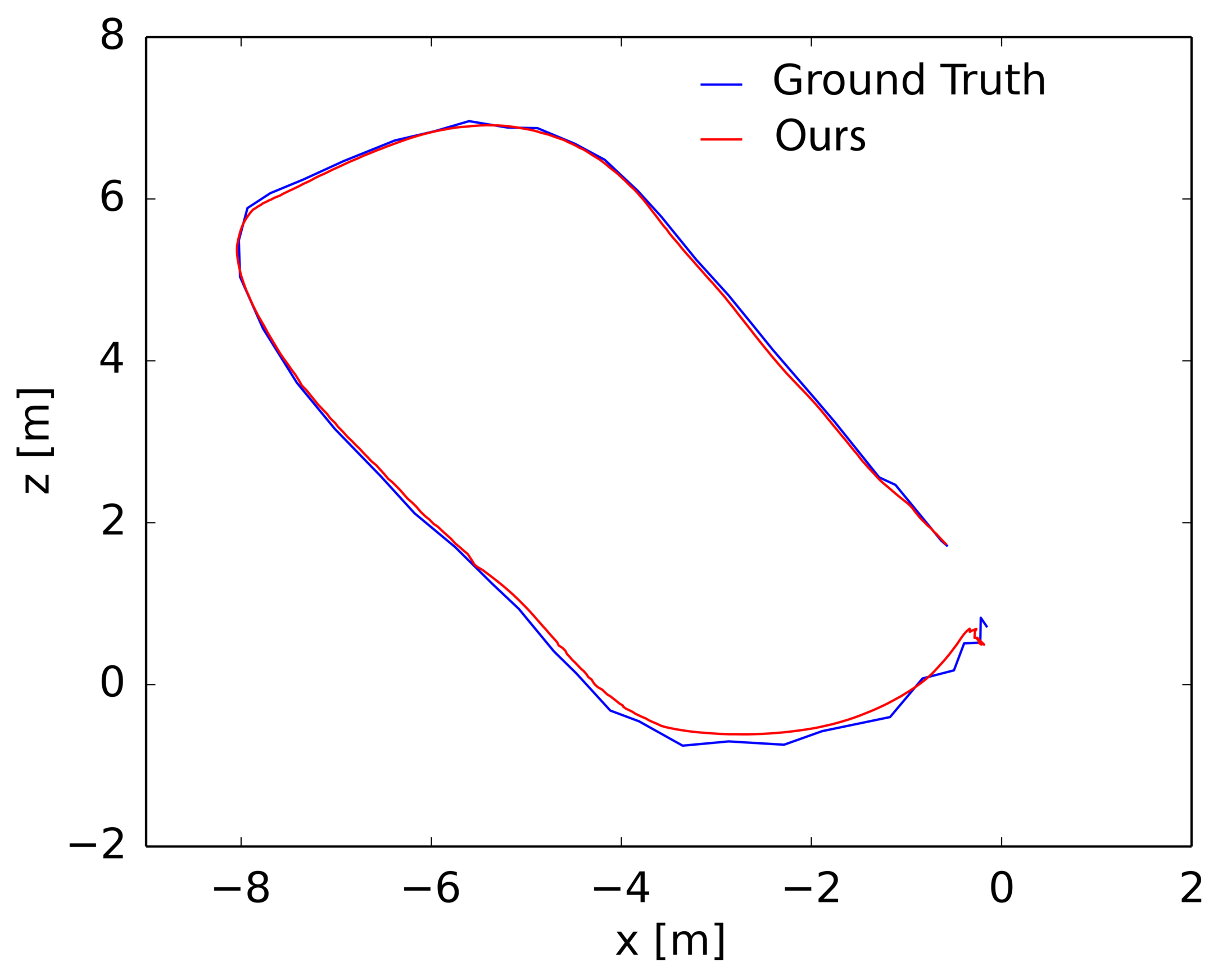}
    \caption[Comparison of LIBVISO2 and our on MAV]{Comparison of LIBVISO2 (left) and our method (right) on the MAV dataset rect1. Even though, the dataset contains no loop closure, our method shows accurate results.}
    \label{fig:rect1}
\end{figure}

The rectification of the images runs in parallel for all six cameras and takes \SI{1}{\milli\second} for a single image when downsampling to half the original resolution.
For an even smaller resolution of $320 \times 256$ the rectification takes \SI{0.7}{\milli\second} and for the full resolution \SI{4}{\milli\second}.

Experiments on the four flights show that retrieving the correct camera motion is more challenging than on the previous datasets.
Especially, the stereo methods do poorly on these datasets. S-PTAM fails to initialize correspondences on all datasets and, thus, cannot be taken into account for comparison.

Exemplary results are shown for a trajectory with repeating loop closures in \cref{fig:loop1}.
It can be seen that the stereo methods Semi-Direct SLAM and LIBVISO2 show a higher offset to the ground truth trajectories than the monocular methods.
Especially LIBVISO2 accumulates high errors at this circular trajectory and the result is not as accurate as before, thereby hindering the semi-direct approach.
As LIBVISO2 performs no loop closure detection, errors in the absolute trajectory cannot be resolved which leads to a globally inconsistent trajectory.
Semi-Direct SLAM uses only the relative motion estimates of LIBVISO with regards to the current key frame.
Thereby, Semi-Direct SLAM is still able to reconstruct a path close to the ground truth with an ATE of \SI{0.63}{\m}.
Contrarily, LSD-SLAM and ORB-SLAM achieve an ATE below \SI{0.31}{\m}.

\begin{table}
	\small
    \begin{center}
    \begin{tabular}{c||c||c||c||c}
    MAV & \multicolumn{4}{c} {\textbf{Absolute Trajectory Error RMSE (Median) in m}}\\
	~~Dataset~ & ~~~Ours~~~ & ~Libviso2~ & ~LSD-SLAM~ & ~ORB-SLAM~ \\
    \hline
    rect1 & \bftab{0.13} (0.11)   & 1.24 (0.49) & 0.30 (0.29) & 0.98 (0.24)  \\
    rect2 & 0.84 (0.81)   & 1.61 (1.59) & \bftab 0.38 (0.37) & 0.59 (0.25)  \\
    loop1 & 0.63 (0.57)   & 1.66 (0.99) & 0.31 (0.28) & \bftab{0.25 (0.21)} \\
    loop2 & 1.58 (0.71)   & 2.61 (1.90) & \bftab{0.54 (0.42)} & 1.19 (0.78) \\
    \hline
    mean   & 0.80 (0.55)   & 1.78 (1.24) & \bftab{0.38 (0.34)} & 0.75 (0.37) \\
    \hline
    \end{tabular}
    \end{center}
    \caption{ATE Results on MAV Dataset}
    \label{tab:mav_ate}
\end{table}

The fact that monocular methods seem to perform better than stereo methods, 
suggests that the underlying lens distortion model was chosen to allow for a rectification mapping and, hence, semi dense
stereo matching, but does not fit the used fish eye lenses very accurately. 
Additionally, the non-rigid mounting of the stereo cameras introduces difficult conditions for stereo correspondence search along fixed epipolar lines.
We assume that the wide non-rigid baseline of \SI{53.37}{\cm} in combination with the perspective rectification onto a plane impedes the stereo correspondence search.
It would generally be more appropriate to model the fish eye lenses as a projection onto a sphere.
As described in \cref{sec:map_update}, we use an additional weighting scheme that downweights the influence of inaccurate depth measurements close to the image borders in order to cope with strong distortions.  	
Moreover, we repeatedly estimate the extrinsic transformation of the cameras online to handle the non-rigidity.
Therefore, we are able to retrieve stereo correspondences and estimate the trajectory on this challenging dataset, unlike S-PTAM which fails to initialize any correspondences.

\cref{fig:rect1} shows trajectories for the sequence rect1 computed by LIBVISO2 and Semi-Direct SLAM. The output of LIBVISO2 shows very noisy estimates and leads to a comparably high ATE of \SI{1.24}{\m}.
In contrast Semi-Direct SLAM produces a smoother trajectory with an ATE of \SI{0.13}{\m}.
However, in general we achieve a higher ATE than the monocular methods.
\cref{tab:mav_ate} summarizes the resulting ATE on all datasets.

In terms of accuracy the monocular methods perform better than all stereo methods. This time S-PTAM is unable to track features on all datasets and fails in recovering any motion.
It is remarkable that monocular methods perform better than stereo methods on these datasets which supports our assumption that the rectification of the fish eye images onto a plane in combination with non-rigidly mounted stereo cameras is very challenging for stereo computations.
Moreover, the wide baseline is problematic as the image overlap between both stereo images is reduced.

On average, we achieve an ATE of \SI{0.8}{\m} while monocular LSD-SLAM achieves an average ATE of \SI{0.38}{\m}.

\subsection{Odometry versus SLAM}

\begin{table}[ht!]
\small
    \begin{center}
    \begin{tabular}{c||c||c||c}
    \shortstack{EuRoC\\Dataset} & \multicolumn{3}{c} {\textbf{\shortstack{Absolute Trajectory Error\\RMSE (Median) in m and Improvement in \%}}}\\
    & \shortstack{Our\\Semi-Direct VO [m]} & \shortstack{Our\\Semi-Direct SLAM [m]} & Improvement [\%]\\
    \hline		
    V1\_01 & 0.26 (0.18) & 0.12 (0.11) & 53.85 (38.89) \\
    V1\_02 & 0.59 (0.59) & 0.11 (0.10) & 81.36 (83.05) \\
    V1\_03 & 0.81 (0.76) & 0.75 (0.44) & \hphantom{0}7.41 (42.11)  \\
    V2\_01 & 0.22 (0.13) & 0.18 (0.12) & 18.18 \hphantom{0}(7.69) \\
    V2\_02 & 0.31 (0.25) & 0.27 (0.22) & 12.90 (12.00)  \\
    V2\_03 & 1.13 (0.97) & 0.87 (0.66) & 23.01 (31.96) \\
    \hline
    mean   & 0.55 (0.48)   & 0.38 (0.28) & 30.72 (42.71) \\
    \hline
    \end{tabular}
    \end{center}
    \caption{Odometry compared to SLAM on EuRoC}
    \label{tab:slamvsvo}
\end{table}

In this section, we will compare the quantitative results of visual odometry to visual SLAM. As visual odometry tends to drift over time, global optimization methods such as bundle adjustment or pose graph optimization help to reduce the drift.

In Semi-Direct SLAM loop closures are detected between key frames and are added as additional constraints to the global pose graph (see \cref{sec:global_map_optimization}).

The trajectories of the EuRoC dataset contain many possible loop closures. Therefore, we show comparative results between visual odometry and SLAM exemplary on this dataset.
Qualitative results are listed in \cref{tab:slamvsvo}. In addition to the ATE as error measure, we also state the percentage improvement gained by SLAM.
We measure the improvement as 
\begin{equation}
 Improvement = \frac{VO-SLAM}{VO}.
\end{equation}

\begin{figure*}[t!]
\centering
\includegraphics[width=0.49\linewidth]{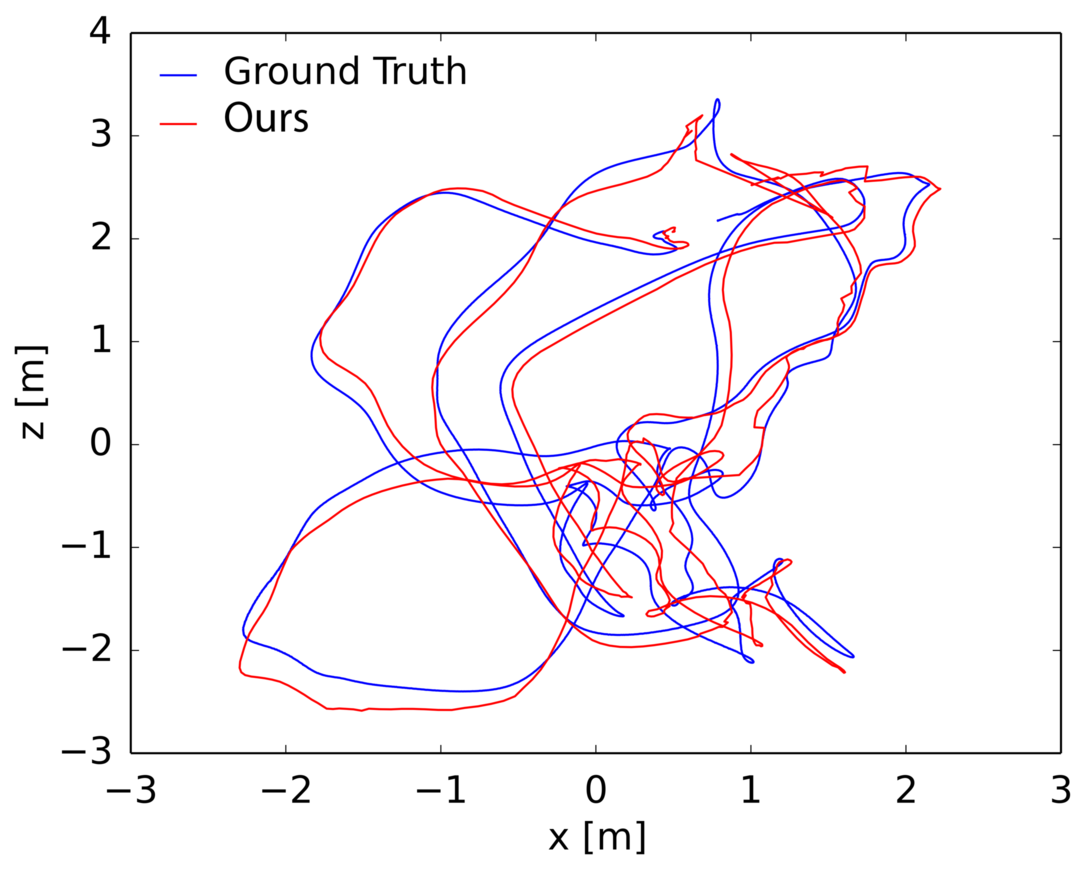}\hfill
\includegraphics[width=0.49\linewidth]{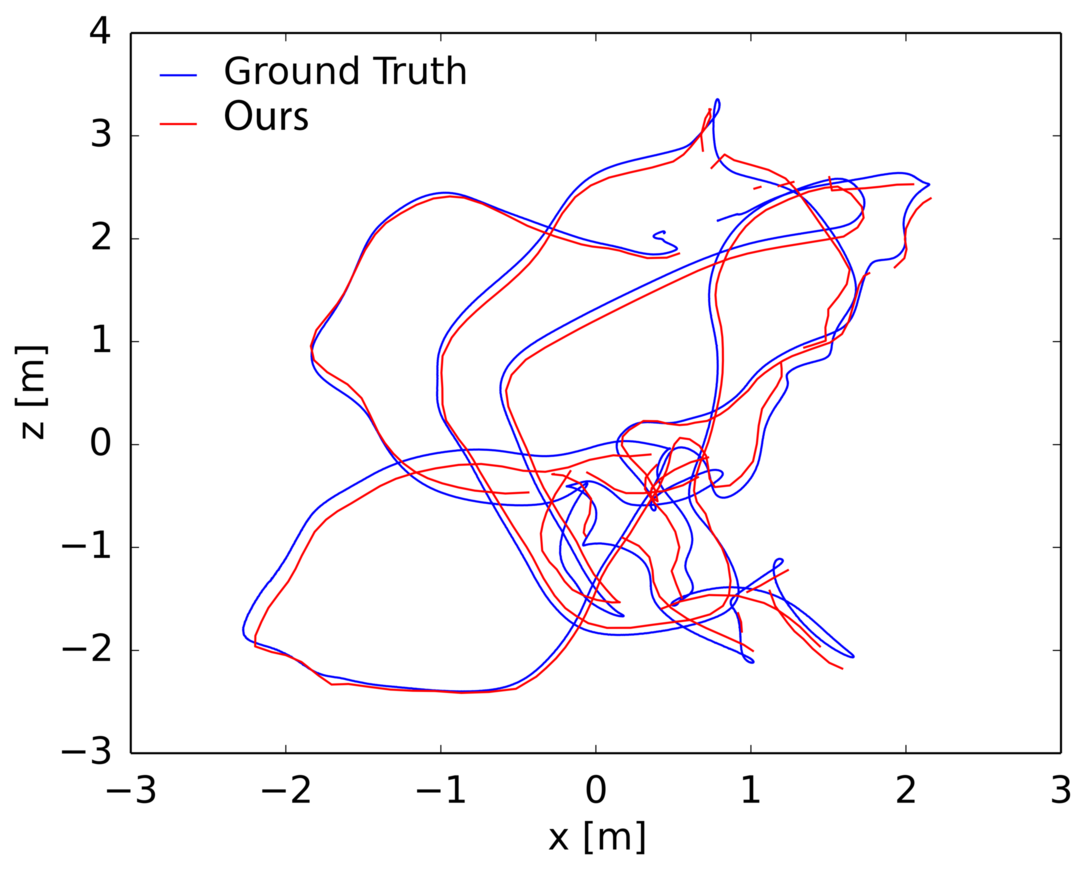}\\
\includegraphics[width=0.49\linewidth]{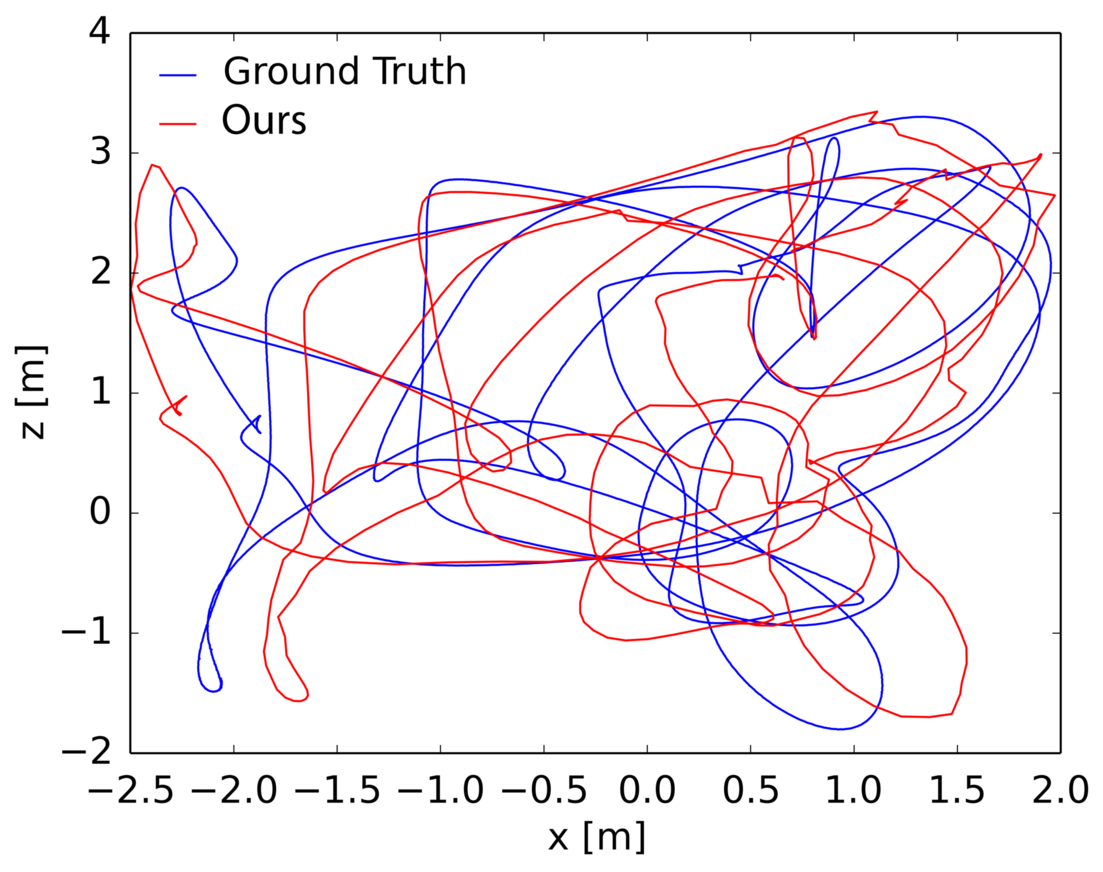}\hfill
\includegraphics[width=0.49\linewidth]{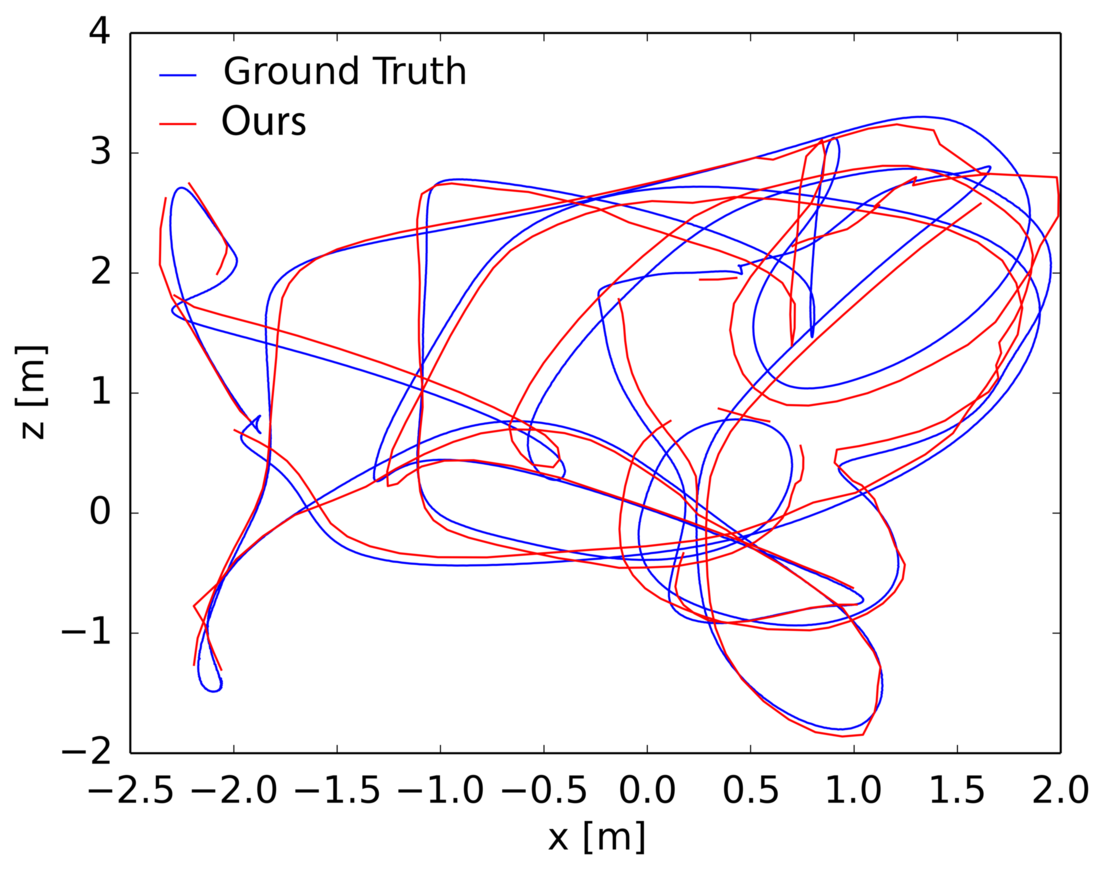}\\
\caption[Comparison of Odometry and SLAM on EuRoC]{Comparison of Semi-Direct Odometry (left) and Semi-Direct SLAM (right) on the EuRoC dataset V1\_01 and V1\_02. With SLAM loop closures are found and accumulated drift is corrected yielding percentage improvements of $53.85$\% and $81.36$\% respectively}
\label{fig:vovsslameuroc}
\end{figure*}

The average improvement for all seven trajectories lies at $30.72$\% denoting an absolute improvement of \SI{0.17}{m} on average.
It can clearly be seen that for each trajectory the odometry result is further improved by SLAM.
The improvements range from $7.41$\% up to $81.36$\%.
The maximum improvement reached an absolute enhancement of \SI{0.48}{m}. 
As the trajectories V1\_01 and V1\_02 show significant improvements of $53.85$\% and $81.36$\% respectively, both of the results are visualized in \cref{fig:vovsslameuroc}.
The advantages of SLAM are visible in either example.
In comparison to the pure odometry, SLAM retrieves trajectories closer to the ground truth.
The bottom row of \cref{fig:vovsslameuroc} highlights the improvement of $81.36$\% on dataset V1\_02.
This dataset is of medium difficulty and contains very dynamic translational and rotational movements. It can be seen, that the odometry might be locally accurate but exhibits accumulated drift.
In the global graph SLAM the drift is corrected by loop closures resulting in a better aligned trajectory.

\begin{figure*}
\centering
\includegraphics[width=0.49\linewidth]{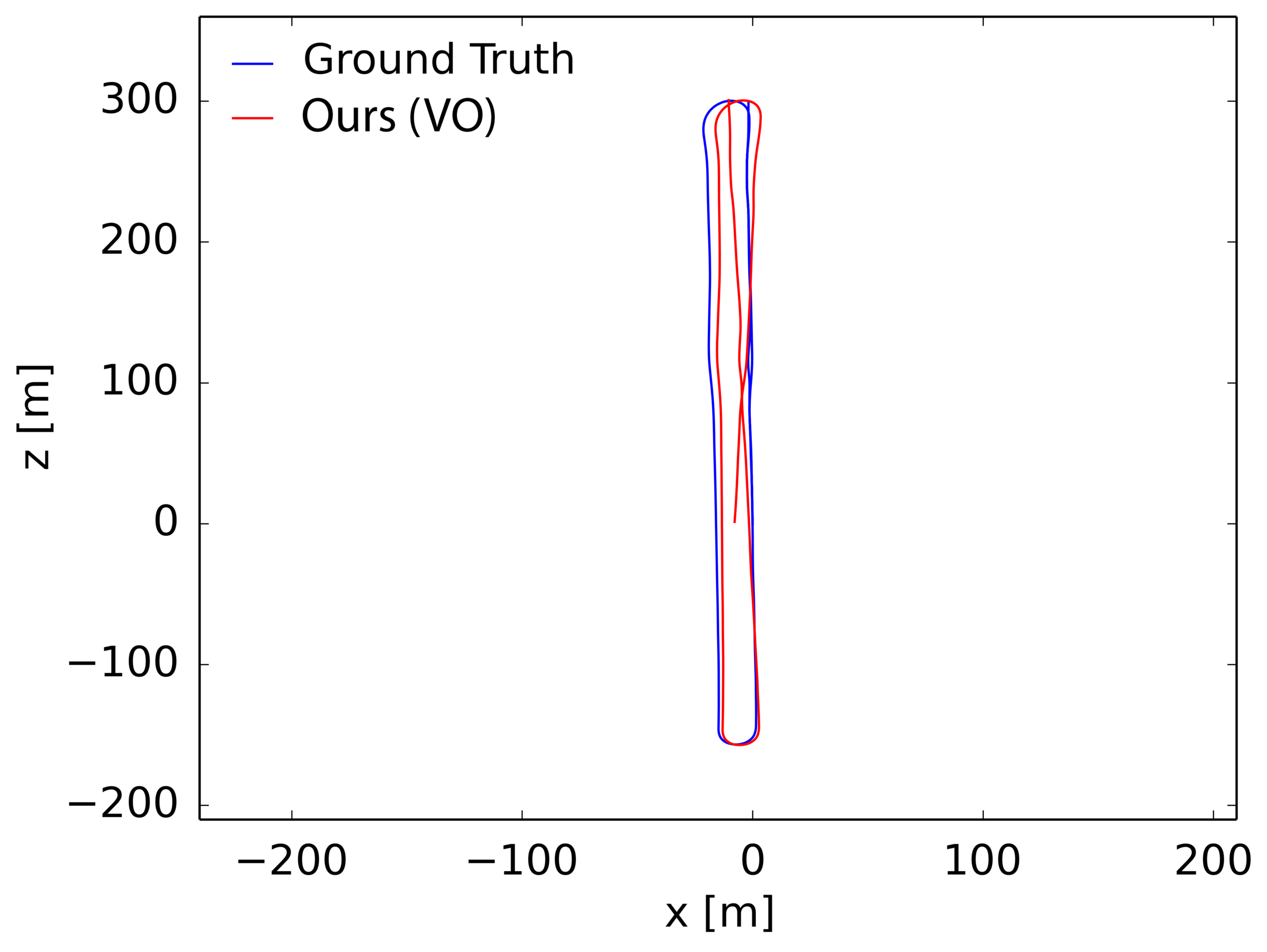}\hfill
\includegraphics[width=0.49\linewidth]{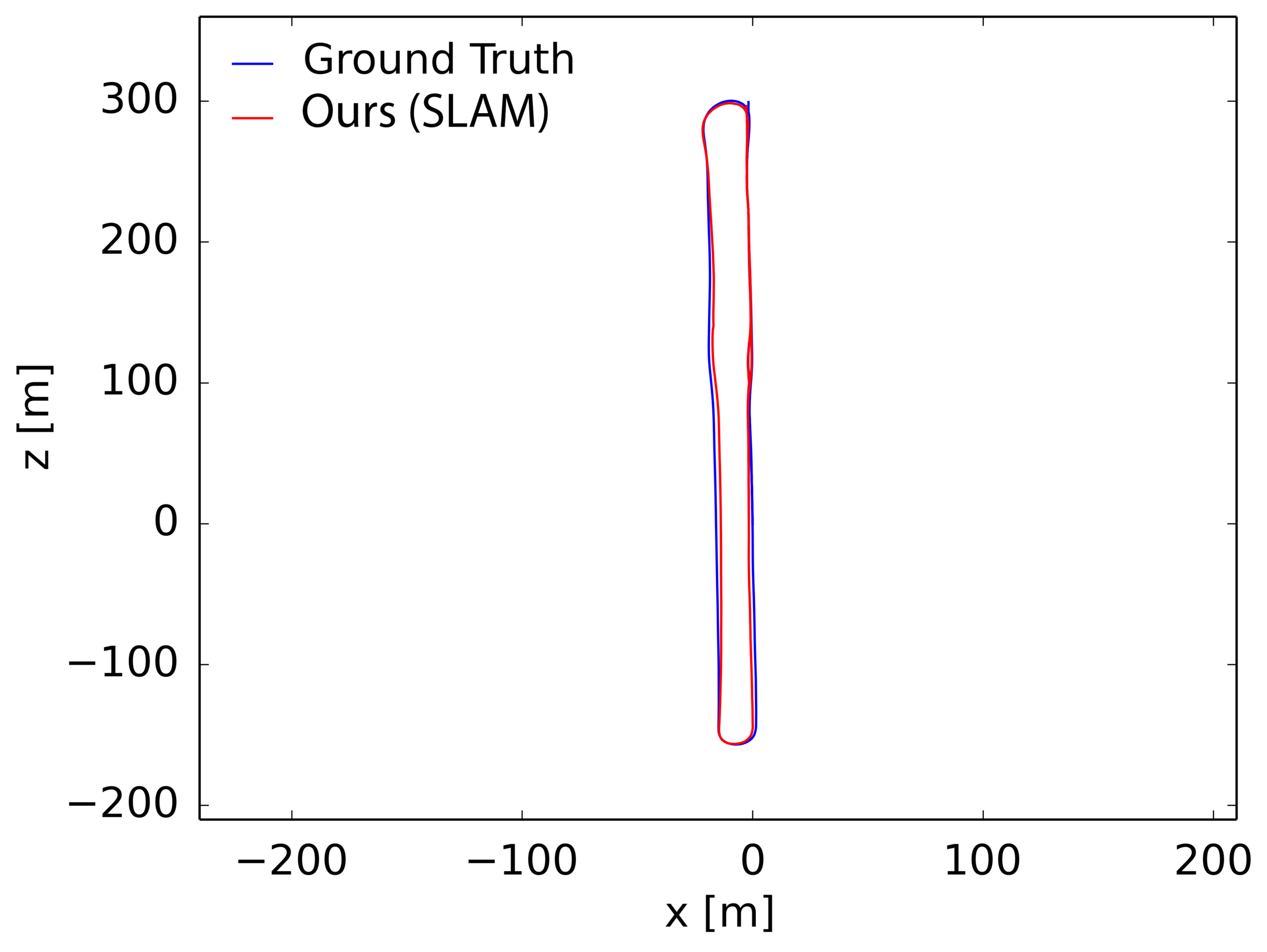}\\
\caption[Comparison of Visual Odometry and SLAM on KITTI]{Comparison of Semi-Direct Odometry (left) and Semi-Direct SLAM (right) on KITTI Sequence 06. Example of a full loop closure found by our SLAM method while pure odometry drifts. SLAM achieves an improvement of $52.9$\%.}
\label{fig:vovsslamkitti}
\end{figure*}

In contrast to the EuRoC dataset the KITTI dataset shows notably less loop closure possibilities. However, when loop closures are found, the global consistency of the map is re-established.
Sequence $06$ contains a distinct loop. While visual odometry produces an ATE of \SI{4.37}{m} on Sequence $06$, the result is corrected after closing the loop and the ATE decreases to \SI{2.06}{m}, showing an improvement of $52.9$\%.
\cref{fig:vovsslamkitti} illustrates this phenomenon: while the odometry drifts over time and does not retrieve the circular path, the SLAM extension closes the loop and continues the trajectory on the previously driven path.

Additionally, we also evaluate the performance of SLAM in comparison to pure odometry on our MAV.
Similarly to above results, loop closures greatly help to reduce the drift on the datasets loop1 and loop2.
While on dataset loop1 the odometry yields an estimate with \SI{1.1}{m} ATE, the visual SLAM recovers the camera motion with \SI{0.63}{m}.
On dataset loop2 the odometry result improves from \SI{2.16}{m} to \SI{1.58}{m} when performing SLAM.
The relative improvements on these datasets are $42.7$\% and $26.9$\% respectively.

\section{Accuracy}
\begin{table}
    \footnotesize
    \begin{center}
    \begin{tabular}{c||c||c||c||c||c}
	 & \multicolumn{5}{c} {\textbf{Absolute Trajectory Error RMSE (Median) in m}}\\
	\hspace{-0.2em}Dataset\hspace{-0.2em} & Ours & Libviso2 & \hspace{-0.2em}LSD-SLAM\hspace{-0.2em} & \hspace{-0.2em}ORB-SLAM\hspace{-0.2em} & S-PTAM\\
	\hline
	\hspace{-0.2em}KITTI\hspace{-0.2em} & \hspace{-0.2em}\bftab{10.28 (8.53)}\hspace{-0.2em}  			& \hspace{-0.2em}32.54 (26.42)\hspace{-0.2em} 			& X           		& \hspace{-0.2em}41.49 (35.66)\hspace{-0.2em} 	&  \hspace{-0.2em}25.74 (20.26)\hspace{-0.2em} \\
	\hspace{-0.2em}EuRoC\hspace{-0.2em} & \hspace{-0.2em}\bftab{\hphantom{0}0.38 (0.28)}\hspace{-0.2em}  	 & \hspace{-0.2em}\hphantom{0}0.85 \hphantom{0}(0.71)\hspace{-0.2em} & 0.53 (0.48) & \hspace{-0.2em}\hphantom{0}1.23 \hphantom{0}(0.94)\hspace{-0.2em} & \hspace{-0.2em}\hphantom{0}1.82 \hphantom{0}(1.52)\hspace{-0.2em} \\
	\hspace{-0.2em}MAV\hspace{-0.2em}   &  \hspace{-0.2em}\hphantom{0}0.80 (0.55)\hspace{-0.2em}           	& \hspace{-0.2em}\hphantom{0}1.78 \hphantom{0}(1.24)\hspace{-0.2em}   &\bftab{0.38 (0.34)} & \hspace{-0.2em}\hphantom{0}0.75 \hphantom{0}(0.37)\hspace{-0.2em}   & X \\
	\hline
    \end{tabular}
    \end{center}
    \caption[Average ATE]{Average ATE Results on the different evaluated datasets}
    \label{tab:avgacc}
\end{table}

\begin{table}
    \small
    \begin{center}
    \begin{tabular}{c||c||c||c}
	RPE & Ours (VO) & Libviso2 & Direct VO\\
	\hline
	Translation Error (\%)                                 & \bftab{0.8061}  & 0.8449 & 0.8168 \\
	Rotation Error (\si{\degree\per\m})    & \bftab{0.0051}  & 0.0052 & 0.0053 \\
	\hline
    \end{tabular}
    \end{center}
    \caption[Relative Pose Errors]{Relative pose errors of the odometry methods. Translational drift is measured in percentage and rotational drift in \si{\degree\per\m}. }
    \label{tab:avgrpe}
\end{table}

We have shown on different challenging datasets that in terms of accuracy we achieve similar results as current state-of-the-art stereo methods. 
The mean results for all datasets are summarized in \cref{tab:avgacc}.
As can be seen in the table, our method achieves a lower ATE than the other evaluated methods on the KITTI and EuRoC datasets.
On our MAV, monocular methods outperform the stereo methods.
However, in comparison to the other stereo methods, our approach performs better and more robustly.
Moreover, we measure relative pose errors as proposed by Geiger \etal \cite{Geiger2012CVPR} to measure the performance and drift of pure odometry over large-scale sequences as in the KITTI dataset.
\cref{tab:avgrpe} summarizes the results of our Semi-Direct Odometry in comparison to LIBVISO2 and Direct Odometry.
Translational and rotational errors are measured separately. Results show that our method shows less translational and rotational drift over time.
Moreover, as already seen above in the exemplary trajectory plots, the fully direct odometry has a higher rotational error than the other methods as direct alignment of frames becomes harder during large rotations.

In summary, our semi-direct approach shows accurate results for all data\-sets. Even on challenging fish eye stereo the whole trajectory can be retrieved and loop closures are found while S-PTAM fails to find any correspondences.

\subsection{Runtime}

\begin{table}
	\scriptsize
\begin{center}
    \begin{tabular}{c||c||c||c||c||c||c||c}
\hspace{-0.6em}Dataset 		& Method 		& Tracking & Mapping& \hspace{-0.6em}Constraint\hspace{-0.6em} & \hspace{-0.4em}Optimization\hspace{-0.4em} & Total & Total\\
& & & & Search & & (VO) & (SLAM)\\
\hline
\multirow{5}{*}{\hspace{-0.6em}KITTI} 	& Ours 				& \SI{26.5}{\ms} 	& \SI{36.6}{\ms}       & \SI{253.5}{\ms}                & \SI{564.6}{\ms}            & \SI{63.1}{\ms}          & \SI{881.2}{\ms} \\
& LSD-SLAM 		& - 		  & -               & -             & -               & -                & - \\
& ORB-SLAM 		& \SI{30.7}{\ms}  &\SI{254.0}{\ms}  &\SI{7.8}{\ms}  &\SI{1315.6}{\ms} & \SI{284.6}{\ms}  &\SI{1608.0}{\ms} \\
& S-PTAM 		  & \SI{71.1}{\ms}  &\SI{5.7}{\ms}    & -             &\SI{2036.9}{\ms} & \SI{77.4}{\ms}   & \SI{2114.3}{\ms}\\
& LIBVISO2 		& \SI{33.8}{\ms} & -               & -             & -               & \SI{33.8}{\ms}    & - \\
\hline
\multirow{5}{*}{\hspace{-0.6em}EuRoC} 	& Ours 				& \SI{22.6}{\ms} & \SI{39.6}{\ms}   & \SI{153.5}{\ms}        & \SI{684.2}{\ms}             & \SI{62.2}{\ms}          & \SI{899.9}{\ms}  \\
& LSD-SLAM 		& \SI{27.6}{\ms}  &\SI{85.6}{\ms}  &\SI{158.1}{\ms}  &\SI{207.3}{\ms} & \SI{113.2}{\ms}  &\SI{478.5}{\ms} \\
& ORB-SLAM 		& \SI{17.9}{\ms}  &\SI{159.2}{\ms}  &\SI{3.7}{\ms}  &\SI{535.6}{\ms} & \SI{177.1}{\ms}  &\SI{716.4}{\ms} \\
& S-PTAM 		  & \SI{47.3}{\ms}  & \SI{1.5}{\ms}   & -             & \SI{976.9}{\ms}            & \SI{48.8}{\ms} & \SI{1025.7}{\ms} \\
& LIBVISO2 		& \SI{24.8}{\ms} & -               & -             & -               & \SI{24.8}{\ms}   & - \\
\hline						

\multirow{5}{*}{\hspace{-0.6em}MAV} 	& Ours 	& \SI{17.5}{\ms} & \SI{25.8}{\ms}       & \SI{140.0}{\ms}                  & \SI{130.7}{\ms}             & \SI{43.3}{\ms}           & \SI{313.3}{\ms}  \\
& LSD-SLAM 		& \SI{28.7}{\ms}  &\SI{67.3}{\ms}  &\SI{314.0}{\ms}  &\SI{637.3}{\ms} & \SI{79.0}{\ms}  &\SI{951.3}{\ms} \\
& ORB-SLAM 		& \SI{24.3}{\ms}  &\SI{221.2}{\ms}  &\SI{11.0}{\ms}  &\SI{353.8}{\ms} & \SI{245.5}{\ms}  &\SI{610.3}{\ms} \\
& S-PTAM 		  & - & -       & -                 & -            & -          & - \\
& LIBVISO2 		& \SI{25.3}{\ms}  & -               & -             & -               & \SI{25.3}{\ms}                & - \\
	\hline
    \end{tabular}
    \end{center}
    \caption[Average Runtimes]{Average runtimes of all evaluated methods}
    \label{tab:runtimesall}
\end{table}

For state-estimation with visual odometry or SLAM on mobile robots, real-time capability is an important factor.
We thereby measure the efficiency of our method in terms of average runtime in \si{\ms}.

We measure the average runtime as well as the runtime of the different blocks because it is oftentimes sufficient if tracking can be done with high frequency since global optimization usually does not run in real-time.
The runtimes are broken down to the individual blocks: tracking, mapping, search for constraints and pose graph optimization.
Timings for all datasets are listed in \cref{tab:runtimesall}. Missing values are denoted with '-', \eg S-PTAM does not perform a constraint search as the other methods, and LIBVISO only performs tracking.
The table clearly highlights that the SLAM parts, consisting of the constraint search and pose graph optimization, are the bottleneck for all systems. 

In general, it can be seen that our approach is able to track incoming frames with \SI{30}{\hertz}. The mapping thread also runs in parallel to tracking at approximately \SI{30}{\hertz}.
However, global optimization is still very costly for all methods. Especially in large-scale sequences the runtime rises.

\subsection{Qualitative Results}

   	\begin{figure*}
      \centering
  	\includegraphics[width=0.90\linewidth]{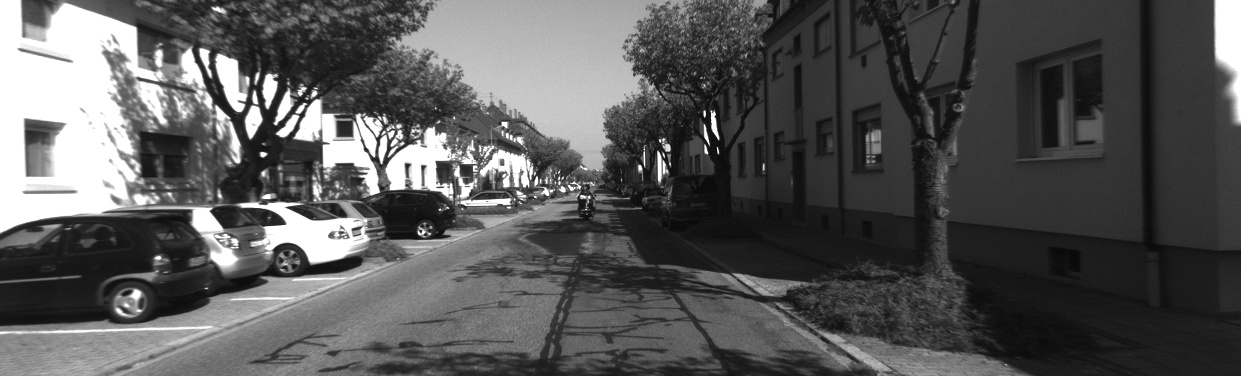}\\\vspace{0.3mm}  	
\includegraphics[width=0.90\linewidth]{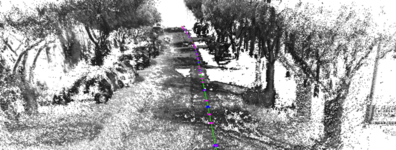}\\\vspace{0.3mm} 
  	\includegraphics[width=0.80\linewidth]{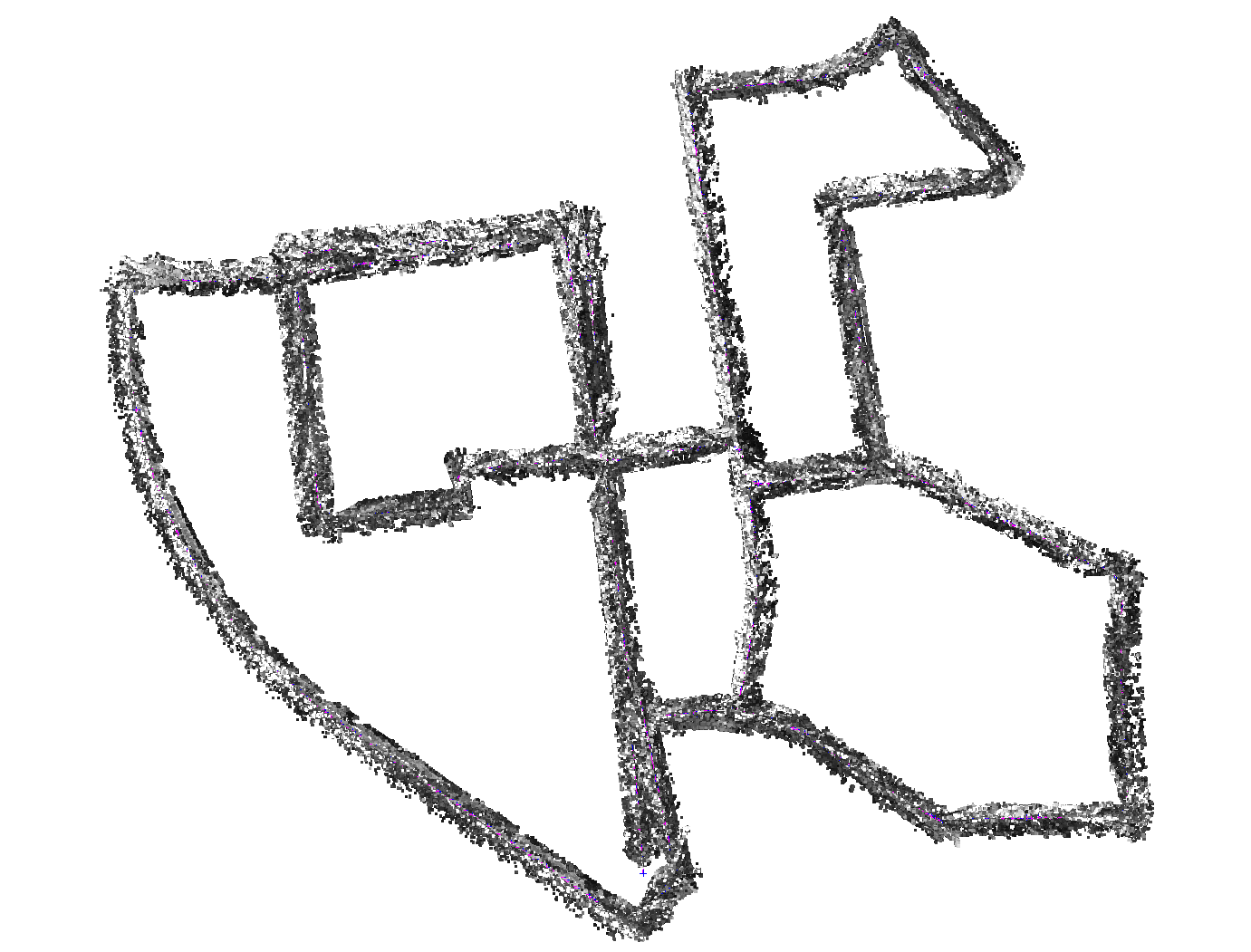}
      \caption[Semi-dense 3D Reconstruction of KITTI Dataset 00]{Semi-dense 3D Reconstruction of KITTI 00: The top image shows the reconstructed scene as captured by the camera. Below the semi-dense 3D reconstruction of this scene and the complete reconstruction of this dataset is shown. }
      \label{fig:kittireconstruction_detail}
   \end{figure*}

A major advantage of our semi-direct approach is that 3D point clouds are estimated at runtime yielding an accurate semi-dense reconstruction of the environment. 
Thus, we are not only able to estimate the current pose of the camera but also maintain a 3D map of the environment which can be used for additional tasks like obstacle avoidance.

   \begin{figure*}[t]
      \centering
  	\includegraphics[width=\linewidth]{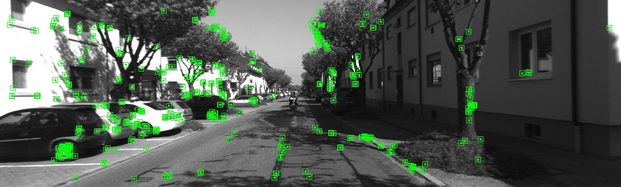}\hfill
  	\includegraphics[width=\linewidth]{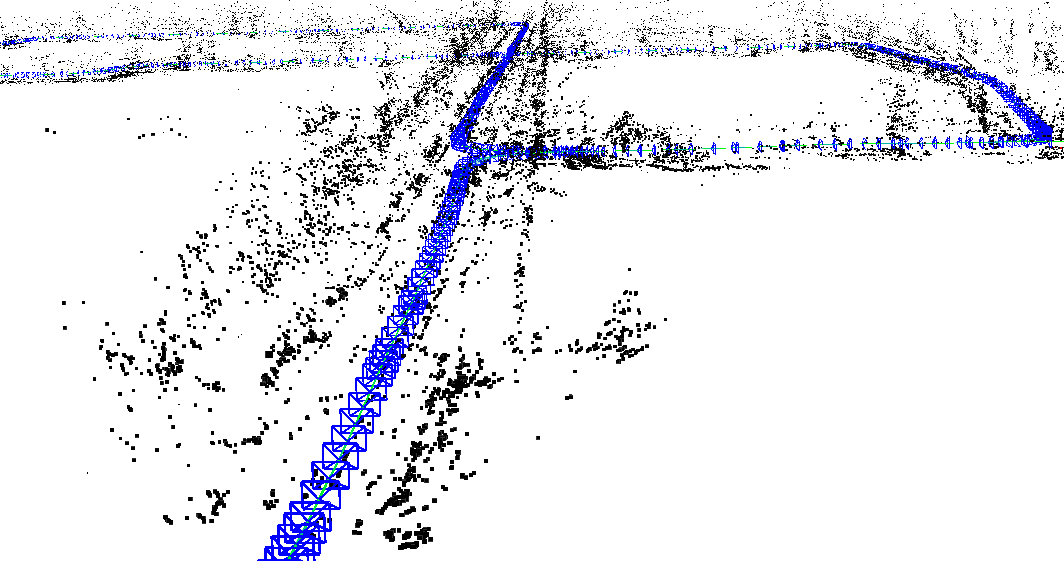}
      \caption[Sparse 3D Reconstruction of KITTI Dataset 00 by ORB-SLAM]{Sparse feature-based 3D Reconstruction of KITTI 00 by ORB-SLAM. The top view shows an exemplary scene where ORB features are tracked. The lower image shows the sparse map that is obtained by tracking ORB features.}
      \label{fig:orbreconstruction}
   \end{figure*}

Exemplary qualitative results are shown for sequence 00 of the KITTI dataset.
As can be seen in \cref{fig:kittireconstruction_detail}, an accurate and consistent 3D reconstruction is achieved by Semi-Direct SLAM. 
For direct comparison to feature-based SLAM methods, the resulting sparse map built by ORB-SLAM is shown in \cref{fig:orbreconstruction}.
While the reconstruction of ORB-SLAM only contains sparse points, our reconstruction allows detailed inference to existing objects in the scene.
Most objects, that are visible in the camera image, can be recovered in our semi-dense map. For example, one can clearly distinguish between individual trees and cars.
Contrarily, in the sparse map of ORB-SLAM one can only guess vaguely where the street is located.

\begin{figure*}
      \centering
  	\includegraphics[width=0.493\linewidth]{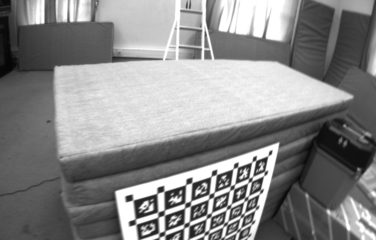}\hfill
\includegraphics[width=0.493\linewidth]{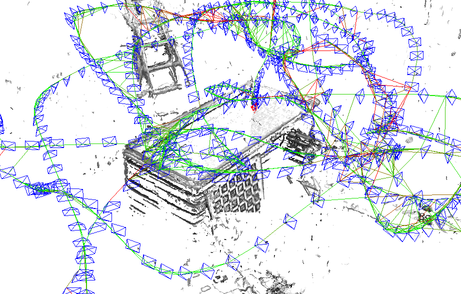}\\\vspace{0.5mm}  	
  	\includegraphics[width=\linewidth]{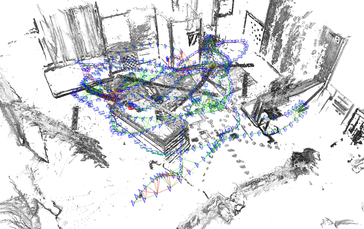}
      \caption[Semi-dense 3D Reconstruction of EuRoC Dataset]{Semi-dense 3D Reconstruction of the EuRoC Dataset V1\_02 with medium difficulty. Results show a globally consistent semi-dense map. The depicted key frame graph visualizes the trajectory. Key frames are shown in blue, while edge-constraints are shown in green and red, depending on their confidence.}
      \label{fig:eurreconstruction}
   \end{figure*}

\cref{fig:eurreconstruction} shows the estimated pose graph of the camera trajectory and reconstructed map of the medium difficult EuRoC dataset V1\_02.
The images demonstrate that our estimated 3D reconstruction is globally consistent.
The objects shown in the exemplary given camera image can easily be retrieved in the reconstructed map.

In conclusion, we state that our method builds globally consistent semi-dense 3D maps of the environment.
It is well suited for large-scale sequences as in the KITTI dataset, as well as for smaller indoor sequences like the EuRoC dataset.
We hence believe that the semi-dense 3D reconstruction yields a great benefit for autonomous visual navigation.

\section{Conclusions}

In this paper, we proposed a novel hybrid visual odometry and SLAM method that combines feature-based tracking with semi-dense direct image alignment.
Our method fuses depth estimates from motion between key frames with instantaneous stereo depth estimates.

The performance of our method has been evaluated in terms of accuracy, runtime, and scene reconstruction on three challenging datasets. 
Our experiments show that for tracking egomotion between image frames, we achieve accuracy similar to the state-of-the-art at high frame rate without the necessity to reduce the image resolution. 
Due to the feature-based tracking as prior for semi-dense direct alignment, our method is computationally less expensive and can estimate the relative camera motion in real-time.  
In future work, we plan to incorporate high frequency IMU readings and to evaluate other feature-based tracking priors, e.g. ORB features.  

\section*{Acknowledgement}

This work has been supported by the German Federal Ministry for Economic Affairs and Energy (BMWi) in the Autonomics for Industry 4.0 project InventAIRy.

\bibliographystyle{elsarticle-num}
\bibliography{references}

\begin{thebibliography}{10}
\expandafter\ifx\csname url\endcsname\relax
  \def\url#1{\texttt{#1}}\fi
\expandafter\ifx\csname urlprefix\endcsname\relax\def\urlprefix{URL }\fi
\expandafter\ifx\csname href\endcsname\relax
  \def\href#1#2{#2} \def\path#1{#1}\fi

\bibitem{engel14eccv}
J.~Engel, T.~Sch\"ops, D.~Cremers, {LSD-SLAM}: Large-scale direct monocular
  {SLAM}, in: European Conf.~on Computer Vision (ECCV), 2014, pp. 834--849.

\bibitem{krombach2016combining}
N.~Krombach, D.~Droeschel, S.~Behnke, Combining feature-based and direct
  methods for semi-dense real-time stereo visual odometry, in: Int.~Conf.~on
  Intelligent Autonomous Systems (IAS), 2016, pp. 855--868.

\bibitem{monoslam}
A.~Davison, I.~Reid, N.~Molton, O.~Stasse, Monoslam: Real-time single camera
  {SLAM}, Pattern Analysis and Machine Intelligence 29~(6) (2007) 1052--1067.

\bibitem{klein07parallel}
G.~Klein, D.~Murray, Parallel tracking and mapping for small {AR} workspaces,
  in: Int.~Symposium on Mixed and Augmented Reality {(ISMAR)}, 2007, pp.
  225--234.

\bibitem{weiss11monocular}
S.~Weiss, D.~Scaramuzza, R.~Siegwart, Monocular-{SLAM}--based navigation for
  autonomous micro helicopters in {GPS}-denied environments, Journal of Field
  Robotics 28~(6) (2011) 854--874.

\bibitem{ORB}
R.~Mur-Artal, J.~Montiel, J.~D. Tard{\'{o}}s, {ORB-SLAM}: A versatile and
  accurate monocular {SLAM} system, Trans.~on Robotics 31~(5) (2015)
  1147--1163.

\bibitem{Geiger2011IV}
A.~Geiger, J.~Ziegler, C.~Stiller, Stereoscan: Dense {3D} reconstruction in
  real-time, in: Intelligent Vehicles Symposium (IV), 2011, pp. 963--968.

\bibitem{bacs}
M.~Nieuwenhuisen, D.~Droeschel, J.~Schneider, D.~Holz, T.~L\"abe, S.~Behnke,
  Multimodal obstacle detection and collision avoidance for micro aerial
  vehicles, in: European Conf.~on Mobile Robots (ECMR), 2013, pp. 7--12.

\bibitem{mur2017orb2}
R.~Mur-Artal, J.~D. Tard{\'o}s, Orb-slam2: An open-source slam system for
  monocular, stereo, and rgb-d cameras, IEEE Transactions on Robotics 33~(5)
  (2017) 1255--1262.

\bibitem{SPTAM}
T.~Pire, T.~Fischer, J.~Civera, P.~D. Crist{\'oforis}, J.~J. Berlles, {Stereo
  Parallel Tracking and Mapping for robot localization}, in: Int.~Conf.~on
  Intelligent Robots and Systems (IROS), 2015, pp. 1373--1378.

\bibitem{SPTAM2017}
T.~Pire, T.~Fischer, G.~Castro, P.~De~Crist{\'o}foris, J.~Civera, J.~J.
  Berlles, S-ptam: Stereo parallel tracking and mapping, Robotics and
  Autonomous Systems 93 (2017) 27--42.

\bibitem{houben2016orbslam}
S.~Houben, J.~Quenzel, S.~Behnke, Efficient multi-camera visual-inertial {SLAM}
  for micro aerial vehicles, in: Int.~Conf.~on Intelligent Robots and Systems
  (IROS), 2016, pp. 1616--1622.

\bibitem{achtelik11onboard}
M.~Achtelik, M.~Achtelik, S.~Weiss, R.~Siegwart, Onboard {IMU} and monocular
  vision based control for {MAVs} in unknown in- and outdoor environments, in:
  Int.~Conf.~on Robotics and Automation (ICRA), 2011, pp. 3056--3063.

\bibitem{Leutenegger15122014}
S.~Leutenegger, S.~Lynen, M.~Bosse, R.~Siegwart, P.~Furgale, Keyframe-based
  visual-inertial odometry using nonlinear optimization, Int.~Journal of
  Robotics Research.

\bibitem{forster2017_manifold_preintegration}
C.~Forster, L.~Carlone, F.~Dellaert, D.~Scaramuzza, On-manifold preintegration
  for real-time visual-inertial odometry, IEEE Transactions on Robotics 33~(1)
  (2017) 1--21.

\bibitem{mur2017visual}
R.~Mur-Artal, J.~D. Tard{\'o}s, Visual-inertial monocular slam with map reuse,
  IEEE Robotics and Automation Letters 2~(2) (2017) 796--803.

\bibitem{comport}
A.~Comport, E.~Malis, P.~Rives, Accurate quadrifocal tracking for robust {3D}
  visual odometry, in: Int.~Conf.~on Robotics and Automation (ICRA), 2007, pp.
  40--45.

\bibitem{engel2013semi}
J.~Engel, J.~Sturm, D.~Cremers, Semi-dense visual odometry for a monocular
  camera, in: Computer Vision (ICCV), 2013 IEEE International Conference on,
  IEEE, 2013, pp. 1449--1456.

\bibitem{engel2015_stereo_lsdslam}
J.~Engel, J.~St\"uckler, D.~Cremers, Large-scale direct {SLAM} with stereo
  cameras, in: Int.~Conf.~on Intelligent Robots and Systems (IROS), 2015, pp.
  1935--1942.

\bibitem{mrsm}
J.~St\"uckler, S.~Behnke, Multi-resolution surfel maps for efficient dense 3d
  modeling and tracking, Journal of Visual Communication and Image
  Representation 25~(1) (2014) 137--147.

\bibitem{gutt}
J.~St\"uckler, A.~Gutt, S.~Behnke, Combining the strengths of sparse interest
  point and dense image registration for rgb-d odometry, in: Int.~Symposium on
  Robotics (ISR) and 8th German Conf.~on Robotics (ROBOTIK), 2014, pp. 1--6.

\bibitem{newcombe2010dense_recon}
R.~A. Newcombe, A.~Davison, Live dense reconstruction with a single moving
  camera, in: Conf.~on Computer Vision and Pattern Recognition (CVPR), 2010,
  pp. 1498--1505.

\bibitem{pizzoli2014remode}
M.~Pizzoli, C.~Forster, D.~Scaramuzza, {REMODE}: Probabilistic, monocular dense
  reconstruction in real time, in: Int.~Conf.~on Robotics and Automation
  (ICRA), 2014, pp. 2609--2616.

\bibitem{engel2018_dso}
J.~Engel, V.~Koltun, D.~Cremers, Direct sparse odometry, Pattern Analysis and
  Machine Intelligence 40~(3) (2018) 611--625.

\bibitem{schops2015_3drecon}
T.~Sch{\"o}ps, T.~Sattler, C.~H{\"a}ne, M.~Pollefeys, 3d modeling on the go:
  Interactive 3d reconstruction of large-scale scenes on mobile devices, in: 3D
  Vision (3DV), 2015 International Conference on, IEEE, 2015, pp. 291--299.

\bibitem{gallup2007_planesweep}
D.~Gallup, J.-M. Frahm, P.~Mordohai, Q.~Yang, M.~Pollefeys, Real-time
  plane-sweeping stereo with multiple sweeping directions, in: Conf.~on
  Computer Vision and Pattern Recognition (CVPR), IEEE, 2007, pp. 1--8.

\bibitem{ORB-SemiDense}
R.~Mur{-}Artal, J.~D. Tard{\'{o}}s, Probabilistic semi-dense mapping from
  highly accurate feature-based monocular {SLAM}, in: Robotics: Science and
  Systems, 2015.

\bibitem{Forster2014ICRA}
C.~Forster, M.~Pizzoli, D.~Scaramuzza, {SVO}: Fast semi-direct monocular visual
  odometry, in: Int.~Conf.~on Robotics and Automation (ICRA), 2014, pp. 15--22.

\bibitem{forster2017svo}
C.~Forster, Z.~Zhang, M.~Gassner, M.~Werlberger, D.~Scaramuzza, Svo: Semidirect
  visual odometry for monocular and multicamera systems, IEEE Transactions on
  Robotics 33~(2) (2017) 249--265.

\bibitem{piazza2018_3dmeshrecon}
E.~Piazza, A.~Romanoni, M.~Matteucci, Real-time cpu-based large-scale 3d mesh
  reconstruction, arXiv preprint arXiv:1801.05230.

\bibitem{younes2018_fdmo}
G.~Younes, D.~Asmar, J.~Zelek, Fdmo: Feature assisted direct monocular
  odometry, arXiv preprint arXiv:1804.05422.

\bibitem{Geiger2010ACCV}
A.~Geiger, M.~Roser, R.~Urtasun, Efficient large-scale stereo matching, in:
  Asian Conf.~on Computer Vision (ACCV), 2010, pp. 25--38.

\bibitem{engel2013iccv}
J.~Engel, J.~Sturm, D.~Cremers, Semi-dense visual odometry for a monocular
  camera, in: Int.~Conf.~on Computer Vision (ICCV), 2013, pp. 1449--1456.

\bibitem{g2o}
R.~K{\"u}mmerle, G.~Grisetti, H.~Strasdat, K.~Konolige, W.~Burgard, G2o: A
  general framework for graph optimization, in: ICRA, 2011, pp. 3607--3613.

\bibitem{Geiger2012CVPR}
A.~Geiger, P.~Lenz, R.~Urtasun, Are we ready for autonomous driving? the
  {KITTI} vision benchmark suite, in: Conf.~on Computer Vision and Pattern
  Recognition (CVPR), 2012, pp. 3354--3361.

\bibitem{Burri25012016}
M.~Burri, J.~Nikolic, P.~Gohl, T.~Schneider, J.~Rehder, S.~Omari, M.~W.
  Achtelik, R.~Siegwart, The {EuRoC} micro aerial vehicle datasets, The
  Int.~Journal of Robotics Research.

\bibitem{sturm12iros}
J.~Sturm, N.~Engelhard, F.~Endres, W.~Burgard, D.~Cremers, A benchmark for the
  evaluation of {RGB-D SLAM} systems, in: Int.~Conf.~on Intelligent Robots and
  Systems (IROS), 2012, pp. 573--580.

\bibitem{ROB:ROB20252}
S.~Kammel, J.~Ziegler, B.~Pitzer, M.~Werling, T.~Gindele, D.~Jagzent,
  J.~Schröder, M.~Thuy, M.~Goebl, F.~v. Hundelshausen, O.~Pink, C.~Frese,
  C.~Stiller, Team annieway's autonomous system for the 2007 darpa urban
  challenge, Journal of Field Robotics 25~(9) (2008) 615--639.

\bibitem{Directmethods}
M.~Irani, P.~Anandan, About direct methods, in: Int. Workshop on Vision
  Algorithms: Theory and Practice, Int.~Conf.~on Computer Vision (ICCV),
  Springer-Verlag, London, UK, UK, 2000, pp. 267--277.

\bibitem{NikolicRBGLFS14}
J.~Nikolic, J.~Rehder, M.~Burri, P.~Gohl, S.~Leutenegger, P.~T. Furgale,
  R.~Siegwart, A synchronized visual-inertial sensor system with {FPGA}
  pre-processing for accurate real-time {SLAM}, in: Int.~Conf.~on Robotics and
  Automation (ICRA), 2014, pp. 431--437.

\bibitem{Beul2017}
M.~Beul, N.~Krombach, M.~Nieuwenhuisen, D.~Droeschel, S.~Behnke,
  \href{https://doi.org/10.1007/978-3-319-54927-9_15}{Autonomous Navigation in
  a Warehouse with a Cognitive Micro Aerial Vehicle}, Springer International
  Publishing, Cham, 2017, pp. 487--524.
\newblock \href {http://dx.doi.org/10.1007/978-3-319-54927-9_15}
  {\path{doi:10.1007/978-3-319-54927-9_15}}.
\newline\urlprefix\url{https://doi.org/10.1007/978-3-319-54927-9_15}

\bibitem{droeschel2016multilayered}
D.~Droeschel, M.~Nieuwenhuisen, M.~Beul, D.~Holz, J.~St{\"u}ckler, S.~Behnke,
  Multilayered mapping and navigation for autonomous micro aerial vehicles,
  Journal of Field Robotics 33~(4) (2016) 451--475.

\end{thebibliography}

\end{document}